\documentclass[lettersize,journal]{IEEEtran}
\usepackage{amsmath,amsfonts}
\usepackage{algorithmic}
\usepackage{algorithm}
\usepackage{array}
\usepackage[caption=false,font=normalsize,labelfont=sf,textfont=sf]{subfig}
\usepackage{textcomp}
\usepackage{mathtools}
\usepackage{stfloats}
\usepackage{url}
\usepackage{verbatim}
\usepackage{graphicx}
\usepackage{cite}
\usepackage{graphicx}
\usepackage{amsmath}
\usepackage{amssymb}
\usepackage{booktabs}
\usepackage{booktabs}
\usepackage{cite}
\usepackage{url}
\usepackage{bm}
\usepackage{multirow}
\usepackage{booktabs}
\usepackage{algorithm,algorithmic}
\usepackage{footnote}
\usepackage{amsmath,amssymb, amsthm,amsfonts}
\usepackage{xcolor}
\usepackage{hyperref}  
\usepackage{bm}
\usepackage{amssymb}
\usepackage{latexsym}
\usepackage{dsfont}
\usepackage{multirow}
\usepackage{amsfonts}
\usepackage{pifont}
\usepackage{amsmath}
\usepackage{multirow}
\usepackage{graphicx}
\usepackage{array}
\usepackage{amssymb}
\usepackage{amsmath}
\usepackage{algorithmic}
\usepackage{color}
\usepackage{cite}

\usepackage[utf8]{inputenc}
\begin{document}

\title{Multi-Modal Cross-Domain Alignment Network for Video Moment Retrieval}

\author{Xiang Fang*, Daizong Liu*~\IEEEmembership{Student~Member,~IEEE}, Pan Zhou~\IEEEmembership{Senior~Member,~IEEE}, Yuchong Hu~\IEEEmembership{Member,~IEEE}
\IEEEcompsocitemizethanks{
\IEEEcompsocthanksitem *indicates co-first authors. This work is supported by National Natural Science Foundation of China (NSFC) under grant no. 61972448. (\emph{Corresponding author: Pan Zhou}.)
\IEEEcompsocthanksitem Xiang Fang is with Hubei Key Laboratory of Distributed System Security, Hubei Engineering Research Center on Big Data Security, School of Cyber Science and Engineering, Huazhong University of Science and Technology, Wuhan, Hubei 430074, China. (E-mail: xfang9508@gmail.com).
\IEEEcompsocthanksitem Daizong Liu is with Wangxuan Institute of Computer Technology, Peking University, No. 128, Zhongguancun North Street, Beijing, China (E-mail: dzliu@stu.pku.edu.cn).
\IEEEcompsocthanksitem Pan Zhou is with Hubei Key Laboratory of Distributed System Security, Hubei Engineering Research Center on Big Data Security, School of Cyber Science and Engineering, Huazhong University of Science and Technology, Wuhan, Hubei 430074, China. (e-mail: panzhou@hust.edu.cn).
\IEEEcompsocthanksitem Yuchong Hu is with the School of Computer Science and Technology, Key Laboratory of Information Storage System Ministry of Education of China, Huazhong University of Science and Technology, Wuhan 430074, China (E-mail: yuchonghu@hust.edu.cn).
\IEEEcompsocthanksitem $\copyright$ 2021 IEEE. Personal use of this material is permitted.  Permission from IEEE must be obtained for all other uses, in any current or future media, including reprinting/republishing this material for advertising or promotional purposes, creating new collective works, for resale or redistribution to servers or lists, or reuse of any copyrighted component of this work in other works.
}}



\maketitle

\begin{abstract}
As an increasingly popular task in multimedia information retrieval, video moment retrieval (VMR) aims to localize the target moment from an untrimmed video according to a given language query. Most previous methods depend heavily on numerous manual annotations (\textit{i.e.}, moment boundaries), which are extremely expensive to acquire in practice. In addition, due to the domain gap between different datasets, directly applying these pre-trained models to an unseen domain leads to a significant performance drop.
In this paper, we focus on a novel task: cross-domain VMR, where fully-annotated datasets are available in one domain (``source domain''), but the domain of interest (``target domain'') only contains unannotated datasets. As far as we know, we present the first study on cross-domain VMR. To address this new task,  we propose a novel \textbf{M}ulti-\textbf{M}odal \textbf{C}ross-\textbf{D}omain  \textbf{A}lignment (MMCDA) network to transfer the annotation knowledge from the source domain to the target domain.  However, due to the domain discrepancy between the source and target domains and the semantic gap between videos and queries, directly applying trained models to the target domain generally leads to a performance drop.  To solve this problem, we develop three novel modules:  (i) a domain alignment module is designed to align the feature distributions between different domains of each modality; (ii) a cross-modal alignment module aims to map both video and query features into a joint embedding space and to align  the feature distributions between different modalities in the target domain; (iii) a specific alignment module tries to obtain the fine-grained similarity between a specific frame and the given query for optimal localization. By jointly training these three modules, our MMCDA can learn domain-invariant and semantic-aligned cross-modal representations. Extensive experiments on three challenging benchmarks {(ActivityNet Captions, Charades-STA and TACoS)} illustrate that our cross-domain method MMCDA outperforms all state-of-the-art single-domain methods.
Impressively, MMCDA raises the performance by more than 7\% in representative cases, which demonstrates its effectiveness.
\end{abstract}

\section{Introduction}\label{section:intro}
\begin{figure}[t]
\centering
\includegraphics[width=0.5\textwidth]{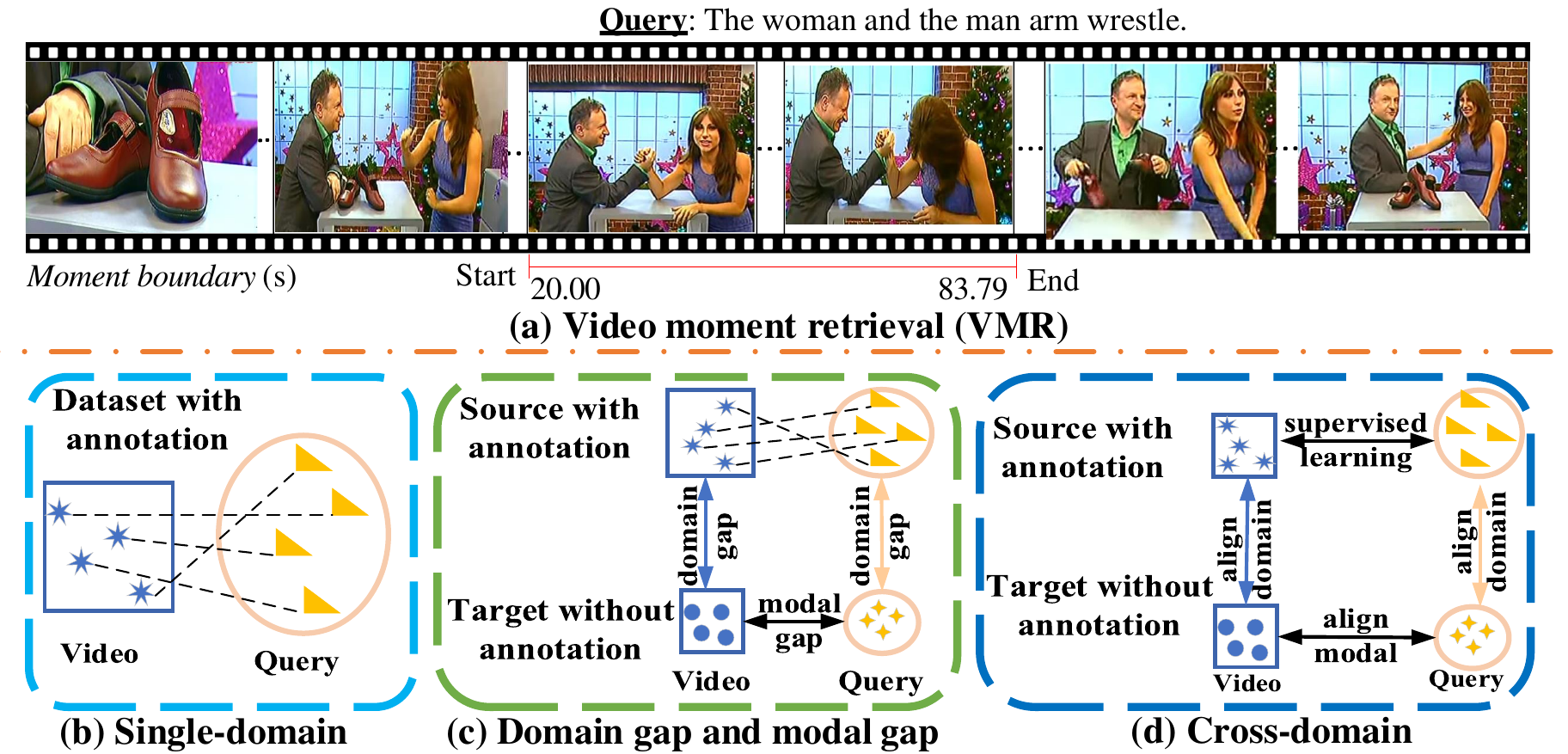}
\vspace{-23pt}
\caption{(a) An example of the video moment retrieval (VMR) task. In an annotated dataset, given a language query, VMR aims to retrieve and rank videos frames based on how well they match the text description. (b) Single-domain models (previous VMR methods) rely on numerous annotations to acquire the annotation knowledge and the semantic relationship between  videos and queries. These single-domain models are trained and tested on the same domain (the same dataset). (c) Due to the domain gap (domain discrepancy) and the modal gap (semantic gap), we cannot directly apply the model trained in the source domain to the target domain. (d) Our cross-domain model aims to transfer the annotation knowledge from the source domain to the target domain by aligning different domains and modalities.}
 \vspace{-0.38cm}
\label{fig:wentimoxing}
\end{figure}

As an important yet challenging multimedia task, video moment retrieval (VMR) \cite{wang2022cross,yang2022video,zeng2022moment} has received increasing attention due to its wide potential applications, such as human-computer interaction \cite{zhu2020vision} and video understanding \cite{liu2023exploring,wang2025taylor,fang2026towardsicml,kuai2026dynamic,wang2025point,fang2025your,zhang2025monoattack,fang2023hierarchical,liu2024towards,yang2025eood,fang2020double,fang2026cogniVerse,lei2025exploring,fang2023you,wang2025dypolyseg,fang2025hierarchical,yan2026fit,fang2025adaptive,wang2026topadapter,cai2025imperceptible,fang2026slap,wang2026reasoning,fang2026immuno,wang2026biologically,fang2026disentangling,wang2025reducing,fang2026advancing,fang2026unveiling,wang2026from,liu2023conditional,liu2026attacking,fang2026rethinking,wang2025seeing,fang2026towards,fang2025multi,fang2024fewer,liu2024pandora,fang2024multi,fang2025turing,fang2024not,liu2023hypotheses,fang2024rethinking,liu2024unsupervised,fang2023annotations,xiong2024rethinking,fang2021unbalanced,wang2025prototype,zhang2025manipulating,fang2026align,tang2024reparameterization,fang2025adaptivetai,tang2025simplification,fang2021animc,cai2026towards,fang2020v}.
Given a sentence query, VMR aims to ground the most relevant moment from an untrimmed video. As shown in Fig.~\ref{fig:wentimoxing}(a), most of the contents in the video are irrelevant to the query while only a short moment matches it.  A well-designed model requires grounding accurate start and end timestamps of the target moment. In practice, VMR is an extremely challenging task because the desired model should (i) cover various  moment lengths in multiple scenarios; (ii) bridge the semantic gap between different modalities (video and query);
(iii) understand the semantic details of different modalities to extract modal-invariant features for optimal  retrieval.

Most previous VMR methods \cite{zeng2021multi,tang2021frame,liu2022exploring1,zhang2021multi,cao2020strong,Liu_2021_CVPR,lei2020tvr,zhao2021cascaded} refer to a fully-supervised setting where all
moment-query pairs and moment boundaries are annotated in detail. For instance, some of them \cite{xiao2021boundary,wang2021structured,Xu0PSSS19,ChenJ19a,ge2019mac,LiuQDZ20,LiuQLDZX20} utilize a scan-and-ranking framework that first pre-defines
a set of candidate segment proposals and then calculates their matching degree with query semantics for ranking and localization;
others \cite{li2021proposal,rodriguez2020proposal,wang2020temporally,ZhangDWWD19,ZhangLZX19} directly predict the moment boundary by a regression strategy. Although these respectable methods have achieved decent progress, they are data-hungry, relying on numerous manual annotations (moment boundaries), which  limits their applications. 
For example, more than 300,000 unannotated videos (with 80,000 hours of video length) are uploaded to YouTube every day. Obviously, it is not feasible to manually annotate such large-scale video data. 
To alleviate the reliance to a certain extent, some weakly-supervised methods \cite{ZhangLZZH20,2021LoGAN,ChenMLW19,DuanHGW0H18,MithunPR19,MaYKLKY20,LinZZWL20,song2020weakly,zhang2020counterfactual} are proposed to only utilize coarse video-level annotations for training. Unfortunately, this still requires us to understand each video and to provide the corresponding video-level annotation, which is significantly time-consuming and labor-intensive. 

Although the above supervised methods have achieved great progress in some cases, they are all trained in a supervised learning manner based on manually annotated datasets, which are always expensive to collect in practice.
In most multimedia applications, we always collect a few fully-annotated datasets and a massive number of unannotated datasets. Hence, we pose a brand-new task: \emph{can we transfer the annotation knowledge from the fully-annotated source dataset to the unannotated target dataset?} 
We demonstrate this new task \emph{``cross-domain video moment retrieval''} (cross-domain VMR) as shown in Fig.~\ref{fig:wentimoxing}(d). To the best of our knowledge, there is no such task proposed in previous works for VMR. Different from previous VMR works under the single-domain setting as shown in Fig.~\ref{fig:wentimoxing}(b), our  cross-domain model first gains the annotation knowledge from the source domain, and then transfers the knowledge into the target domain as shown in Fig.~\ref{fig:wentimoxing}(d).
Due to the domain discrepancy and semantic gap as shown in Fig.~\ref{fig:wentimoxing}(c), if previous single-domain VMR methods are directly employed to the unannotated target dataset, they will suffer from a performance drop. 
Therefore, cross-domain VMR is a significant and challenging task.

{Although many single-modal domain adaptation works \cite{wang2022information,zhang2022latent,jing2022adversarial,wang2022uncertainty,tao2022dreamt} have been proposed, they can only learn the feature representation of one modality. As a significant multimedia task, the VMR task requires the alignment knowledge of two modalities. Therefore, previous single-modal domain adaptation methods cannot be directly used for the VMR task due to the inability to preserve the relations between two modalities at the same time.
Recently, some multi-modal domain adaptation methods have been proposed for some specific multimedia tasks, \textit{e.g.}, action recognition \cite{munro2020multi} and emotion recognition \cite{wang2022multi}. Someone might have a question: if we can introduce these methods to our cross-domain VMR task? Unfortunately, there is a large gap between VMR and other multimedia tasks. Directly introducing these works into VMR meets limitations of the more complex multi-modal scenarios and the larger domain gap in VMR datasets. Different from previous multi-modal domain adaptation tasks that focus on matching visual and textual information for a classification-based task, our proposed cross-domain VMR is more challenging and needs to transfer the query-related moment knowledge across different datasets in a more complicated retrieval task.}


To solve this task, we propose a Multi-Modal Cross-Domain  Alignment (MMCDA) network to learn domain-invariant and semantic-aligned cross-modal representations.
As shown in Fig.~\ref{fig:kuangjia}, MMCDA contains three main parts: (i) two shared multi-modal feature encoders to extract discriminative modal-specific features, (ii) a cross-modal attention part to learn domain-agnostic video-query interaction, (iii) a multi-modal  cross-domain alignment part to close the domain discrepancy and the semantic gap via three carefully-designed alignment modules: domain alignment, cross-modal alignment, and specific alignment. 

To alleviate the domain discrepancy, we first design a brand-new domain alignment module to minimize the maximum mean discrepancy (MMD) distance between different domains. Moreover, an intra-sample distribution function and an inter-sample distribution function are proposed to align the feature distributions of the source and target domains.
Based on this domain alignment module, we can obtain the domain-invariant representation for each modality. However, only aligning different domains is not enough 
due to the existence of the semantic gap between videos and queries in the unannotated target domain. To bridge such a semantic gap, we further develop a cross-modal alignment module. In particular, we first
project these multi-modal features into a joint embedding space, which aims to maximize the similarity between the video and corresponding query  and to minimize the similarity between the video and non-matching query, versa vice. Then, we align the feature distributions of different modalities based on the intra-sample distribution function and the inter-sample distribution function. Considering that some specific frames near a moment boundary might be assigned to another moment incorrectly, we also propose a specific alignment module to improve the  retrieval accuracy. Particularly, we enhance the characteristics of the specific frame by maximizing its probability belonging to the corresponding query.
By integrating the aforementioned three alignment modules, our MMCDA is able to achieve significant performance on the target domain without any annotations.

Our main contributions are summarized as follows:
\begin{itemize}
\item We present a novel task: cross-domain VMR, which aims to transfer the annotation knowledge from the source domain to the target domain. To our best knowledge, our proposed MMCDA is the first attempt to make VMR adaptive to different domains rather than a single domain in previous VMR methods.
\item Three alignments modules (domain alignment, cross-modal alignment, and specific alignment) are designed in MMCDA to learn  domain-invariant and semantic-aligned cross-modal representations.
\item Experimental results on three popular datasets {(ActivityNet Captions, Charades-STA and TACoS)} show the effectiveness of MMCDA. To our surprise, it significantly outperforms state-of-the-art single-domain VMR methods in representative cases. Extensive ablation studies  illustrate the functions of each module.
\end{itemize}

\section{Related Works}  \label{section:related}

\noindent \textbf{Fully-supervised video moment retrieval.}
Most of the existing methods are  fully supervised \cite{zhang2020temporal}, requiring all annotated moment-query pairs and labeled moment boundaries. With adequate annotations, these
methods try to align multi-modal features to predict the moment boundary from an untrimmed video according to a given query.
Based on a two-stage multi-modal matching strategy, many works \cite{xiao2021boundary,wang2021structured,Xu0PSSS19,ChenJ19a,ge2019mac,LiuQDZ20,LiuQLDZX20} first sample candidate moment proposals, and subsequently integrate a given query with moment representations by a matrix operation. 
For example, by introducing a multi-level model, Xu \emph{et al.} \cite{Xu0PSSS19} integrates visual and textual features earlier and further re-generates queries as an auxiliary task. To enhance the video representation understanding, Chen \emph{et al.} \cite{ChenKN17} captures the evolving fine-grained frame-by-word interactions between videos and queries. Although these methods achieve good performances, they are severely proposal-dependent and time-consuming. 
Without using proposals, the latest  methods \cite{li2021proposal,rodriguez2020proposal,wang2020temporally,ZhangDWWD19,ZhangLZX19} integrate the query representation with those video moments individually, and then calculate their matching scores.
{By using a two-phase cross-modal interaction, Sun \emph{et al.} \cite{sun2021maban} leverage multi-agent reinforcement learning to reason each  moment boundary.
To sufficiently understand cross-modal semantic,
Hu \emph{et al.} \cite{hu2021coarse} propose a deep cross-modal semantic alignment network for moment localization.}


\noindent \textbf{Weakly-supervised video moment retrieval.}
The aforementioned fully-supervised methods are annotation-dependent. In order to alleviate the dependence to a certain extent, several weakly-supervised VMR methods \cite{DuanHGW0H18,MithunPR19,ChenMLW19,LinZZWL20} are under a weakly-supervised setting,
which  only utilizes coarse video-level annotations to obtain accurate retrieval. 
For a weakly-supervised VMR task, Duan \emph{et al.} \cite{DuanHGW0H18} decompose it into two sub-tasks: event captioning and query localization. Duan \emph{et al.} \cite{DuanHGW0H18} first assume that each caption describes only one temporal moment, and then design a cycle network to train the model. 
As the pioneering work for weakly-supervised VMR, Mithun \emph{et al.} \cite{MithunPR19} learn a joint representation between the video and the query by proposing a text-guided-attention  network and utilizing an attention weight. To improve the exploration and exploitation, Lin \emph{et al.} \cite{LinZZWL20} choose the top-K proposals and measure the semantic similarity between the video and the query. By proposing Semantic Completion Network, Lin \emph{et al.} \cite{LinZZWL20} treat the masked query as input and predict the masked words from the video features.

\noindent \textbf{Unsupervised domain adaptation.}
Unsupervised domain adaptation (UDA) aims to transfer the knowledge from the label-rich source domain to the  unlabelled target domain, which is called \emph{domain shift} \cite{Na_2021_CVPR}.
This shift usually encounters some challenges: (i) the collected data are from diverse resources \cite{ganin2016domain}; (ii) the training data in the target domain is limited \cite{hosseini2018augmented}; (iii) the source and target domains have different modalities \cite{huang2018mhtn}.
A general solution is to align the feature distribution between the two domains. Many UDA methods align the feature distribution by minimizing the disagreement of data distribution in different domains \cite{GrettonBRSS12}, or learning an adversarial domain-classifier to find domain-invariant features \cite{TzengHSD17}.
{For example, for the activity recognition task, Song \emph{et al.} \cite{song2021spatio} establish the cross-modal domain alignment via self-supervised contrastive framework to learn the joint clip-level and video-level representation alignment for video-based UDA. Considering that video data is usually associated with multi-modal information (RGB and optical flow), Kim \emph{et al.} \cite{kim2021learning} propose a unified framework for video domain
adaptation, which simultaneously regularizes cross-modal
and cross-domain feature representations. As for the text-video retrieval task, Chen \emph{et al.} \cite{chen2021mind} propose a new benchmark and investigate the UDA task for text-video retrieval in this setting.
Based on standard backpropagation training,
Ganin \emph{et al.} \cite{ganin2015unsupervised} present  a network that allows large-scale training by massive annotated data in the source domain and a large number of unannotated data in the target domain.
By introducing adversarial learning to model domain shift between different domains, Long \emph{et al.} \cite{long2018deep} design a deep domain adaptation hashing network for binary representation generation. Long \emph{et al.} \cite{long2018conditional} propose a conditional adversarial domain adaptation network, which conditions the adversarial adaptation methods on discriminative information conveyed when predicting classification results.
} However, these methods cannot apply to our cross-domain VMR task because (i) most UDA methods only align the unimodal features, while we need to align the multi-modal features (both queries and videos); (ii) many UDA methods aim to tackle  classification-based tasks (\textit{e.g.}, {activity recognition \cite{li2021semisupervised}, object detection \cite{guan2021uncertainty} and text-video retrieval \cite{chen2021mind}}) rather than our complicated retrieval task. 
Thus, cross-domain VMR is more challenging than general UDA tasks.
%

\section{Proposed MMCDA Network}   \label{section:meth}
Firstly, we carefully formulate our novel problem and introduce our proposed MMCDA. Then, we present our framework  (feature encoder, cross-modal attention, and multi-modal cross-domain alignment). In the multi-modal cross-domain alignment, we highlight three well-designed alignment modules: domain alignment, cross-modal alignment, and specific alignment.  Finally, we demonstrate our training and inference phases in detail.

\begin{figure*}[t]
\centering
\includegraphics[width=\textwidth]{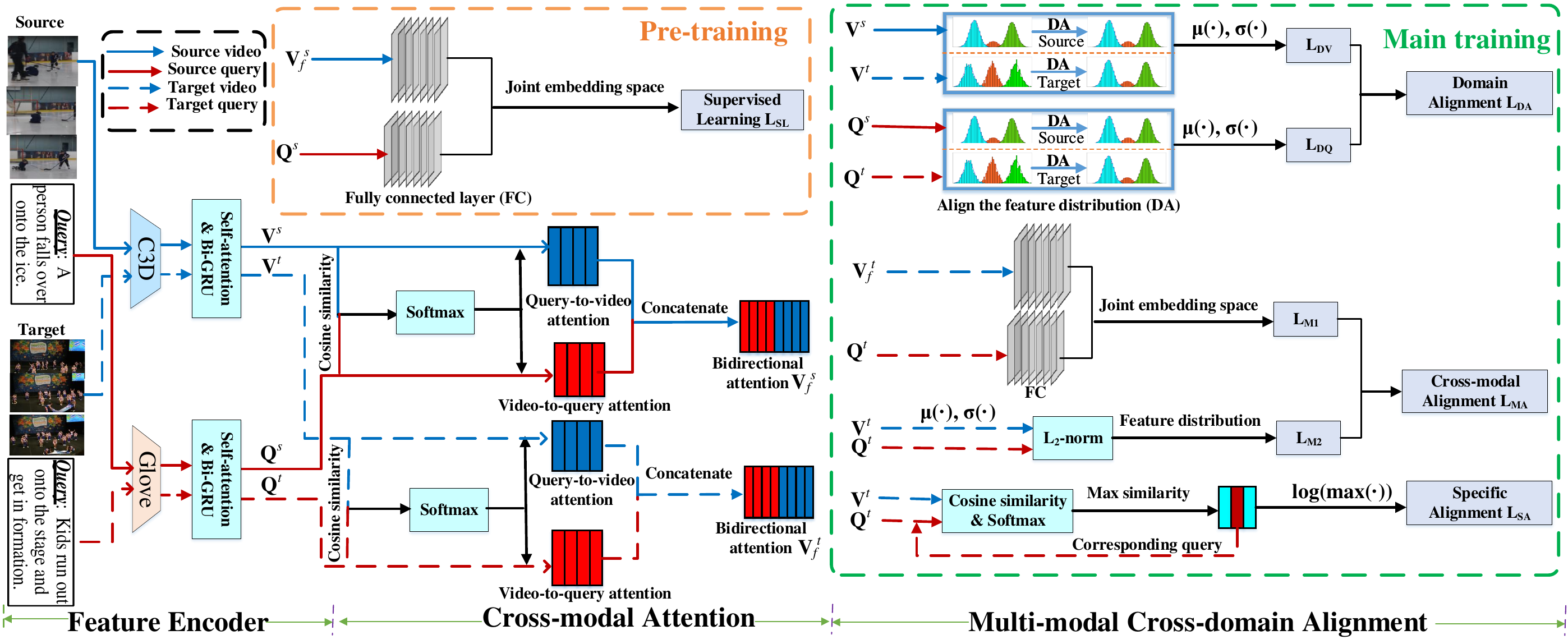}
\vspace{-25pt}
\caption{A brief framework of our proposed MMCDA for the cross-domain VMR task. MMCDA contains three parts: the shared feature encoders to extract the video and query features in each domain, the cross-modal attention to capture the video-query relations, and the multi-modal cross-domain alignment to close the domain discrepancy and the semantic gap. MMCDA aims to transfer the alignment knowledge from the source domain to the target domain. Best viewed in color.
}
 \vspace{-0.38cm}
\label{fig:kuangjia}
\end{figure*}
\subsection{Problem Formulation}
In our cross-domain VMR task, we collect a fully-annotated source dataset as:
{\[\bm{X}^s=\{ \{\bm{V}_i^{s }\}_{i=1}^{I^s}, \{\bm{Q}_j^{s}\}_{j=1}^{J^s},\{(s_j,e_j)^s\}_{j=1}^{J^s}\},\]
where $\bm{V}_{i}^{s}$ is the $i$-th untrimmed video; $\bm{Q}_j^{s}$ and $(s_j,e_j)^s$ are the $j$-th sentence query and moment boundary, respectively; $(s_j,e_j)$ is the start and end timestamps of the $j$-th moment, which corresponds to the $j$-th query semantically.}

Also, we are given an unannotated target dataset:
{\small\[\bm{X}^{t}=\{ \{\bm{V}_{k}^{t}\}_{k=1}^{I^t}, \{\bm{Q}_{l}^{t}\}_{l=1}^{J^t},\{(\cdot,\cdot)^t \}_{l=1}^{J^t}\},\]\normalsize
where ``$(\cdot,\cdot)$'' means that the target moment boundary  is not available.}  $I^s$/$I^t$ and $J^s$/$J^t$  are the numbers of source/target
videos and source/target queries, respectively\footnote{In this paper, the superscript $s$ represents the source domain, and the superscript $t$ represents the target domain.}.
To predict the target moment boundary $(\cdot,\cdot)^t$, we propose MMCDA to transfer the annotation knowledge from the source domain into the target domain.



As shown in Fig. \ref{fig:kuangjia}, our  MMCDA contains three main parts: feature encoder, cross-modal attention, and  multi-modal cross-domain alignment.  The feature encoder aims to extract the discriminative features ($\bm{V}_i^{s}$/$\bm{V}_k^{t}$ and $\bm{Q}_j^{s}/\bm{Q}_l^{t}$) with modal-specific information. The cross-modal attention tries to learn the domain-agnostic video-query  interaction by integrating both query-to-video  and video-to-query information as the bidirectional attention ($\bm{V}_{f}^{s}$/$\bm{V}_{f}^{t}$).
The multi-modal cross-domain alignment  aims to learn domain-invariant and semantic-aligned cross-modal representations through three alignment modules: domain alignment, cross-modal alignment, and specific alignment.

\subsection{Feature Encoder}
\label{subsection:bianma}
\noindent \textbf{Video encoding.}
Given a video input, following \cite{LiuQLDZX20},  we first utilize a pre-trained C3D model \cite{TranBFTP15} to extract the frame-wise features. Then,  we leverage a multi-head self-attention module \cite{VaswaniSPUJGKP17} to obtain the semantic dependencies in the long video context. To incorporate the contextual information, we further exploit a bi-directional GRU
(Bi-GRU) network \cite{chung2014empirical}.
In the source domain, we denote the encoded video representation as $\bm{V}_{i}^{s}= \{\bm{v}_{1}^{s}, \bm{v}_{2}^{s}, \cdots, \bm{v}_{T^{s}}^s\}^{\top} \in\mathbb{R}^{T_i^{s} \times d}$, where $T_i^s$ is the frame number of the $i$-th source video, $d$ is the embedding dimension.
Similarly, we leverage the same video encoder for the target domain.  The extracted features can be represented  as
$\bm{V}_{k}^{t} = \{\bm{v}_{1}^{t}, \bm{v}_{2}^{t}, \cdots, \bm{v}_{T^{t}}^t\}^{\top} \in \mathbb{R}^{T_k^{t} \times d}$,
where $T_k^{t}$ is the frame number of the $k$-th target video.

\noindent \textbf{Query encoding.} Given a query as input, a Glove embedding \cite{PenningtonSM14} is employed to obtain the word-wise features. Similar to the video encoding, we feed these word features into another Bi-GRU network and a multi-head self-attention module to further integrate the sequential query representations. In the source domain, we denote the encoded query representation as $\bm{Q}_{j}^{s} = \{\bm{q}_{1}^{s}, \bm{q}_{2}^{s}, \cdots,  \bm{q}_{N^{s}}^s\}^{\top} \in \mathbb{R}^{N_j^{s} \times d}$, where $N_j^{s}$ is the word number of the $j$-th source query, $d$ is the embedding dimension. Similarly, in the target domain, we use the same query encoder to obtain the query representation $\bm{Q}_{l}^{t} = \{\bm{q}_{1}^{t}, \bm{q}_{2}^{t}, \cdots, \bm{q}_{N^{t}}^t\}^{\top} \in \mathbb{R}^{N_l^{t} \times d}$,
where $N_l^t$ is the word number of the $l$-th target query.


\subsection{Cross-Modal Attention}
\label{subsection:zhuyili}
After encoding both videos and queries, we learn domain-agnostic video-query interactions through two attention mechanisms, shown in Fig. \ref{fig:kuangjia}.

Firstly, we  compute two cosine similarity matrices: (i) $\bm{C}^s(\bm{V}_{i}^s,\bm{Q}_{j}^s)\in \mathbb{R}^{T_i^{s}\times N_j^{s}}$ between $\bm{V}_{i}^s$ and $\bm{Q}_{j}^s$ in the source domain, (ii) $\bm{C}^t({\bm{V}_{k}^t,\bm{Q}_{l}^t})\in \mathbb{R}^{T_k^{t} \times N_l^{t}}$ between $\bm{V}_{k}^t$ and $\bm{Q}_{l}^t$ in the target domain:
\begin{align}\label{xiangsidu}
\bm{C}_{i,j}^s=\frac{||\bm{v}_i^{s\top}\bm{q}_j^s||_2}{||\bm{v}_i^s||_2||\bm{q}_j^s||_2},
\bm{C}_{k,l}^{t}=\frac{||\bm{v}_{k}^{t\top}\bm{q}_{l}^{t}||_2}{||\bm{v}_{k}^{t}||_2||\bm{q}_{l}^{t}||_2},
\end{align}
where $\bm{Q}_{j}^s$/$\bm{Q}_{l}^t$ is the corresponding query semantically-matching video $\bm{V}_{i}^s$/$\bm{V}_{k}^t$, $(i,j)$ and $(k,l)$ is the coordinates of $\bm{C}^s$ and  $\bm{C}^{t}$; $\bm{v}_i^s$ ,$\bm{v}_k^{t}$and $\bm{q}_j^s$ $\bm{q}_l^{t}$ are the  frames and  words in source and target domains, respectively.
Specifically, $\bm{C}_r^s$ and $\bm{C}_c^s$ mean the row-wise and column-wise normalizations of $\bm{C}^s(\bm{V}_{i}^s,\bm{Q}_{j}^s)$ by the Softmax function, respectively.

Then, in the source domain, we design the bidirectional attention module as follows:
\[\mathcal{X}_{i}^s=\bm{C}_r^s\bm{Q}_{i}^s,\mathcal{Y}_{i}^s=\bm{C}_r^s\bm{C}_c^{s\top}\bm{V}_{i}^s.\]
Similarly, in the target domain, the bidirectional attention module is as follows:
\[\mathcal{X}_{k}^t=\bm{C}_r^{t}\bm{Q}_{k}^{t},\mathcal{Y}_{k}^{t}=\bm{C}_r^{t}\bm{C}_c^{t\top}\bm{V}_{k}^{t},\]
where $\mathcal{X}_{i}^s\in \mathbb{R}^{T_i^s \times d}$ and $\mathcal{X}_{k}^t\in \mathbb{R}^{T_k^t \times d}$ are the video-to-query attention weights; $\mathcal{Y}_{i}^s\in \mathbb{R}^{T_i^s \times d}$ and $\mathcal{Y}_{k}^t\in \mathbb{R}_k^{T^t \times d}$ are the query-to-video attention weights.

Finally, we compose the fused frame-wise feature sequence with the output of the bidirectional attention:
\begin{align}\label{shuangxiangdeshuchu}
\begin{split}
\{\bm{V}_{f}^{s}\}_{i}=\text{Bi-GRU}([\bm{V}_{i}^s;\mathcal{X}_{i}^s;\bm{V}_{i}^s\odot\mathcal{X}_{i}^s;\bm{V}_{i}^s\odot\mathcal{Y}_{i}^s]), \\
\{\bm{V}_{f}^{t}\}_{k}=\text{Bi-GRU}([\bm{V}_{k}^t;\mathcal{X}_{ k}^t;\bm{V}_{k}^t\odot\mathcal{X}_{k}^t;\bm{V}_{k}^t\odot\mathcal{Y}_{k}^t]),
\end{split}
\end{align}
where $\{\bm{V}_{f}^{s}\}_{i} \in \mathbb{R}^{T_i^s \times d}$; $\{\bm{V}_{f}^{t}\}_{k} \in \mathbb{R}^{T_k^t \times d}$; $\odot$ is the element-wise multiplication operation.

\subsection{Multi-Modal Cross-Domain Alignment}
\label{subsection:zongkuangjia}
So far, we encode videos and queries  ($\bm{V}_i^{s},\bm{Q}_j^{s},\bm{V}_k^{t},\bm{Q}_l^{t}$) in both domains, and obtain the fused features ($\bm{V}_{f}^{s}, \bm{V}_{f}^{t}$) by cross-modal interactions.
However, when we transfer the localization model trained on the source domain into the target domain, we still meet the following challenges: (i) how to close the domain discrepancy of the same modality between different domains;  (ii) how to close the semantic gaps of the video and query in each domain;  (iii) how to learn more representative frame-wise features to perform more accurate localization. As shown in Fig. \ref{fig:kuangjia}, we address the above three challenges by proposing three alignment modules: (i) a domain alignment (DA) module  to align the feature distributions between different domains of each modality; (ii)  a cross-modal alignment (MA) module to map both video and query features into a joint embedding space and to align  the feature distributions between different modalities in the target domain; (iii) a specific alignment (SA) module to enhance the characteristics of a specific frame by maximizing its probability belonging to the corresponding query.

\noindent \textbf{Domain alignment.}
In our cross-domain VMR task, for each
modality (video or query), the domain discrepancy  is mainly due to the gap of feature distributions across different domains. To alleviate the domain discrepancy, we propose a domain alignment module for relieving the divergence of domain statistics. The basic idea is that the moment calculated on different domains are the
same if the distributions of source and target domains are identical. To alleviate the domain discrepancy, we aim to shift the feature distributions in different domains as close as possible for each modality. In our task, the feature distribution consists of two folds: intra- and  inter-sample distributions\footnote{A sample means a video or a query.}. Therefore, we propose two domain alignment loss functions $\mathcal{L}_{DV}$ and $\mathcal{L}_{DQ}$ as follows:
\begin{align}
\mathcal{L}_{DV}=\sum_{\bm{V}_i^s,\bm{V}_{k}^t} MMD(\mu(\bm{V}_{i}^s),\mu(\bm{V}_{k}^t))+ MMD(\sigma(\bm{V}^s),\sigma(\bm{V}^{t})), \nonumber\\
\mathcal{L}_{DQ}=\sum_{\bm{Q}_{j}^s,\bm{Q}_{l}^t} MMD(\mu(\bm{Q}_{j}^s),\mu(\bm{Q}_{l}^t))+ MMD(\sigma(\bm{Q}^{s}),\sigma(\bm{Q}^{t})),\nonumber
\end{align}
where $\mathcal{L}_{DV}$ is the video-based domain alignment loss; $\mathcal{L}_{DQ}$ is the query-based domain alignment loss; $\mu(\cdot)$ and $\sigma(\cdot)$ are the intra-sample distribution function and the inter-sample distribution function respectively, which are defined by:
\begin{align}
\mu(\bm{V}_i^s)=\frac{1}{T_i^s}\sum_{a=1}^{T_i^s}\bm{v}_a^s, \sigma(\bm{V}^s)=\sqrt{\frac{1}{I^s}\sum_{i=1}^{I^s}(\mu(\bm{V}_i^s)-\frac{\sum_{i=1}^{I^s}\mu(\bm{V}_i^s)}{I^s})^2},\nonumber\\
\mu(\bm{Q}_j^s)=\frac{1}{N_j^s}\sum_{b=1}^{N_j^s}\bm{q}_b^s, \sigma(\bm{Q}^s)=\sqrt{\frac{1}{J^s}\sum_{j=1}^{J^s}(\mu(\bm{Q}_j^s)-\frac{\sum_{j=1}^{J^s}\mu(\bm{Q}_j^s)}{J^s})^2},\nonumber\\
\mu(\bm{V}_k^t)=\frac{1}{T_k^t}\sum_{c=1}^{T_k^t}\bm{v}_c^t, \sigma(\bm{V}^t)=\sqrt{\frac{1}{I^t}\sum_{k=1}^{I^t}(\mu(\bm{V}_k^t)-\frac{\sum_{k=1}^{I^t}\mu(\bm{V}_k^t)}{I^t})^2},\nonumber\\
\mu(\bm{Q}_l^t)=\frac{1}{N_l^t}\sum_{d=1}^{N_l^t}\bm{q}_d^t, \sigma(\bm{Q}^t)=\sqrt{\frac{1}{J^t}\sum_{l=1}^{J^t}(\mu(\bm{Q}_l^t)-\frac{\sum_{l=1}^{J^t}\mu(\bm{Q}_l^t)}{J^t})^2},\nonumber
\end{align}
where $\mu(\bm{V}_i^s)$ (or $\mu(\bm{V}_k^t)$) means the intra-video distribution, which is learned by averaging these frame features in the $i$-th source video (or the $k$-th target video); $\sigma(\bm{V}^s)$ (or $\sigma(\bm{V}^t)$) means the inter-video distribution, which is obtained by computing the standard deviation of different video features in the source domain (or the target domain). Similarly, $\mu(\bm{Q}_j^s)$ (or $\mu(\bm{Q}_l^t)$) means the intra-query distribution; $\sigma(\bm{Q}^s)$ (or $\sigma(\bm{Q}^t)$) means the inter-query distribution.

In $\mathcal{L}_{DV}$ and $\mathcal{L}_{DQ}$, $MMD(\cdot)$ is a function  to restrict two sets of variables $(\bm{U},\bm{W})$ matching with each other:
\begin{align}\label{mmd}
MMD(\bm{U},\bm{W})&=\frac{1}{N_{u}^2}\sum_{j=1}^{N_{u}}\sum_{i=1}^{N_{u}}\bm{\phi}(\bm{u}_i,\bm{u}_j)+\frac{1}{N_{w}^2}\sum_{j=1}^{N_{w}}\sum_{i=1}^{N_{w}}\bm{\phi}(\bm{w}_{i},\bm{w}_{j})\nonumber\\
&+\frac{1}{N_{u}N_{w}}\sum_{i=1}^{N_u}\sum_{j=1}^{N_w}\bm{\phi}(\bm{u}_i,\bm{w}_j),
\end{align}
where $\bm{\phi}(\cdot,\cdot)$ means the Gaussian kernel; $\bm{u}_{i}$ (resp. $\bm{w}_j$) is the $i$-th (resp. $j$-th) sample from the set $\bm{U}$ (resp. $\bm{W}$); $N_{u}$ and $N_w$ are the sample numbers of $\bm{U}$ and $\bm{W}$, respectively.

The total domain alignment loss function is as follows:
\begin{align}\label{total_yuduiqi}
\mathcal{L}_{DA}=\mathcal{L}_{DV}+\mathcal{L}_{DQ}.
\end{align}
Thus, by reducing the distance between two distributions based on Eq.~\eqref{mmd}, we align the feature distributions of the same modal in different domains.

\noindent \textbf{Cross-modal alignment.} Besides alleviating the domain discrepancy, we also need to bridge the semantic gap between videos and queries in the target domain, which is the key to the standard VMR task.
Thus, we develop a cross-modal alignment module by mapping videos and queries into a  joint embedding space and aligning the feature distributions of different modalities.

On the one hand, we introduce two projection matrices $\bm{P}_{\bm{V}}\in\mathbb{R}^{d\times d}$ and $\bm{P}_{\bm{Q}}\in\mathbb{R}^{d\times d}$ to project the video and query features into their joint embedding spaces as $\{\bm{V}_p^{t}\}_{{k}}=\{\bm{V}_{f}^{t}\}_{k}\bm{P}_{\bm{V}}$ and $\{\bm{Q}_{p}^{t}\}_{l}=\bm{Q}_{l}^t \bm{P}_{\bm{Q}}$, respectively.
Based on the joint embedding, the cross-modal consistent loss is formulated as:
{\small
\begin{align}\label{motaiduiqi1}
\mathcal{L}_{M1}=&\mathcal{L}_{triplet}(\{\bm{Q}_{p}^{t}\}_{l},\{\bm{V}_p^{t}\}_{l+},\{\bm{V}_p^{t}\}_{l-},m^t)\nonumber\\
&+\mathcal{L}_{triplet}(\{\bm{V}_p^{t}\}_{k},\{\bm{Q}_{p}^{t}\}_{k+},\{\bm{Q}_{p}^{t}\}_{k-},m^t),
\end{align}\normalsize
where $m^t$ is a margin in the target domain, the triplet $(\{\bm{Q}_{p}^{t}\}_{l},\{\bm{V}_p^{t}\}_{l+},\{\bm{V}_p^{t}\}_{l-},m^t)$  are matching video $\{\bm{V}_p^{t}\}_{l+}$ and non-matching video $\{\bm{V}_p^{t}\}_{l-}$ to query $\{\bm{Q}_{p}^{t}\}_{l}$; similarly,  triplet $(\{\bm{V}_p^{t}\}_{k},\{\bm{Q}_{p}^{t}\}_{k+},\{\bm{Q}_{p}^{t}\}_{k-},m^t)$ are video $\{\bm{V}_p^{t}\}_{k}$ with matching query $\{\bm{Q}_p^{t}\}_{k+}$ and non-matching query $\{\bm{Q}_p^{t}\}_{k-}$;
$\mathcal{L}_{triplet}$ is a triplet loss defined as follows:}
\begin{align}\label{triplet}
\mathcal{L}_{triplet}(\bm{A},\bm{P},\bm{N},m)=&\sum_{\bm{A},\bm{P}}\sum_{\bm{N}}\max(m+||\bm{C}(\bm{A},\bm{P})||_2\nonumber\\
&-||\bm{C}(\bm{A},\bm{N})||_2),
\end{align}
where $||\cdot||_2$ means L2-norm,  $\bm{A}$ is an anchor input, $\bm{P}$ is a positive input matching $\bm{A}$, $\bm{N}$ is a negative input that does not match $\bm{A}$,
$m$ is a certain margin between positive pair ($\bm{A},\bm{P}$)
and negative pair ($\bm{A},\bm{N}$).
All these matching/non-matching samples are selected by pairwise ranking and the hardest negative mining strategies \cite{wang2021robust}. By minimizing Eq.~\eqref{motaiduiqi1}, the similarity between a video feature and the corresponding query feature is maximized, and the similarity between all other non-matching ones is minimized, which aligns the textual semantics and the matching visual semantics in the target domain.

On the other hand, we align the feature
distributions of videos and queries. Based on the intra-sample distribution function $\mu(\cdot)$ and the inter-sample distribution function $\sigma(\cdot)$, we develop the following cross-modal feature distribution loss:
\begin{align}\label{motaiduiqi2}
\mathcal{L}_{M2}=\sum_{\bm{V}_{k}^{t},\bm{Q}_{l}^{t}}||\mu(\bm{V}_{k}^{t})-\mu(\bm{Q}_{l}^{t})||_2^2+||\sigma(\bm{V}^{t})-\sigma(\bm{Q}^{t})||_2^2.
\end{align}
Based on Eq.~\eqref{motaiduiqi2}, we can make the feature distributions of queries and matching videos as close as possible.

Combining Eq.~\eqref{motaiduiqi1} and Eq.~\eqref{motaiduiqi2}, we can bridge the semantic gap between videos and queries by the following cross-modal alignment loss:
\begin{align}\label{motaiduiqi}
\mathcal{L}_{MA}=\mathcal{L}_{M1}+ \mathcal{L}_{M2}.
\end{align}

\noindent \textbf{Specific alignment.}
Due to the lack of annotations in the target domain, the learned frame-wise representations might be not optimally discriminative. To avoid some frames near the target moment boundary  being localized in  incorrect boundaries, we develop a specific alignment module to learn more fine-grained and representative features of each specific frame by correlating to the most relevant query information.

Given a set of specific frame features $\{\bm{v}_c^t\}_{c=1}^{T_k^t}$ and a set of query features $\{\bm{Q}_{l}^t\}_{l=1}^{N_l^t}$, we first calculate their cosine similarity $\bm{C}^t(\bm{v}_c^t,\bm{Q}_{l}^t)$ as follows:
\begin{align}\label{xiangsidu_spe}
\bm{C}^t(\bm{v}_c^t,\bm{Q}_{l}^t)=\frac{||\bm{v}_c^{t\top}\bm{Q}_{l}^{t}||_2}{||\bm{v}_{c}^{t}||_2||\bm{Q}_{l}^{t}||_2}.
\end{align}
To obtain the probability that  $\bm{v}_c^t$ corresponds to $\bm{Q}_{l}^t$,  we regularize the similarity by the Softmax function:
\begin{align}\label{bilihua}
\bm{s}_{cl}^t&=Softmax(\bm{C}^t(\bm{v}_c^t,\bm{Q}_{l}^t)).
\end{align}
Then, we choose the query with the greatest probability as the correct query corresponding to $\bm{v}_c^t$. Finally, we employ the specific alignment loss as follows:
\begin{align}\label{momentloss}
\mathcal{L}_{SA}=-\sum_{c}\log(\max(\bm{s}_{cl}^t)|_{l=1}^{N_l^t}).
\end{align}
By minimizing Eq.~\eqref{momentloss}, we can maximize the similarity between the specific frame $\bm{v}_c^t$ and the corresponding query $\bm{Q}_{l}^t$. In this way, we can enhance the characteristics of the specific frame by maximizing its probability belonging to the corresponding query.

\subsection{Training and Testing}
\noindent \textbf{Pre-training for source domain.}
Since we have adequate annotations in the source domain, we first train our MMCDA network in the source domain under a fully-supervised setting.
Given $\bm{V}_0^{s}$ and $\bm{Q}^s$, we introduce two projection matrices $\bm{P}_v\in\mathbb{R}^{d\times T^s}$ and $\bm{P}_q\in\mathbb{R}^{d\times N^s}$ to project the video and query features into the joint semantic representation. On the joint representation, the projection for the video feature is derived as $\bm{V}_p^{s}=\bm{P}_v\bm{V}_0^{s}$ and the projection for the query feature is derived as $\bm{Q}_p^{s}=\bm{P}_q\bm{Q}^s$.
Similar to the joint embedding in the cross-domain alignment (Eq.~\eqref{motaiduiqi1}), we utilize the
pairwise ranking loss with the hardest negative samples as follows:
\begin{align}\label{jianduiloss}
\mathcal{L}_{SL}=&\mathcal{L}_{triplet}(\{\bm{Q}_{p}^{s}\}_{j},\{\bm{V}_p^{s}\}_{j+},\{\bm{V}_p^{s}\}_{j-},m^s)\nonumber\\
&+\mathcal{L}_{triplet}(\{\bm{V}_p^{s}\}_{i},\{\bm{Q}_{p}^{s}\}_{i+},\{\bm{Q}_{p}^{s}\}_{i-},m^s),
\end{align}
where $m^s$ is the certain margin in the source domain.
Eq.~\eqref{jianduiloss} maximizes the similarity between an embedded video feature and the corresponding query feature, and minimizes similarity to all other non-matching ones.
In Eq.~\eqref{jianduiloss}, the first term ensures that given a query input, the matching video should have a larger similarity than the non-matching videos. As for its second term, when we have a video input, the matching query inputs should have a larger similarity than the non-matching query inputs.

\begin{figure}[t]
\centering
\includegraphics[width=0.5\textwidth]{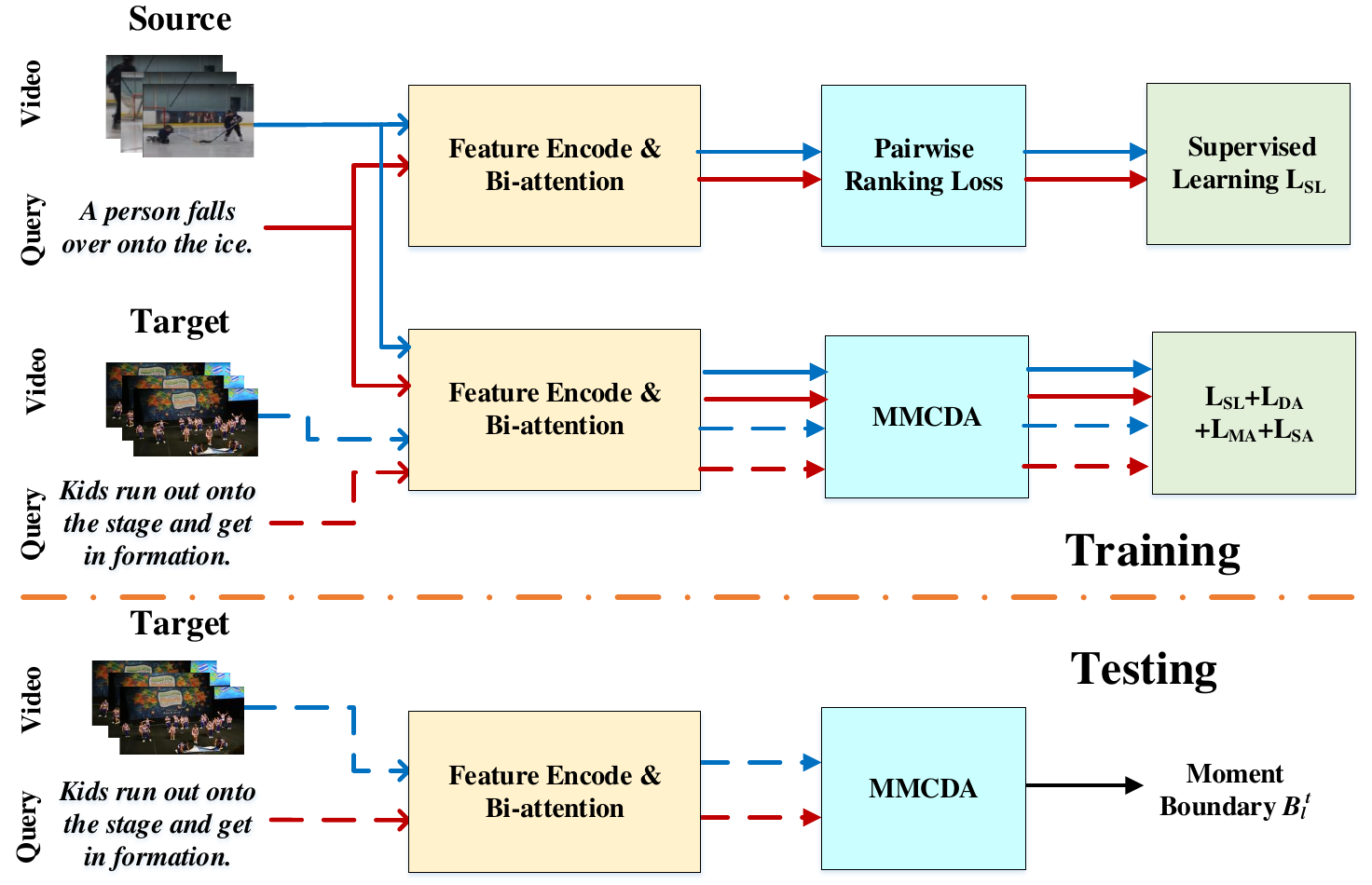}
\vspace{-25pt}
\caption{The training and testing process of our MMCDA.}
 \vspace{-0.38cm}
\label{fig:train_test}
\end{figure}

\noindent \textbf{Main training.}
For the target domain, we do not have the annotated temporal boundaries of each moment-query pair. To transfer the annotation knowledge from the source domain into the target domain, we combine different alignment modules (Eq.~\eqref{total_yuduiqi}, Eq.~\eqref{motaiduiqi}, Eq.~\eqref{momentloss}, and Eq.~\eqref{jianduiloss}) into a multi-task loss function (\textit{i.e.}, the final loss function $L_{final}$) as follows:
\begin{align}\label{source_loss_zong}
\mathcal{L}_{final}=\mathcal{L}_{SL}+\gamma_1\mathcal{L}_{DA}+\gamma_2\mathcal{L}_{MA}+\gamma_3\mathcal{L}_{SA},
\end{align}
where $\gamma_1$, $\gamma_2$, and $\gamma_3$ are parameters to balance the importance between different modules.

Note that our training procedure is divided into two stages shown in Fig. \ref{fig:train_test}. At the first stage, we only use $\mathcal{L}_{SL}$ for pre-training in the source domain. At the second stage, we leverage the final objective function $\mathcal{L}_{final}$ for main training in the source and target domains. The procedure not only effectively gains the annotation knowledge from the source domain, but also transfers the knowledge to the target domain. We can learn the domain-invariant representation \cite{Zhong0LL019} by the domain alignment (Eq.~\eqref{total_yuduiqi}) and obtain the semantic aligned cross-modal representations by the cross-modal alignment (Eq.~\eqref{motaiduiqi}) and the semantic alignment (Eq.~\eqref{momentloss}). Thus, based on Eq.~\eqref{source_loss_zong}, we can learn domain-invariant and semantic aligned cross-modal representations to improve the generalization of our network.




\noindent \textbf{Inference.}
During the inference, we first feed each video-query pair of the target domain into our model to extract its domain-invariant video/query features.
After obtaining both representative video and query features, we follow \cite{MithunPR19,liu2022unsupervised} to compute the cosine similarities between frame-wise video features and the query feature to determine whether each frame is semantic-related to the query.
Finally, we locate the frame with the highest similarity score as the basic predicted moment, and add the left/right frames into the moment if the ratio of their scores to the frame score of the closest moment boundary is less than a certain threshold.
In our all experiments, this inference threshold is set to 0.8 in ActivityNet Captions, TACoS and 0.9 in Charades-STA.
We repeat this step to construct the moment until no frame can be added.

\section{Experimental Results}\label{section:exp}
\begin{table*}[t!]
\centering
\caption{Statistics of the datasets, where L(video) is average length of videos in seconds, L(query) is average number of words in sentence query, L(moment) is average length of moments in seconds, and S(moment) is the standard deviation of moment length in seconds.}
\vspace{-10pt}
\setlength{\tabcolsep}{2mm}{
\begin{tabular}{lccccccccccccccccccccc}
\hline
Dataset  & Domain & \# Videos  & \# Annotations & L(video) & L(query) & L(moment) & S(moment)\\\hline
ActivityNet Captions & Open& 14,926 & 54,926  &117.61s & 14.78 & 36.18s & 40.18s\\\hline
Charades-STA& Indoors& 6,672 & 16,128  & 30.59s& 7.22& 8.22s& 3.59s\\\hline
TACoS& Cooking& 127 & 18,818 & 287.14s & 10.05 & 5.45s & 7.56s\\\hline
\end{tabular}}
\label{dataset}
\end{table*}

\subsection{Datasets and Evaluation Metrics} \label{section:data}
We conduct experiments on three challenging benchmark datasets: TACoS~\cite{RegneriRWTSP13}, Charades-STA~\cite{GaoSYN17}, and ActivityNet Captions \cite{KrishnaHRFN17}, summarized in Table~\ref{dataset}.

\noindent \textbf{TACoS.} TACoS~\cite{RegneriRWTSP13} contains 127 videos, and the average video length is about 7 minutes.
These videos are mainly about cooking scenarios, thus lacking diversity.
The cooking activities appear in the same kitchen scene with some slightly varied cooking objects.
For a fair comparison, we follow the same splitting protocol as \cite{GaoSYN17}.

\noindent \textbf{Charades-STA.} Introduced for the video moment retrieval task, Charades-STA~\cite{GaoSYN17,SigurdssonVWFLG16} is mainly about daily indoor activities with 9,848 videos and 16,128 video-query pairs. Videos in the Charades dataset are
mainly about indoor everyday activities.
The average video length is about 0.5 minutes. The annotations are generated by query decomposition
and keywords matching with a manual check.
As the original Charades dataset has only temporal activity localization and video-level paragraph
description annotations, labels for video moment retrieval are added in Charades-STA later.

\noindent \textbf{ActivityNet Captions.} ActivityNet Captions \cite{KrishnaHRFN17} is a large-scale dataset from open activities including 20,000 untrimmed videos with 100,000 queries from YouTube. These videos in the ActivityNet Captions dataset are open-domain
and much longer with an average duration of about 2 minutes and queries are about 13.5 words.

Note that in our experiments, we use one dataset as the source domain and the rest datasets as the target domain.
We abbreviate ``TACoS'', ``Charades-STA'', and ``ActivityNet Captions'' as ``T'', ``C'', and ``A'' respectively. ``A$\rightarrow$T'' means that we transfer the annotation knowledge from the ActivityNet Captions dataset (source domain) to the TACoS dataset (target domain).
We consider six transfer tasks:
A$\rightarrow$T, C$\rightarrow$T, A$\rightarrow$C, T$\rightarrow$C,  T$\rightarrow$A, and C$\rightarrow$A. Meanwhile, we design three fully-supervised single-domain tasks: A$\rightarrow$A, C$\rightarrow$C, and T$\rightarrow$T. In each single-domain task, for our MMCDA, we only perform pre-training (rather than main training) on a dataset and testing on the same dataset.
%
%
%


Following \cite{ZhangSJZ20}, we use the metric ``R@$n$, IoU=$m$'' for evaluation. ``R@$n$, IoU=$m$'' calculates the ratio of at least one of top-$n$ selected moments having an intersection over union (IoU) larger than $m$. The larger metric means the better performance. In our experiments, we use $n \in \{1,5\}$ for all datasets, $m \in \{0.3, 0.5,0.7\}$ for ActivityNet Captions and Charades-STA, $m \in \{0.1, 0.3,0.5\}$ for TACoS.

\subsection{Implementation Details} \label{section:shiyanxijie}
For video encoding, following \cite{LiuQDZ20}, we resize every frame into $112\times 112$
pixels shape to encode videos, and then utilize a pre-trained C3D model \cite{TranBFTP15} to extract frame-wise features for TACoS, ActivityNet Captions, and Charades-STA datasets. The length of each video is set to 200 in TACoS and ActivityNet Captions and 64 in Charades-STA.
For query encoding, each word is embedded in a 300-dimensional vector with the Glove embedding \cite{PenningtonSM14}.
The hidden state dimension of the Bi-GRU network is set to 512. The hyperparameters $\gamma_1$, $\gamma_2$, $\gamma_3$ are set to 1.0, 0.5, 0.2 respectively according to empirical study. Besides, the batch size, $m^s$, and $m^t$ are set to 128, 0.2, and 0.2, respectively.
We train our whole MMCDA model for 100 epochs with batch size of 16 and early stopping strategy.
Parameter optimization is performed by Adam optimizer with leaning rate $4\times 10^{-4}$ for ActivityNet Captions and Charades-STA and $3\times 10^{-4}$ for TACoS, and linear decay of learning rate and gradient clipping of 1.0.

\subsection{Baseline Comparison} \label{section:compared}
To evaluate the effectiveness of our proposed MMCDA, we compare it to several state-of-the-art VMR models, which are divided into two categories by the viewpoints of fully-supervised (FS) and weakly-supervised (WS) models. (i) FS models: 2D-TAN \cite{ZhangPFL20}, ABLR \cite{yuan2019find}, ACRN \cite{LiuWN0CC18}, ACL-K \cite{ge2019mac}, CBP \cite{Yu0SYLBB18}, {CLEAR \cite{hu2021coarse}}, CMIN \cite{ZhangLZX19}, CSMGAN \cite{LiuQLDZX20}, CTRL \cite{GaoSYN17}, GDP \cite{ChoMGBBSB14},  SAP \cite{ChenJ19a}, SCDM \cite{YuanMWL019}, SM-RL \cite{WangHW19}, {MABAN \cite{sun2021maban}}, MCN  \cite{hendricks2018localizing},
TGN \cite{ChenKN17}, VSLNet \cite{ZhangSJZ20}. (ii) WS models: CTF \cite{ChenMLW19}, LoGAN \cite{2021LoGAN}, RTBPN \cite{ZhangLZZH20}, SCN \cite{LinZZWL20}, TGA \cite{MithunPR19}, MARN \cite{song2020weakly}, VGN \cite{zhang2020counterfactual}, VLANet \cite{MaYKLKY20}.

Following \cite{zhang2021natural,munro2020multi}, we directly cite the results of compared VMR methods from corresponding works, which are trained and tested on the same datasets in FS or WS settings. Similarly, ``CD'' means ``cross-domain''. Note that no weakly-supervised  method reports its results on TACoS. The best results are \textbf{bold}.

\subsection{Does Our Domain Adaption Strategy Work?} \label{section:Plug}
To fairly compare with these single-domain baseline models, we serve our proposed MMCDA as a plug-and-play module for several state-of-the-art models (2D-TAN \cite{ZhangPFL20}, CSMGAN \cite{LiuQLDZX20}, and CBLN \cite{Liu_2021_CVPR}) for investigating its effectiveness.
Specifically,  we develop the following three settings in Table~\ref{shiyanjieguo_jichajiyong}:

\textbf{1) Baseline (BL)}: we directly report the performance of these baseline models. This is under a singe-domain setting, where their training and testing are on the same source dataset.

\textbf{2) Source Only (SO)}: we only use the annotated source dataset to train a baseline model, and directly test the trained model on another unannotated target dataset. This is under a simple cross-domain setting.

\textbf{3) Ours}: we add our multi-modal cross-domain alignment loss ($\gamma_1\mathcal{L}_{DA}+\gamma_2\mathcal{L}_{MA}+\gamma_3\mathcal{L}_{SA}$) to 2D-TAN, CSMGAN, and CBLN, which can make them adaptive to our transfer tasks. This is under our cross-domain setting.

From Table~\ref{shiyanjieguo_jichajiyong}, we conclude the following observations:
\vspace{-2pt}
\begin{itemize}
    \item Directly testing the trained SO model on the target dataset will severely degenerate the retrieval performance, since there are a large domain discrepancy between two datasets and a huge semantic gap between videos and queries in the target dataset.
    \item Our domain adaption strategy is effective for these single-domain models, outperforming these baselines and their SO variants by a large margin.
\end{itemize}
 It shows that our multi-modal cross-domain alignment can bring large improvement on single-domain methods.

\subsection{Performance Comparison and Analysis}  \label{subsection:shiyanjieguo}

\begin{table*}[t!]
\centering
\caption{Our proposed MMCDA serves as a plug-and-play module for several state-of-the-art models on different transfer tasks.}
\vspace{-10pt}
\setlength{\tabcolsep}{1.0mm}{
\begin{tabular}{c|c|cccc|cccc|ccccccccccc}
\hline
\multirow{3}*{Model}&\multirow{3}*{Setting} & \multicolumn{4}{c|}{T$\rightarrow$A}& \multicolumn{4}{c|}{A$\rightarrow$C}  & \multicolumn{4}{c}{C$\rightarrow$T}  \\\cline{3-14}
~&~ & R@1  & R@1 & R@5  & R@5& R@1  & R@1 & R@5  & R@5& R@1  & R@1 & R@5 & R@5 \\
~&~   &  IoU=0.3 & IoU=0.5  & IoU=0.3 & IoU=0.5&  IoU=0.3 & IoU=0.5  & IoU=0.3 & IoU=0.5 & IoU=0.3 & IoU=0.5  & IoU=0.3 & IoU=0.5 \\\hline
\multirow{3}*{2D-TAN \cite{ZhangPFL20}}&BL&37.29&25.32&57.81&45.04&59.45&44.51&77.13&61.96&53.25&39.81&85.15&79.33\\
~&SO  & 32.68 & 20.10  & 52.46 & 42.12&53.51&40.72&72.03&55.12&40.76&36.34&78.73&73.16 \\
~&\textbf{Ours} & \textbf{41.47} & \textbf{28.83}  & \textbf{60.37} & \textbf{49.53}& \textbf{71.92}& \textbf{52.98}& \textbf{98.76}& \textbf{89.52}& \textbf{61.02}& \textbf{46.21}& \textbf{90.47}& \textbf{84.33}\\\hline
\multirow{3}*{CSMGAN \cite{LiuQLDZX20}}&BL&33.90&27.09&53.98&41.22&68.52&49.11&87.68&77.43&78.43&60.04&95.43&89.01\\
~&SO  & 30.07 & 21.90  & 50.12 & 37.81&61.97&43.62&81.95&70.18&73.74&55.93&87.82&79.72 \\
~&\textbf{Ours}  & \textbf{39.41} & \textbf{33.65} &  \textbf{63.18} & \textbf{50.69}& \textbf{72.84}& \textbf{53.86}& \textbf{99.02}& \textbf{89.96}& \textbf{82.26}& \textbf{66.04}& \textbf{97.28}& \textbf{92.16} \\\hline
\multirow{3}*{CBLN \cite{Liu_2021_CVPR}}&BL&38.98&27.65&59.96&46.24&66.34&48.12&88.91&79.32&62.43&47.94&93.56&88.20\\
~&SO&33.06&24.39&53.82&42.63&61.43&43.67&80.72&71.84&55.68&42.39&85.32&80.16\\
~&\textbf{Ours}  &\textbf{43.72}&\textbf{35.74}& \textbf{65.71}&\textbf{49.68}&\textbf{73.66}&\textbf{55.60}&\textbf{99.57}&\textbf{89.91}&\textbf{71.03}&\textbf{51.32}&\textbf{96.08}&\textbf{90.73}\\\hline\hline
\multirow{3}*{Model}&\multirow{3}*{Setting} & \multicolumn{4}{c|}{C$\rightarrow$A}& \multicolumn{4}{c|}{A$\rightarrow$T}  & \multicolumn{4}{c}{T$\rightarrow$C}  \\\cline{3-14}
~&~ & R@1  & R@1 & R@5  & R@5& R@1  & R@1 & R@5  & R@5& R@1  & R@1 & R@5 & R@5 \\
~&~   &  IoU=0.3 & IoU=0.5  & IoU=0.3 & IoU=0.5&  IoU=0.3 & IoU=0.5  & IoU=0.3 & IoU=0.5 & IoU=0.3 & IoU=0.5  & IoU=0.3 & IoU=0.5 \\\hline
\multirow{3}*{2D-TAN \cite{ZhangPFL20}}&BL&53.25& 39.81& 85.15& 79.33& 59.45& 44.51& 77.13& 61.96& 37.29& 25.32& 57.81& 45.04\\
~&SO  &  50.27& 34.82& 76.75& 74.18& 48.26& 37.52& 68.31& 49.35& 30.49& 18.76& 50.38& 40.77\\
~&\textbf{Ours} & \textbf{79.25}& \textbf{62.13}& \textbf{95.83}& \textbf{89.17}& \textbf{63.48}& \textbf{49.76}& \textbf{87.21}& \textbf{78.32}& \textbf{48.74}& \textbf{35.82}& \textbf{86.31}& \textbf{70.06} \\\hline
\multirow{3}*{CSMGAN \cite{LiuQLDZX20}}&BL&78.43& 60.04& 95.43& 89.01& 68.52& 49.11& 87.68& 77.43& 33.90& 27.09& 53.98& 41.22\\
~&SO  & 72.92& 53.48& 86.74& 78.94& 60.83& 44.06& 82.34& 71.03& 31.52& 20.71& 51.34& 38.03 \\
~&\textbf{Ours}  &\textbf{84.38}& \textbf{63.99}& \textbf{97.83}& \textbf{93.72}& \textbf{70.51}& \textbf{53.80}& \textbf{79.23}& \textbf{80.16}& \textbf{49.14}& \textbf{36.52}& \textbf{65.86}& \textbf{71.38}  \\\hline
\multirow{3}*{CBLN \cite{Liu_2021_CVPR}}&BL&62.43& 47.94& 93.56& 88.20& 66.34& 48.12& 88.91& 79.32& 38.98& 27.65& 59.96& 46.24\\
~&SO&54.96& 41.72& 84.05& 80.32& 62.07& 44.01& 79.82& 72.05& 32.94& 23.97& 54.06& 43.01\\
~&\textbf{Ours} &\textbf{85.13}& \textbf{64.75}& \textbf{98.26}& \textbf{94.98}& \textbf{72.98}& \textbf{52.87}& \textbf{89.94}& \textbf{80.96}& \textbf{49.25}& \textbf{38.29}& \textbf{87.32}& \textbf{71.54}\\\hline
\end{tabular}
}
\vspace{-0.25cm}
\label{shiyanjieguo_jichajiyong}
\end{table*}
\noindent \textbf{Comparison on TACoS.}
As shown in Table~\ref{shiyanjieguo_TACOS}, we compare our cross-domain MMCDA with the state-of-the-art fully-supervised and weakly-supervised methods on the TACoS dataset, where MMCDA reaches the highest results in terms of all metrics. {Particularly, compared with the best fully-supervised model CLEAR, MMCDA(A$\rightarrow$T) achieves  1.42\% absolute improvements in ``R@1, IoU=0.5''.} The main reason is that in this dataset, the cooking activities appear in the same kitchen scene with some slightly varied cooking objects. Thus, these baseline models are difficult to localize such fine-grained activities correctly. Different from these single-domain mmodels, MMCDA(A$\rightarrow$T) and MMCDA(C$\rightarrow$T) can learn these fine-grained activity representations from ActivityNet Captions and Charades-STA datasets, respectively. Besides, MMCDA(A$\rightarrow$T) and MMCDA(C$\rightarrow$T) also outperform MMCDA(T$\rightarrow$T) with a clear margin, which shows the effectiveness of our multi-modal cross-domain alignment module in the VMR task.

\noindent \textbf{Comparison on Charades-STA.}
We also compare our cross-domain MMCDA(A$\rightarrow$C) with fully-supervised and weakly-supervised models on the Charades-STA dataset in Table~\ref{shiyanjieguo_Char}, where MMCDA achieves a new state-of-the-art performance on most metrics.
{Compared with the best fully-supervised method MABAN, our proposed MMCDA achieves competitive performance.
Specifically, compared with  MABAN, our MMCDA(A$\rightarrow$C) brings 3.51\% improvements in ``R@1, IoU=0.7''. Note that, our proposed MMCDA is simpler than MABAN, and we can achieve much better performance when we apply our domain adaption strategy to other strong baselines shown in Table \ref{shiyanjieguo_jichajiyong}.} Compared with the best weakly-supervised method LoGAN, our MMCDA(A$\rightarrow$C) brings 21.23\% and 21.91\% improvements in ``R@1, IoU=0.7'' and ``R@5, IoU=0.7'', respectively. It verifies the benefits of transferring the annotation knowledge from the ActivityNet Captions dataset to the Charades-STA dataset through our multi-modal cross-domain alignment module.

\noindent \textbf{Comparison on ActivityNet Captions.}
Table~\ref{shiyanjieguo_act} also reports the  grounding performance on the ActivityNet Captions dataset. Obviously, our MMCDA(C$\rightarrow$A) also performs better than supervised methods in most  cases with significant improvements.
{Particularly, compared with the best fully-supervised approach CLEAR, our MMCDA(C$\rightarrow$A) obtains 10.29\% and 18.14\%  improvements in the strict metrics ``R@1, IoU=0.3'' and ``R@1, IoU=0.7'', respectively. Also, our MMCDA(C$\rightarrow$A) significantly outperforms the second best fully-supervised beseline MABAN. In particular, compared to MABAN, MMCDA(C$\rightarrow$A) improves the performance about 10.26\% and 21.85\% in the strict metrics ``R@1, IoU=0.5'' and ``R@1, IoU=0.7'', respectively.} Compared with the weakly-supervised approach SCN, our MMCDA(C$\rightarrow$A) brings 23.46\% and 23.71\% improvements in ``R@1, IoU=0.5'' and ``R@5, IoU=0.5'', respectively.




\begin{table}[t!]
\centering
\caption{Performance on the TACoS dataset.}
\vspace{-0.32cm}
\setlength{\tabcolsep}{0.2mm}{
\begin{tabular}{c|c|cccccccccccccccccc}
\hline
\multirow{2}*{Model}&\multirow{2}*{Mode}&R@1 & R@1 & R@1 & R@5 & R@5 & R@5\\
~& ~& IoU=0.1 & IoU=0.3 & IoU=0.5 & IoU=0.1 & IoU=0.3 & IoU=0.5 \\\hline
2D-TAN \cite{ZhangPFL20}&FS&47.59& 37.29& 25.32& 70.31& 57.81& 45.04\\
ABLR \cite{yuan2019find}&FS &  34.70& 19.50& 9.40& - &- &-\\
ACRN \cite{LiuWN0CC18}&FS &  24.22& 19.52& 14.62& 47.42& 34.97& 24.88\\
CBP \cite{Yu0SYLBB18}&FS &- &27.31& 24.79& -& 43.64& 37.40\\
CMIN \cite{ZhangLZX19}&FS &  32.48& 24.64& 18.05& 62.13& 38.46& 27.02\\
CSMGAN \cite{LiuQLDZX20}&FS &42.74& 33.90& 27.09& 68.97& 53.98& 41.22\\
GDP \cite{ChoMGBBSB14}&FS  &  39.68& 24.14& 13.50& - &- &-\\
SCDM \cite{YuanMWL019}&FS &-& 26.11& 21.17& -& 40.16& 32.18\\
TGN \cite{ChenKN17}&FS &  41.87& 21.77& 18.90& 53.40& 39.06& 31.02\\
VSLNet \cite{ZhangSJZ20}&FS &-&29.61& 24.27& -&-&-\\
{CLEAR \cite{hu2021coarse}}&{FS}&{-}&{42.18}&{30.27}&{-}&{63.61}&{51.76}\\\hline\hline
MMCDA(T$\rightarrow$T)&FS&36.58&22.63&14.95&54.01&45.77&34.78\\
\textbf{MMCDA(A$\rightarrow$T)}&CD &\textbf{55.38} & \textbf{43.07} & \textbf{31.69} & \textbf{77.54} & \textbf{63.81} & \textbf{51.82}\\
MMCDA(C$\rightarrow$T)&CD &47.97 & 35.41 & 26.88 & 69.53 & 58.10 & 42.62 \\\hline
\end{tabular}
}
\vspace{-0.45cm}
\label{shiyanjieguo_TACOS}
\end{table}
\begin{table}[t!]
\centering
\caption{Performance on the Charades-STA dataset.}
\vspace{-0.32cm}
\setlength{\tabcolsep}{0.2mm}{
\begin{tabular}{c|c|ccccccccccccccccccc}
\hline
\multirow{2}*{Model}&\multirow{2}*{Mode}&R@1 & R@1 & R@1 & R@5 & R@5 & R@5\\
~&~&  IoU=0.3 & IoU=0.5 & IoU=0.7 & IoU=0.3 & IoU=0.5 & IoU=0.7 \\\hline
2D-TAN \cite{ZhangPFL20}&FS &-& 39.81& 23.25& -& 79.33& 52.15\\
ACL-K \cite{ge2019mac}&FS &-& 30.48& 12.20&-&  64.84&35.13\\
SAP \cite{ChenJ19a}&FS&-& 27.42& 13.36&-& 66.37& 38.15\\
SM-RL \cite{WangHW19} &FS&-& 24.36& 11.17&-& 61.25& 32.08\\
VSLNet \cite{ZhangSJZ20}&FS&64.30& 47.31& 30.19& -&-& -\\
{MABAN \cite{sun2021maban}}&{FS}&{-}&{\textbf{56.29}}&{32.26}&{-}&{-}&{-}\\\hline
LoGAN \cite{2021LoGAN}&WS& 51.67& 34.68& 14.54& 92.74& 74.30& 39.11\\
 RTBPN \cite{ZhangLZZH20}&WS& 60.04& 32.36& 13.24& 97.48& 71.85& 41.18\\
 SCN \cite{LinZZWL20}&WS& 42.96& 23.58& 9.97& 95.56& 71.80& 38.87\\
  TGA \cite{MithunPR19}&WS& 29.68& 17.04& 6.93& 83.87& 58.17& 26.80\\
 VLANet \cite{MaYKLKY20}&WS&45.24& 31.83& 14.17& 95.70& 82.85& 33.09\\\hline\hline
 MMCDA(C$\rightarrow$C)& FS&39.42&24.53&10.04&75.21&53.52&30.87 \\
\textbf{MMCDA(A$\rightarrow$C)} &CD& \textbf{71.58} & 54.80 & \textbf{35.77} & \textbf{98.36} & \textbf{88.90} & \textbf{61.02}\\
MMCDA(T$\rightarrow$C) &CD& 48.69 & 35.17 & 17.28 & 85.40 & 69.93 & 39.74 \\\hline
\end{tabular}
}
\vspace{-0.35cm}
\label{shiyanjieguo_Char}
\end{table}

\noindent \textbf{Analysis.} From the above results, we can observe that:
{(i) In all the datasets, our proposed MMCDA on these cross-domain settings (A$\rightarrow$T, C$\rightarrow$T, C$\rightarrow$A, T$\rightarrow$A, A$\rightarrow$C, T$\rightarrow$C) performs better than these fully-supervised settings (T$\rightarrow$T, A$\rightarrow$A, C$\rightarrow$C). There are two reasons as follows: Firstly, our proposed MMCDA on the fully-supervised setting is a simple model (only with $\mathcal{L}_{SL}$), which is only used as a baseline approach. Secondly, we focus on the cross-domain setting in this paper. The better performance on our cross-domain setting shows the effectiveness of our MMCDA network.
(ii) Although our MMCDA is under the cross-domain setting, it can significantly outperforms  fully-supervised and weakly-supervised VMR methods.} It is because MMCDA can effectively transfer the annotation knowledge from the source domain to the target domain. (iii) For different cross-domain VMR tasks (same target dataset and different source datasets), MMCDA performs differently.
For example, MMCDA performs better in the A$\rightarrow$C task than in the T$\rightarrow$C task. One possible reason is that
the Activity Captions dataset contains multiple and sufficient scenarios that are able to generalize  MMCDA to other datasets. In the contrast, the TACoS dataset only contains cooking videos (monotonous and simple) with a small number of videos, which may limit the generalization ability of our MMCDA.

\begin{table}[t!]
\centering
\caption{Performance on the ActivityNet Captions dataset.}
\vspace{-0.32cm}
\setlength{\tabcolsep}{0.2mm}{
\begin{tabular}{c|c|cccccccccccccccccc}
\hline
\multirow{2}*{Model}&\multirow{2}*{Mode}&R@1 & R@1 & R@1 & R@5 & R@5 & R@5\\
~  &~& IoU=0.3 & IoU=0.5 & IoU=0.7 & IoU=0.3 & IoU=0.5 & IoU=0.7 \\\hline
ACRN \cite{LiuWN0CC18}&FS& 49.70 & 31.67 & 11.25 & 76.50 & 60.34 & 38.57\\
CTRL \cite{GaoSYN17}&FS& 47.43 & 29.01 & 10.34 & 75.32 & 59.17 & 37.54\\
MCN  \cite{hendricks2018localizing}&FS& 39.35& 21.36 & 6.43 & 68.12 & 53.23 & 29.70\\
SCDM \cite{YuanMWL019}&FS& 54.80 & 36.75 & 19.86 & 77.29 & 64.99 & 41.53 \\
TGN \cite{ChenKN17}&FS& 45.51& 28.47 & - & 57.32 & 43.33 & -\\
{MABAN \cite{sun2021maban}}&{FS}&{-}&{42.42}&{24.34}&{-}&{-}&{-}\\
  {CLEAR \cite{hu2021coarse}}& {FS}& {59.96}& {45.33}& {28.05}& {85.83}& {77.26}& {\textbf{62.13}}\\
\hline
 CTF \cite{ChenMLW19}&WS& 44.30& 23.60&-&-&-&-\\
  RTBPN \cite{ZhangLZZH20}&WS& 49.77& 29.63&-& 79.89& 60.56&-\\
 SCN \cite{LinZZWL20}&WS& 47.23& 29.22& -& 71.45& 55.69&-\\
 VGN \cite{zhang2020counterfactual}&WS&46.17&28.79&-&71.23&55.13\\
 MARN \cite{song2020weakly}&WS&47.01& 29.95&-& 72.02& 57.49&-\\\hline\hline
 MMCDA(A$\rightarrow$A)&FS&40.18&16.72&11.73&43.81&36.74&30.52\\
\textbf{MMCDA(C$\rightarrow$A)}&CD& \textbf{70.25} & \textbf{52.68} & \textbf{46.19} & \textbf{88.73} & \textbf{79.40} & 60.17\\
MMCDA(T$\rightarrow$A) &CD& 42.58 & 20.90 & 14.69 & 58.75 & 46.81 & 40.13 \\\hline
\end{tabular}
}
\label{shiyanjieguo_act}
\end{table}
\subsection{Ablation Study}\label{subsection:xiaorongshiyan}
In this section, we will perform in-depth ablation studies to evaluate the effectiveness of each component in our MMCDA on different transfer tasks.
\begin{table*}[t!]
\centering
\caption{{Main ablation study on all the tasks.}}
\vspace{-0.32cm}
\setlength{\tabcolsep}{1mm}{
\begin{tabular}{c|cccc|cccc|cccccccccccc}
\hline
\multirow{3}*{Model} & \multicolumn{4}{c|}{A$\rightarrow$T}&\multicolumn{4}{c|}{T$\rightarrow$C}&\multicolumn{4}{c}{C$\rightarrow$A}\\\cline{2-13}
~& R@1  & R@1 & R@5  & R@5& R@1  & R@1 & R@5  & R@5& R@1  & R@1 & R@5 & R@5\\
~   &  IoU=0.3 & IoU=0.5  & IoU=0.3 & IoU=0.5&  IoU=0.3 & IoU=0.5  & IoU=0.3 & IoU=0.5 & IoU=0.3 & IoU=0.5  & IoU=0.3 & IoU=0.5 \\\hline
MMCDA(w/o DA)&  34.62 & 23.50  & 54.79 & 42.41&39.84&23.76&76.98&58.73&61.96&43.87&75.42&67.49 \\
MMCDA(w/o MA) & 36.61  & 25.64  & 56.30 & 44.19&43.95&23.89&82.91&65.77&64.26&46.82&79.51&70.86 \\
MMCDA(w/o SA) & 38.57  & 27.98  & 58.45 & 46.66 &47.06&30.84&84.05&66.15&68.72&50.78&83.45&77.46 \\\hline
\textbf{MMCDA(full)}&  \textbf{43.07} & \textbf{31.69}  & \textbf{63.81} & \textbf{51.82}& \textbf{48.69}& \textbf{35.17}& \textbf{85.40}& \textbf{69.93}& \textbf{70.25}& \textbf{52.68}& \textbf{88.73}& \textbf{79.40} \\\hline\hline
\multirow{3}*{{Model}}& \multicolumn{4}{c|}{{T$\rightarrow$A}}& \multicolumn{4}{c|}{{C$\rightarrow$T}}  & \multicolumn{4}{c}{{A$\rightarrow$C}}  \\\cline{2-13}
~& {R@1}  & {R@1} & {R@5}  & {R@5}& {R@1}  & {R@1} & {R@5}  & {R@5}& {R@1}  & {R@1} & {R@5} & {R@5}\\
~   &  {IoU=0.3} & {IoU=0.5}  & {IoU=0.3} & {IoU=0.5}&  {IoU=0.3} & {IoU=0.5}  & {IoU=0.3} & {IoU=0.5} & {IoU=0.3} & {IoU=0.5}  & {IoU=0.3} & {IoU=0.5} \\\hline
{MMCDA(w/o DA)}& {38.54}&{16.27}&{52.85}&{40.06}&{31.70}&{20.46}&{51.06}&{36.79}&{67.15}&{46.91}&{94.02}&{82.38}\\
{MMCDA(w/o MA)} &{39.04}&{17.68}&{54.72}&{41.65}&{33.46}&{22.73}&{54.93}&{37.01}&{68.42}&{47.88}&{96.35}&{84.72}\\
{MMCDA(w/o SA)} &{40.35}&{19.79}&{57.21}&{44.92}&{34.87}&{23.96}&{56.49}&{39.20}&{69.19}&{49.96}&{97.83}&{86.54}\\
{\textbf{MMCDA(full)}}& {\textbf{42.58}} & {\textbf{20.90}}  & {\textbf{58.75}} & {\textbf{46.81}}& {\textbf{35.41}} & {\textbf{26.88}} &  {\textbf{58.10}} & {\textbf{42.62}}& {\textbf{71.58}} & {\textbf{54.80}} &  {\textbf{98.36}} & {\textbf{88.90}} \\\hline
\end{tabular}
}
\label{shiyanjieguo_xiaorong}
\end{table*}
\begin{figure*}[t!]
\centering
\subfloat[{A$\rightarrow$C}]{\label{fig:cha_xunliantu}\includegraphics[width=0.27\textwidth]{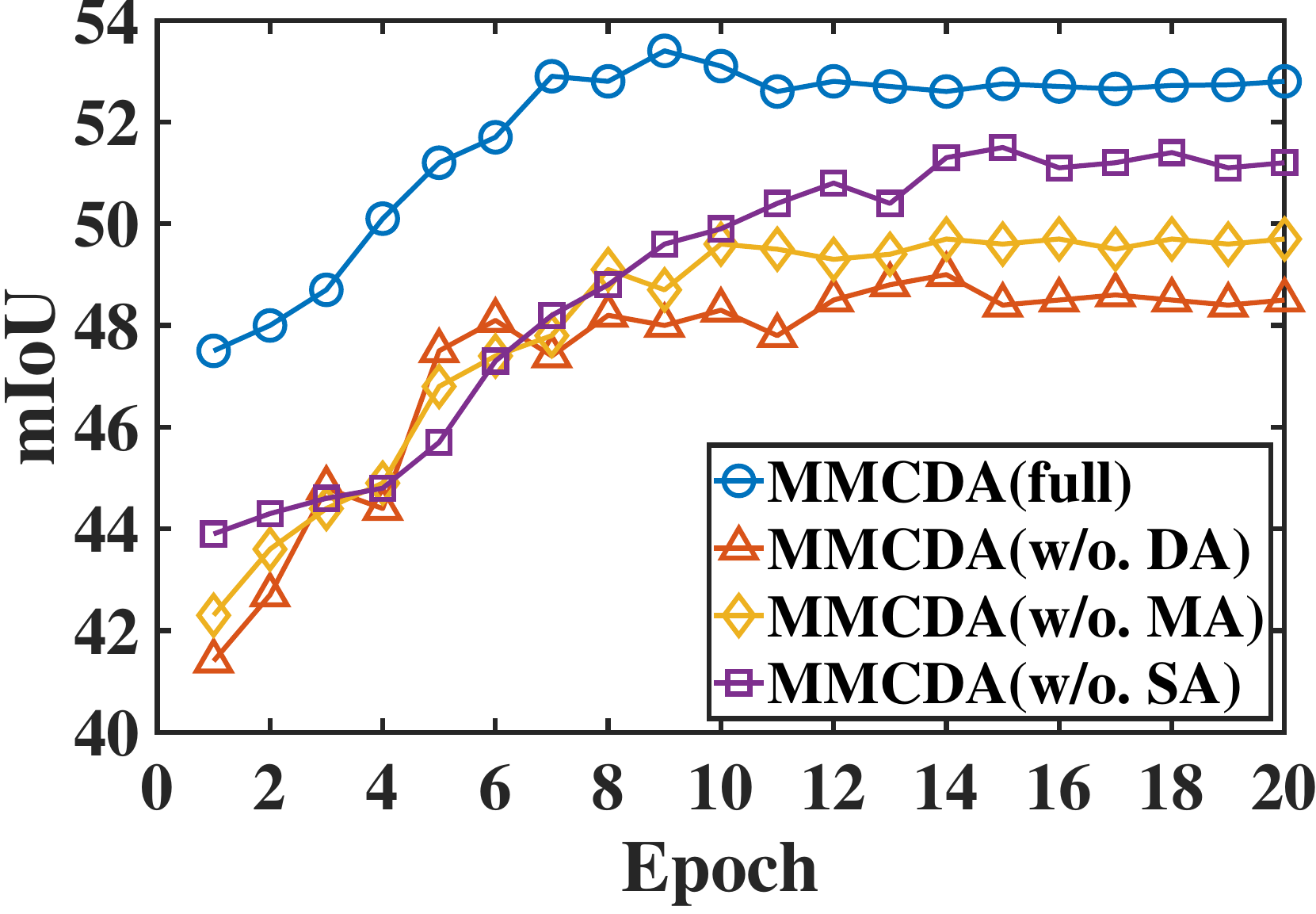} }
\subfloat[{C$\rightarrow$A}]{\label{fig:act_xunliantu}\includegraphics[width=0.27\textwidth]{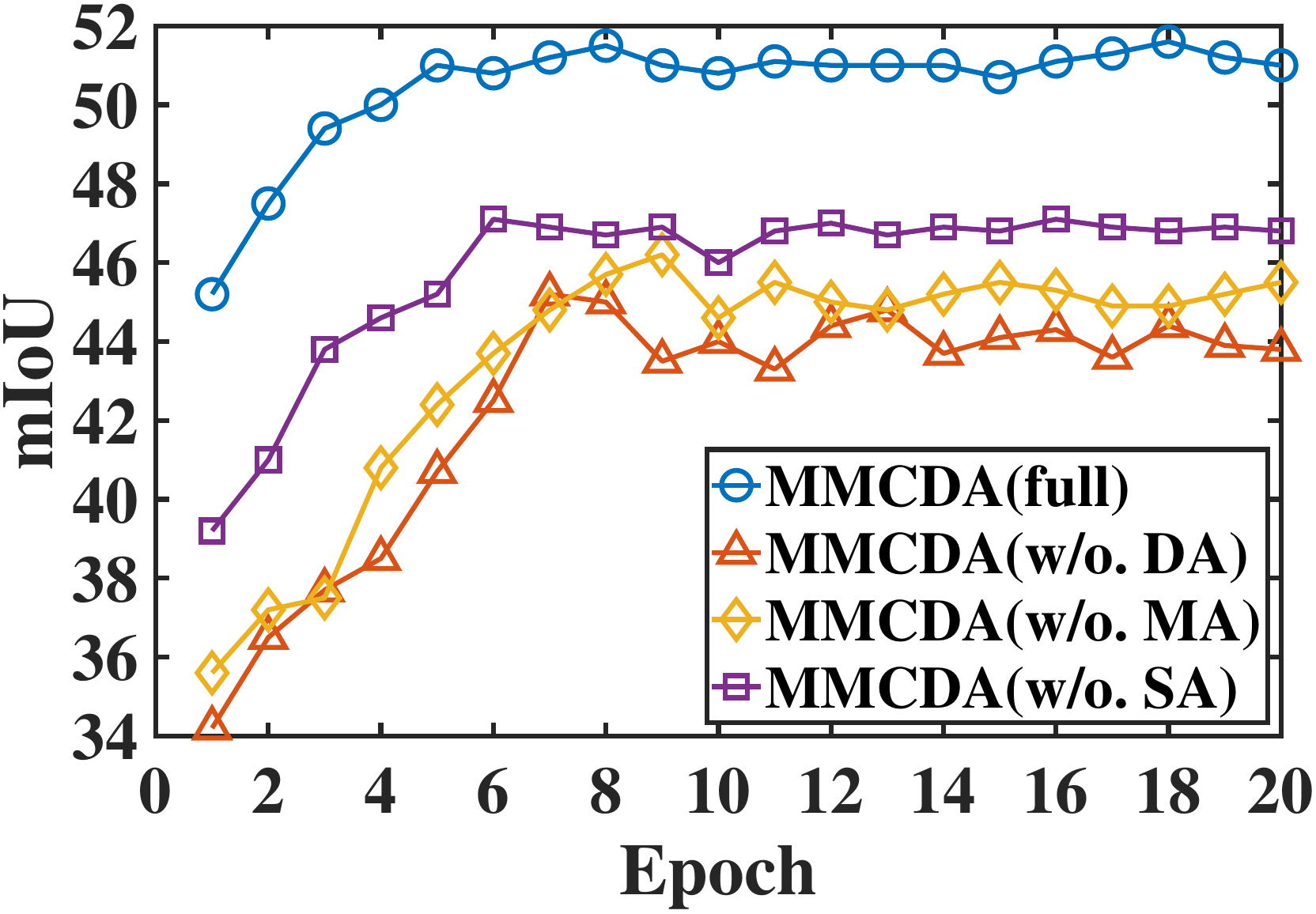} }
\subfloat[{A$\rightarrow$T}]{\label{fig:tacos_xunliantu}\includegraphics[width=0.27\textwidth]{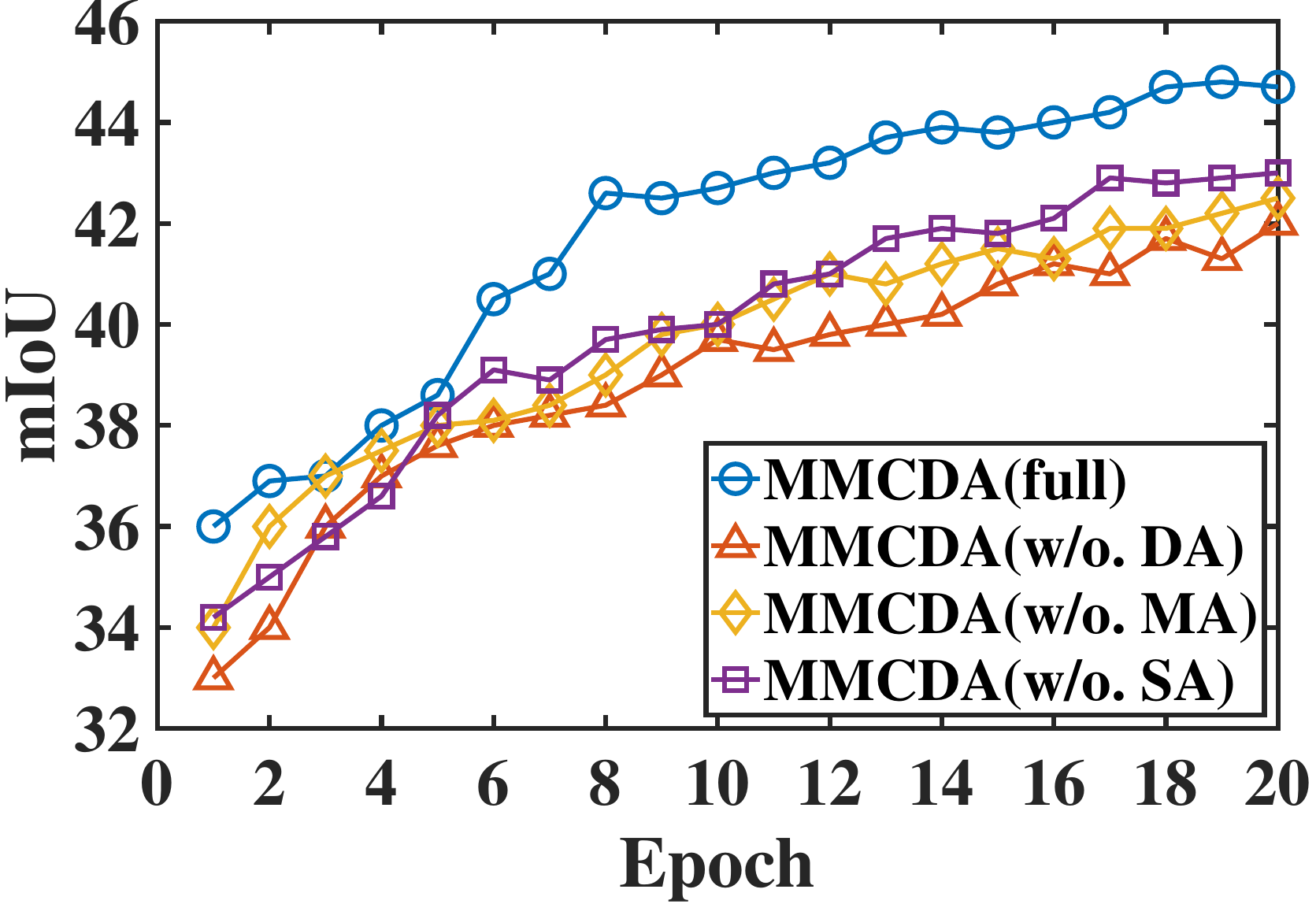} }
\subfloat[{T$\rightarrow$A}]{\label{fig:act_xunliantu_2}\includegraphics[width=0.27\textwidth]{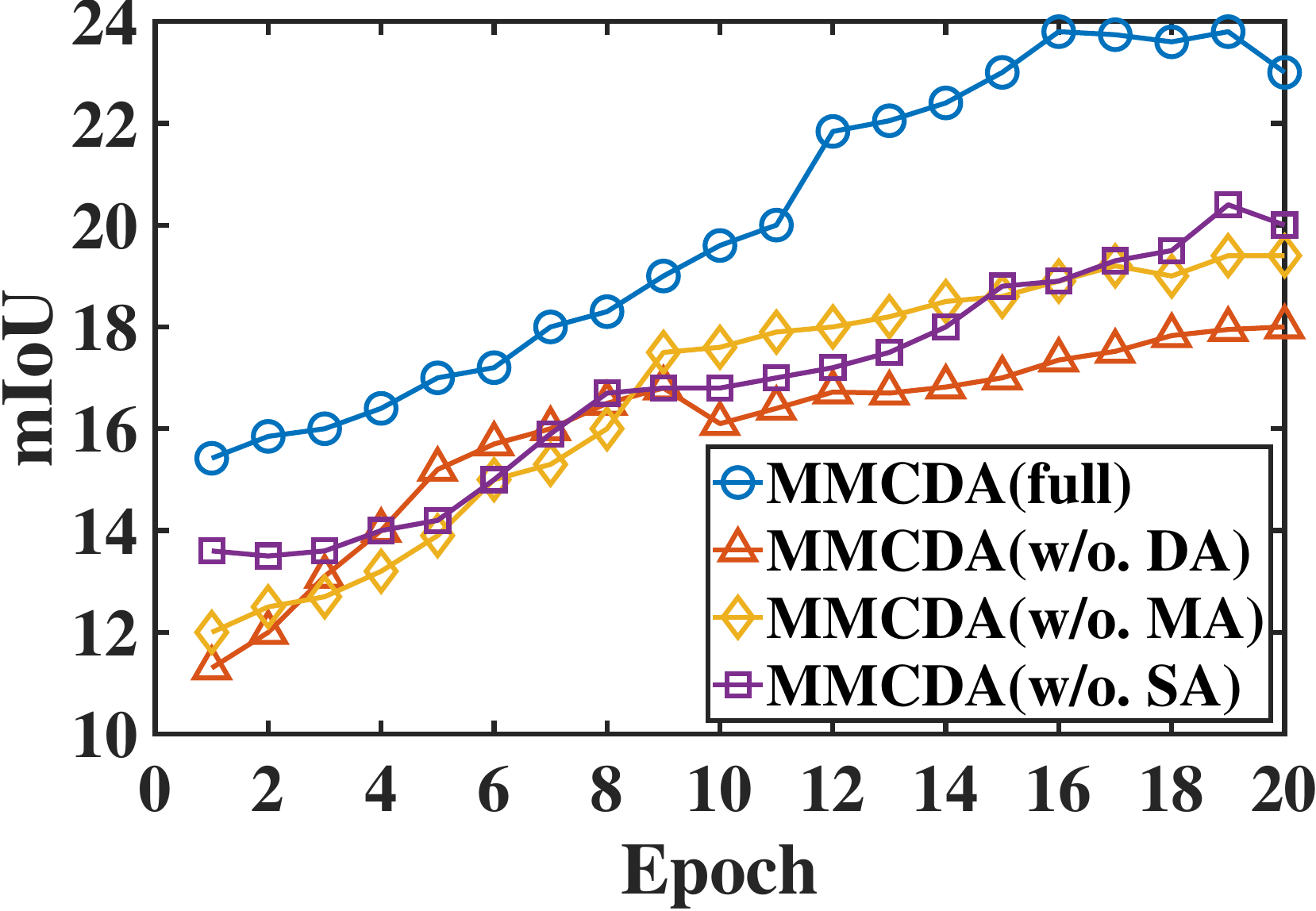} }
\subfloat[{C$\rightarrow$T}]{\label{fig:tacos_xunliantu_2}\includegraphics[width=0.27\textwidth]{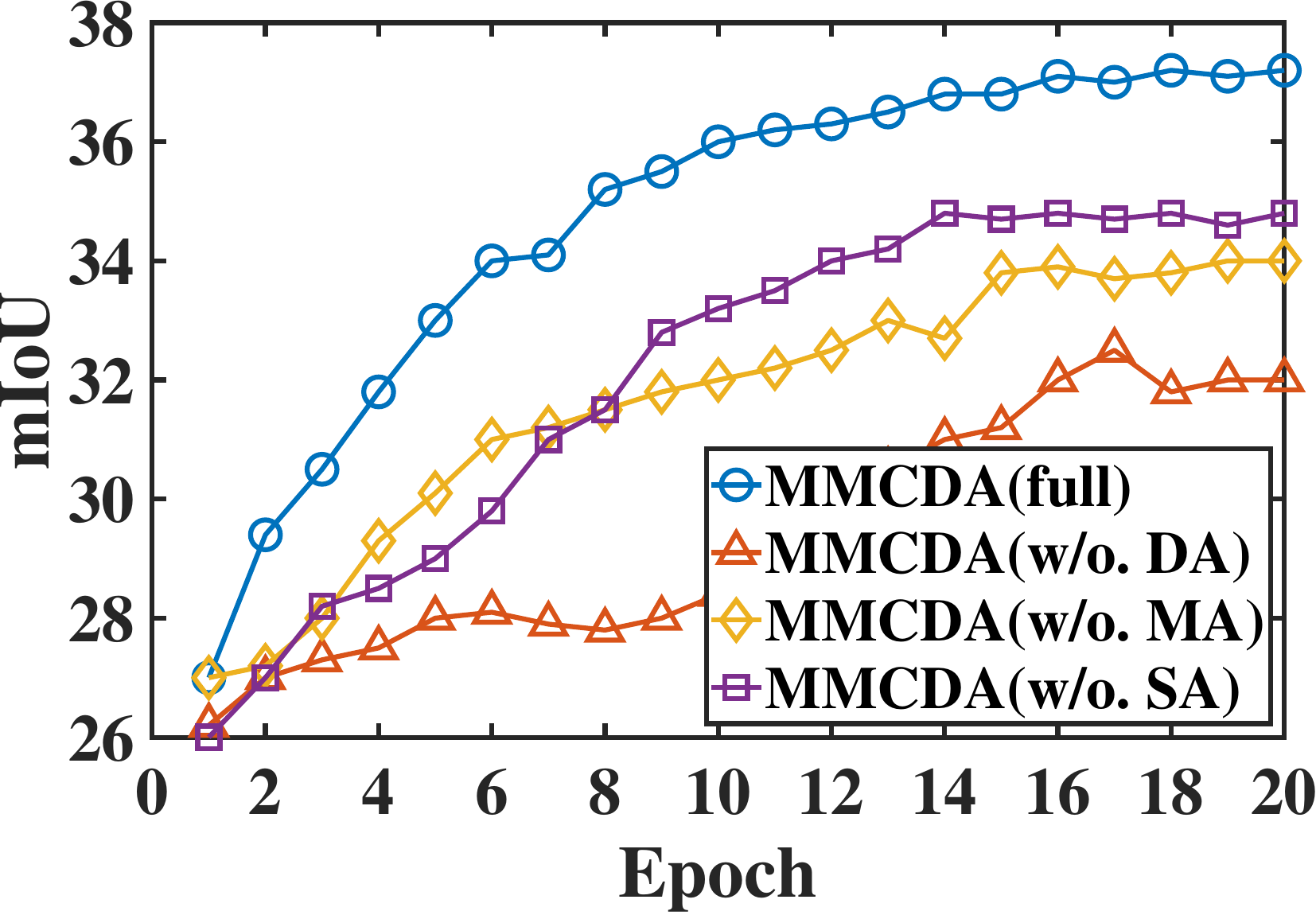} }
\subfloat[{T$\rightarrow$C}]{\label{fig:cha_xunliantu_2}\includegraphics[width=0.27\textwidth]{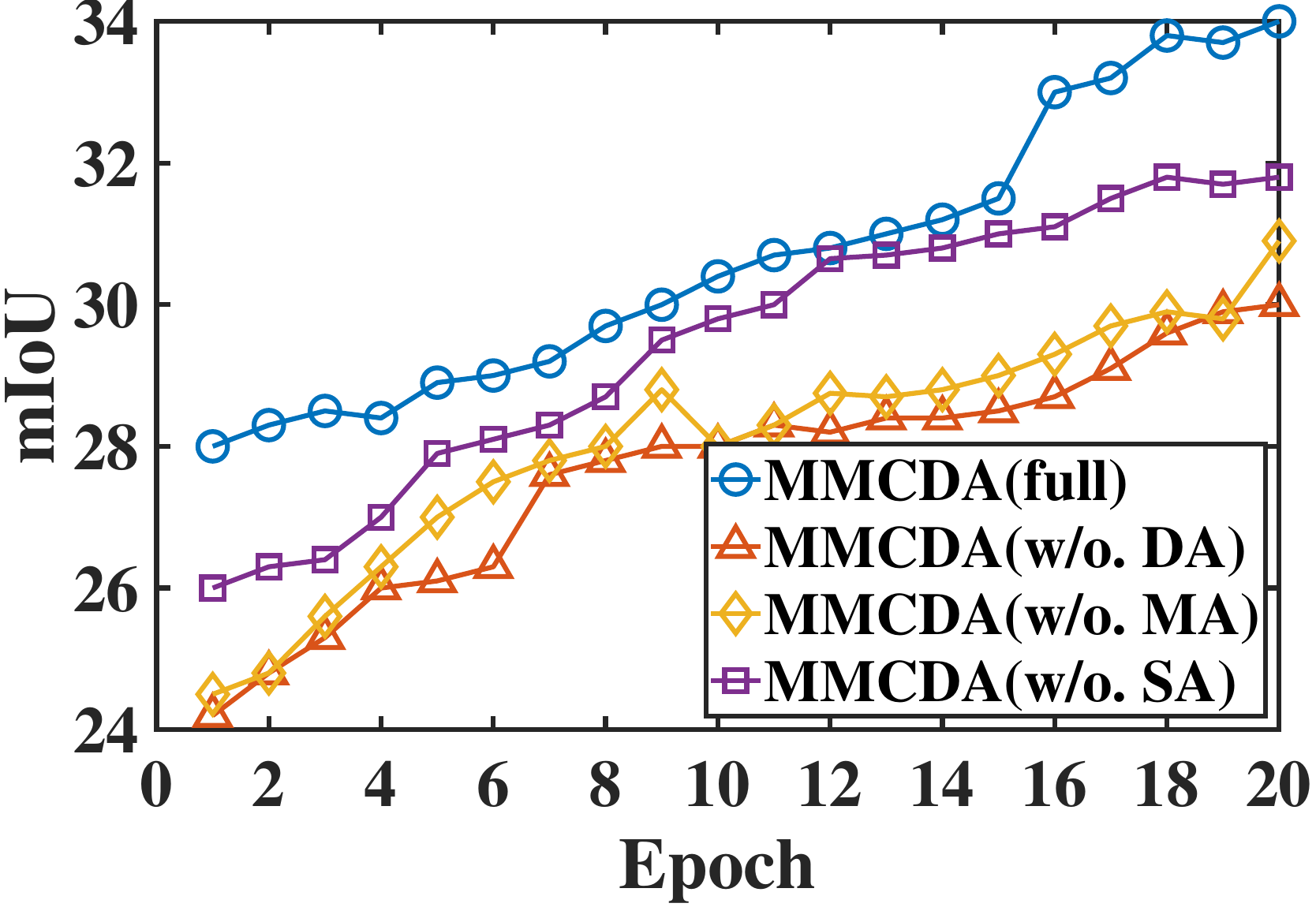} } 
\vspace{-10pt}
\caption{Performance of each ablation module on the target domain during training epochs.}
\label{fig:xunliantu}
\end{figure*}

\noindent \textbf{Main ablation study.}
To systematically evaluate the effectiveness of each module of our proposed MMCDA, {we first conduct an main ablation study on all six transfer tasks.} Table~\ref{shiyanjieguo_xiaorong} shows the corresponding results. In the main ablation study, we design the following ablation models: 1) \textbf{MMCDA(w/o\footnote{In our experiments, ``w/o'' means ``without'', while ``w/'' means ``with''.} DA)}: we remove the domain alignment module (Eq.~\eqref{total_yuduiqi}) from MMCDA;
2) \textbf{MMCDA(w/o MA)}: we remove the cross-modal alignment module (Eq.~\eqref{motaiduiqi}) from MMCDA; 3) \textbf{MMCDA(w/o SA)}: we remove the specific alignment module (Eq.~\eqref{momentloss}) from MMCDA;
4) \textbf{MMCDA(full)}: our full MMCDA model.

From the Table \ref{shiyanjieguo_xiaorong}, we can observe that  MMCDA(full) outperforms other ablation models in all the tasks, which shows the effectiveness of each module. Especially, compared with MMCDA(w/o DA), MMCDA(full) brings 9.11\% improvement in ``R@5, IoU=0.5'' on the A$\rightarrow$T transfer task; compared with MMCDA(w/o MA), MMCDA(full) raises the performance around 7.51\% in ``R@5, IoU=0.3''; compared with MMCDA(w/o SA), MMCDA(full) obtains the performance improvement about 5.41\% in ``R@5, IoU=0.1''. It is because the domain alignment module is able to close the domain discrepancy between different datasets for transferring the annotation knowledge; the cross-modal alignment module can reduce the semantic gaps between videos and queries in each dataset for matching a given query and the corresponding video; the specific alignment module aims to learn more fine-grained and representative frame-wise features for more accurate localization.


Based on the above results, we can obtain the following conclusions: 1) Compared with each model, the full model perform the best on all the datasets, which illustrates the effectiveness of each alignment module of our MMCDA. 2) Almost all ablation models outperform all state-of-the-art methods, which shows that our MMCDA can effectively handle the cross-domain video moment retrieval task. Meanwhile, it also illustrates that the excellent performance of our MMCDA does not rely on one specific key alignment module, and our full model is robust to address this cross-domain task. 3) We can observe that both MMCDA(w/o SA) and MMCDA(w/o MA) outperforms MMCDA(w/o DA). It is because the domain alignment is the key of the cross-domain VMR task to  transfer the knowledge from source domain to target domain.

\begin{table*}[t!]
    \centering
    \caption{Ablation study on the domain alignment and cross-modal alignment on all the tasks.}
    \vspace{-10pt}
    \setlength{\tabcolsep}{1.0mm}{
    \begin{tabular}{c|c|cccccc|ccccccc}
    \hline
    \multirow{3}*{Module} & \multirow{3}*{Changes} & \multicolumn{6}{c|}{A$\rightarrow$C}&\multicolumn{6}{c}{C$\rightarrow$A}\\\cline{3-14}
    &&R@1, & R@1,& R@1, & R@5, & R@5, & R@5, &R@1, & R@1,& R@1, & R@5, & R@5, & R@5, \\
     ~ & ~ & IoU=0.3 &IoU=0.5 & IoU=0.7& IoU=0.3 & IoU=0.5 & IoU=0.7& IoU=0.3 &IoU=0.5 & IoU=0.7& IoU=0.3 & IoU=0.5 & IoU=0.7 \\ \hline
     ~ & \textbf{w/ $\mathcal{L}_{DV}$} & \textbf{71.58} & \textbf{54.80} & \textbf{35.77} & \textbf{98.36} & \textbf{88.90} & \textbf{61.02}& \textbf{70.25} & \textbf{52.68} & \textbf{46.19} & \textbf{88.73} & \textbf{79.40} & \textbf{60.17} \\
     Domain & w/o $\mathcal{L}_{DV}$ &65.83&44.32&30.89&90.15&81.30&53.37&64.24&44.99&39.51& 80.76&72.09&53.83 \\ \cline{2-14}
     Alignment  & \textbf{w/ $\mathcal{L}_{DQ}$} & \textbf{71.58} & \textbf{54.80} & \textbf{35.77} & \textbf{98.36} & \textbf{88.90} & \textbf{61.02}& \textbf{70.25} & \textbf{52.68} & \textbf{46.19} & \textbf{88.73} & \textbf{79.40} & \textbf{60.17} \\
     ~ & w/o $\mathcal{L}_{DQ}$ &66.34&47.38&30.85&92.31&82.47&54.94&65.03&45.38&40.12&81.33&73.24&54.63  \\ \hline
      & \textbf{w/ $\mathcal{L}_{M1}$} & \textbf{71.58} & \textbf{54.80} & \textbf{35.77} & \textbf{98.36} & \textbf{88.90} & \textbf{61.02}& \textbf{70.25} & \textbf{52.68} & \textbf{46.19} & \textbf{88.73} & \textbf{79.40} & \textbf{60.17} \\
     Cross-modal& w/o $\mathcal{L}_{M1}$ &68.34&48.96&32.75&95.17&84.06&59.08&67.62&48.36&43.75&82.63&75.90&55.41  \\ \cline{2-14}
     Alignment  & \textbf{w/ $\mathcal{L}_{M2}$} & \textbf{71.58} & \textbf{54.80} & \textbf{35.77} & \textbf{98.36} & \textbf{88.90} & \textbf{61.02}& \textbf{70.25} & \textbf{52.68} & \textbf{46.19} & \textbf{88.73} & \textbf{79.40} & \textbf{60.17}  \\
     ~ & w/o $\mathcal{L}_{M2}$ &70.03&50.21&33.84&97.18&86.35&60.38&68.42&49.17&44.63&83.01&76.34&57.29  \\ \hline
         \hline
    \multirow{3}*{{Module}} & \multirow{3}*{{Changes}} & \multicolumn{6}{c|}{{A$\rightarrow$T}}&\multicolumn{6}{c}{{T$\rightarrow$A}}\\\cline{3-14}
    &&{R@1,} & {R@1,}& {R@1,} & {R@5,} & {R@5,} & {R@5,} &{R@1,} & {R@1,}& {R@1,} & {R@5,} & {R@5,} & {R@5,} \\
     ~ & ~ & {IoU=0.1} &{IoU=0.3} & {IoU=0.5}& {IoU=0.1} & {IoU=0.3} & {IoU=0.5}& {IoU=0.3} &{IoU=0.5} & {IoU=0.7}& {IoU=0.3} & {IoU=0.5} & {IoU=0.7} \\ \hline
     ~ & {\textbf{w/ $\mathcal{L}_{DV}$}}&{\textbf{55.38}} & {\textbf{43.07}} & {\textbf{31.69}} & {\textbf{77.54}} & {\textbf{63.81}} & {\textbf{51.52}}&{\textbf{42.58}} & {\textbf{20.90}} & {\textbf{14.69}} & {\textbf{58.75}} & {\textbf{46.81}} & {\textbf{40.13}}\\
     {Domain} & {w/o $\mathcal{L}_{DV}$}&{53.42}&{40.81}&{27.35}&{73.80}&{60.35}&{48.26}&{39.97}&{18.96}&{13.21}&{54.96}&{43.28}&{37.95}\\
     {Alignment}  & {\textbf{w/ $\mathcal{L}_{DQ}$}}&{\textbf{55.38}} & {\textbf{43.07}} & {\textbf{31.69}} & {\textbf{77.54}} & {\textbf{63.81}} & {\textbf{51.52}}&{\textbf{42.58}} & {\textbf{20.90}} & {\textbf{14.69}} & {\textbf{58.75}} & {\textbf{46.81}} & {\textbf{40.13}} \\
     ~ & {w/o $\mathcal{L}_{DQ}$}&{53.94}&{41.12}&{28.75}&{74.28}&{61.37}&{48.93}&{40.26}&{19.35}&{13.62}&{55.78}&{44.89}&{38.74}\\\hline
     & {\textbf{w/ $\mathcal{L}_{M1}$}}&{\textbf{55.38}} & {\textbf{43.07}} & {\textbf{31.69}} & {\textbf{77.54}} & {\textbf{63.81}} & {\textbf{51.52}}&{\textbf{42.58}} & {\textbf{20.90}} & {\textbf{14.69}} & {\textbf{58.75}} & {\textbf{46.81}} & {\textbf{40.13}}\\
     {Cross-modal}& {w/o $\mathcal{L}_{M1}$}&{54.26}&{41.38}&{28.94}&{75.62}&{61.53}&{49.04}&{40.72}&{19.46}&{13.95}&{56.27}&{45.08}&{38.96}\\
     {Alignment}  & {\textbf{w/ $\mathcal{L}_{M2}$}}&{\textbf{55.38}} & {\textbf{43.07}} & {\textbf{31.69}} & {\textbf{77.54}} & {\textbf{63.81}} & {\textbf{51.52}}&{\textbf{42.58}} & {\textbf{20.90}} & {\textbf{14.69}} & {\textbf{58.75}} & {\textbf{46.81}} & {\textbf{40.13}}\\
      ~ & {w/o $\mathcal{L}_{M2}$}&{54.03}&{41.27}&{29.05}&{75.73}&{61.45}&{49.72}&{40.93}&{19.32}&{13.87}&{56.94}&{45.36}&{39.18}\\\hline
         \hline
    \multirow{3}*{{Module}} & \multirow{3}*{{Changes}} & \multicolumn{6}{c|}{{C$\rightarrow$T}}&\multicolumn{6}{c}{{T$\rightarrow$C}}\\\cline{3-14}
    &&{R@1,} & {R@1,}& {R@1,} & {R@5,} & {R@5,} & {R@5,} &{R@1,} & {R@1,}& {R@1,} & {R@5,} & {R@5,} & {R@5,} \\
     ~ & ~ & {IoU=0.1} &{IoU=0.3} & {IoU=0.5}& {IoU=0.1} & {IoU=0.3} & {IoU=0.5}& {IoU=0.3} &{IoU=0.5} & {IoU=0.7}& {IoU=0.3} & {IoU=0.5} & {IoU=0.7} \\ \hline
     ~ & {\textbf{w/ $\mathcal{L}_{DV}$}}&{\textbf{47.97}} & {\textbf{35.41}} & {\textbf{26.88}} & {\textbf{69.53}} & {\textbf{58.10}} & {\textbf{42.62}}&{\textbf{48.69}} & {\textbf{35.17}} & {\textbf{17.28}} & {\textbf{85.40}} & {\textbf{69.93}} & {\textbf{39.74}}\\
     {Domain} & {w/o $\mathcal{L}_{DV}$}&{45.74}&{33.02}&{23.43}&{66.08}&{55.49}&{39.90}&{45.03}&{33.72}&{16.35}&{83.92}&{67.30}&{35.98}\\
     {Alignment}  & {\textbf{w/ $\mathcal{L}_{DQ}$}}&{\textbf{47.97}} & {\textbf{35.41}} & {\textbf{26.88}} & {\textbf{69.53}} & {\textbf{58.10}} & {\textbf{42.62}}&{\textbf{48.69}} & {\textbf{35.17}} & {\textbf{17.28}} & {\textbf{85.40}} & {\textbf{69.93}} & {\textbf{39.74}} \\
     ~ & {w/o $\mathcal{L}_{DQ}$}&{46.02}&{32.84}&{23.72}&{65.11}&{55.76}&{40.31}&{45.32}&{33.60}&{16.03}&{83.75}&{67.84}&{36.20}\\\hline
     & {\textbf{w/ $\mathcal{L}_{M1}$}}&{\textbf{47.97}} & {\textbf{35.41}} & {\textbf{26.88}} & {\textbf{69.53}} & {\textbf{58.10}} & {\textbf{42.62}}&{\textbf{48.69}} & {\textbf{35.17}} & {\textbf{17.28}} & {\textbf{85.40}} & {\textbf{69.93}} & {\textbf{39.74}}\\
     {Cross-modal}& {w/o $\mathcal{L}_{M1}$}&{46.84}&{33.78}&{24.13}&{66.49}&{56.17}&{40.96}&{45.43}&{33.85}&{16.84}&{84.12}&{68.01}&{36.87}\\
     {Alignment}  & {\textbf{w/ $\mathcal{L}_{M2}$}}&{\textbf{47.97}} & {\textbf{35.41}} & {\textbf{26.88}} & {\textbf{69.53}} & {\textbf{58.10}} & {\textbf{42.62}}&{\textbf{48.69}} & {\textbf{35.17}} & {\textbf{17.28}} & {\textbf{85.40}} & {\textbf{69.93}} & {\textbf{39.74}}\\
      ~ & {w/o $\mathcal{L}_{M2}$}&{47.16}&{34.02}&{24.75}&{66.14}&{56.85}&{41.77}&{46.37}&{34.12}&{16.99}&{84.24}&{68.72}&{37.41}\\\hline
    \end{tabular}}
    \label{tab:ablation_two_align}
\end{table*}

\noindent \textbf{Training process of different ablation models.} Following \cite{LinZZWL20}, we conduct six experiments to analyze the training process of different ablation models on each transfer task. Fig. \ref{fig:xunliantu} shows the results, where ``mIoU'' is the evaluation metric. We can obtain the following representative observations:
(i) On each epoch, MMCDA(full) outperforms ablation models, which further demonstrates the effectiveness of each module. (ii) MMCDA(full) performs better in the A$\rightarrow$C task than in the C$\rightarrow$A task. It is because  ActivityNet Captions  (20,000 videos) contains more  annotated videos (annotation knowledge) than Charades-STA (9,848 videos). (iii) For each transfer task, MMCDA(full) converges faster than ablation models, which shows that the full model is more efficient on time-consuming.

\begin{table*}[t!]
   \centering
    \caption{Ablation study on hyperparameters $\gamma_1$, $\gamma_2$, $\gamma_3$.}
    \begin{center}
 \vspace{-10pt}
    \begin{tabular}{c|cccc|cccc|ccccccc}
        \toprule
        \multirow{3}*{$\gamma_1$}& \multicolumn{4}{c|}{A$\rightarrow$T}&\multicolumn{4}{c|}{T$\rightarrow$C}&\multicolumn{4}{c}{C$\rightarrow$A}\\\cline{2-13}
~& R@1  & R@1 & R@5  & R@5& R@1  & R@1 & R@5  & R@5& R@1  & R@1 & R@5 & R@5\\
~   &  IoU=0.3 & IoU=0.5  & IoU=0.3 & IoU=0.5&  IoU=0.3 & IoU=0.5  & IoU=0.3 & IoU=0.5 & IoU=0.3 & IoU=0.5  & IoU=0.3 & IoU=0.5\\\midrule
        0.8  & 42.89&30.04&61.31&50.92&47.62&33.09&83.01&68.26&68.83&50.12&87.90&77.32\\
        0.9  & 43.01&31.45&62.14&51.03&48.17&34.92&84.89&68.91&69.47&51.83&88.20&78.74\\
        \textbf{1.0} & \textbf{43.07} & \textbf{31.69}  & \textbf{63.81} & \textbf{51.52}& \textbf{48.69}& \textbf{35.17}& \textbf{85.40}& \textbf{69.93}& \textbf{70.25}& \textbf{52.68}& \textbf{88.73}& \textbf{79.40}   \\
        1.1  &42.31&30.07&62.19&50.10&47.98&34.03&84.92&68.33&69.15&51.32&87.39&78.10\\
        1.2&40.24&28.19&60.32&49.88&46.63&33.77&83.45&67.51&68.06&50.87&86.58&77.93\\\bottomrule\bottomrule
        \multirow{3}*{$\gamma_2$}& \multicolumn{4}{c|}{A$\rightarrow$T}&\multicolumn{4}{c|}{T$\rightarrow$C}&\multicolumn{4}{c}{C$\rightarrow$A}\\\cline{2-13}
~& R@1  & R@1 & R@5  & R@5& R@1  & R@1 & R@5  & R@5& R@1  & R@1 & R@5 & R@5\\
~   &  IoU=0.3 & IoU=0.5  & IoU=0.3 & IoU=0.5&  IoU=0.3 & IoU=0.5  & IoU=0.3 & IoU=0.5 & IoU=0.3 & IoU=0.5  & IoU=0.3 & IoU=0.5\\\midrule
        0.3&41.10&30.25&62.11&50.19&46.07&33.91&83.15&67.83&68.49&50.14&86.46&77.03\\
        0.4& 42.11&31.12&62.98&51.17&47.86&34.40&84.52&69.02&69.19&51.30&87.98&78.13\\
        \textbf{0.5}& \textbf{43.07} & \textbf{31.69}  & \textbf{63.81} & \textbf{51.52}& \textbf{48.69}& \textbf{35.17}& \textbf{85.40}& \textbf{69.93}& \textbf{70.25}& \textbf{52.68}& \textbf{88.73}& \textbf{79.40}\\
        0.6&41.39&30.18&62.90&50.30&47.26&33.32&84.41&68.89&69.97&51.28&87.26&78.74\\
        0.7 & 40.92&29.83&61.47&50.01&46.74&32.19&83.38&68.42&69.02&50.10&86.04&77.31\\\bottomrule\bottomrule
        \multirow{3}*{$\gamma_3$}& \multicolumn{4}{c|}{A$\rightarrow$T}&\multicolumn{4}{c|}{T$\rightarrow$C}&\multicolumn{4}{c}{C$\rightarrow$A}\\\cline{2-13}
~& R@1  & R@1 & R@5  & R@5& R@1  & R@1 & R@5  & R@5& R@1  & R@1 & R@5 & R@5\\
~   &  IoU=0.3 & IoU=0.5  & IoU=0.3 & IoU=0.5&  IoU=0.3 & IoU=0.5  & IoU=0.3 & IoU=0.5 & IoU=0.3 & IoU=0.5  & IoU=0.3 & IoU=0.5\\\midrule
        0.1&42.33&31.02&63.15&50.90&47.96&34.03&83.47&68.99&69.72&51.03&87.52&78.49\\
        \textbf{0.2}&  \textbf{43.07} & \textbf{31.69}  & \textbf{63.81} & \textbf{51.52}& \textbf{48.69}& \textbf{35.17}& \textbf{85.40}& \textbf{69.93}& \textbf{70.25}& \textbf{52.68}& \textbf{88.73}& \textbf{79.40}\\
        0.3&42.25&30.91&62.90&50.93&48.07&34.16&84.91&69.18&68.37&51.48&87.96&79.01\\
        0.4  &41.93&29.85&62.04&50.17&47.82&34.03&84.25&68.72&68.09&51.19&87.14&78.17\\
        0.5  & 41.31&29.11&61.69&49.82&46.79&33.25&83.62&67.04&66.27&50.53&85.52&76.38\\
        \bottomrule
    \end{tabular}
    \end{center}
    \label{tab:chaocan}
\end{table*}

\noindent \textbf{Analysis on domain alignment and cross-modal alignment.}
As shown in Table \ref{tab:ablation_two_align}, we also conduct ablation study within the domain alignment and cross-modal alignment examine the effectiveness of each loss function. For the domain alignment, the video-based domain alignment loss $\mathcal{L}_{DV}$ and the query-based domain alignment loss $\mathcal{L}_{DQ}$ bring significant improvement in all metrics on both tasks. Especially,  $\mathcal{L}_{DV}$ brings 8.36\%  improvement in ``R@1, IoU=0.5'' on the A$\rightarrow$C transfer task, and raises the performance around 7.97\% in ``R@5, IoU=0.3'' on the C$\rightarrow$A transfer task. Besides, $\mathcal{L}_{DQ}$ obtains the performance improvement about 6.43\% in ``R@5, IoU=0.5'' on the A$\rightarrow$C transfer task, and brings 7.40\% improvement in ``R@5, IoU=0.3'' on the C$\rightarrow$A transfer task. As for the cross-modal alignment, the cross-modal consistent loss $\mathcal{L}_{M1}$ and the cross-modal feature distribution loss $\mathcal{L}_{M2}$ also effectively improve the grounding accuracy. For example, on the A$\rightarrow$C transfer task, $\mathcal{L}_{M1}$ raises the performance around 4.24\% and 4.84\% in ``R@1, IoU=0.3'' and ``R@5, IoU=0.5'' respectively, while it achieves the performance improvement about 4.32\% and 6.10\% in ``R@1, IoU=0.5'' and ``R@5, IoU=0.3'' respectively on the C$\rightarrow$A transfer task. Moreover, $\mathcal{L}_{M2}$ brings 2.42\% improvement in ``R@1, IoU=0.5'' on the A$\rightarrow$C transfer task, and brings 5.62\%  improvement in ``R@5, IoU=0.3'' on the C$\rightarrow$A transfer task.

\begin{table*}[t!]
    \centering
    \caption{{Ablation study on the video and query encoders for all the tasks.}}
    \vspace{-10pt}
    \setlength{\tabcolsep}{1.0mm}{
    \begin{tabular}{c|c|cccccc|cccccccccc}
    \hline
    \multirow{3}*{Module} & \multirow{3}*{Changes} & \multicolumn{6}{c|}{{A$\rightarrow$C}}&\multicolumn{6}{c}{{C$\rightarrow$A}}\\\cline{3-14}
    &&{R@1,} & {R@1,}& {R@1,} & {R@5,} & {R@5,} & {R@5,} &{R@1,} & {R@1,}& {R@1,} & {R@5,} & {R@5,} & {R@5,} \\
     ~ & ~ & {IoU=0.3} &{IoU=0.5} & {IoU=0.7}& {IoU=0.3} & {IoU=0.5} & {IoU=0.7}& {IoU=0.3} &{IoU=0.5} & {IoU=0.7}& {IoU=0.3} & {IoU=0.5} & {IoU=0.7} \\ \hline
     ~& \textbf{w/  Bi-GRU} & {\textbf{71.58}} & {\textbf{54.80}} & {\textbf{35.77}} & {\textbf{98.36}} & {\textbf{88.90}} & {\textbf{61.02}}& {\textbf{70.25}} & {\textbf{52.68}} & {\textbf{46.19}} & {\textbf{88.73}} & {\textbf{79.40}} & {\textbf{60.17}}\\
     Video & w/o Bi-GRU &{69.83}& 53.01&33.87&{96.02}&87.42&60.58&{68.94}&{50.36}&{43.51}&{87.25}&{78.02}&{59.13} \\ \cline{2-14}
     Encoder & \textbf{w/ Self-attention} &  {\textbf{71.58}} & {\textbf{54.80}} & {\textbf{35.77}} & {\textbf{98.36}} & {\textbf{88.90}} & {\textbf{61.02}}& {\textbf{70.25}} & {\textbf{52.68}} & {\textbf{46.19}} & {\textbf{88.73}} & {\textbf{79.40}} & {\textbf{60.17}} \\
     ~ & w/o Self-attention &{69.42}&  53.76&34.89&{97.08}&86.31&59.64&{68.02}&{51.13}&{44.26}&{86.98}&{77.86}&{58.40}\\ \hline
    ~ & \textbf{w/ Bi-GRU} &  {\textbf{71.58}} & {\textbf{54.80}} & {\textbf{35.77}} & {\textbf{98.36}} & {\textbf{88.90}} & {\textbf{61.02}}& {\textbf{70.25}} & {\textbf{52.68}} & {\textbf{46.19}} & {\textbf{88.73}} & {\textbf{79.40}} & {\textbf{60.17}} \\
     Query  & w/o Bi-GRU & {68.60}& 51.72&34.17&{96.50}&87.16&59.93&{68.36}&{51.47}&{44.82}&{87.05}&{78.14}&{59.37}\\ \cline{2-14}
     Encoder& \textbf{w/ Self-attention} & {\textbf{71.58}} & {\textbf{54.80}} & {\textbf{35.77}} & {\textbf{98.36}} & {\textbf{88.90}} & {\textbf{61.02}}& {\textbf{70.25}} & {\textbf{52.68}} & {\textbf{46.19}} & {\textbf{88.73}} & {\textbf{79.40}} & {\textbf{60.17}} \\
     ~ & w/o Self-attention & {69.28}& 52.25&33.75&{96.72}&85.92&59.13&{68.74}&{51.26}&{45.30}&{86.41}&{77.59}&{58.81}\\ \hline
    \hline
    \multirow{3}*{{Module}} & \multirow{3}*{{Changes}} & \multicolumn{6}{c|}{{A$\rightarrow$T}}&\multicolumn{6}{c}{{T$\rightarrow$A}}\\\cline{3-14}
    &&{R@1,} & {R@1,}& {R@1,} & {R@5,} & {R@5,} & {R@5,} &{R@1,} & {R@1,}& {R@1,} & {R@5,} & {R@5,} & {R@5,} \\
     ~ & ~ & {IoU=0.1} &{IoU=0.3} & {IoU=0.5}& {IoU=0.1} & {IoU=0.3} & {IoU=0.5}& {IoU=0.3} &{IoU=0.5} & {IoU=0.7}& {IoU=0.3} & {IoU=0.5} & {IoU=0.7} \\ \hline
     ~& {\textbf{w/  Bi-GRU}} & {\textbf{55.38}} & {\textbf{43.07}} & {\textbf{31.69}} & {\textbf{77.54}} & {\textbf{63.81}} & {\textbf{51.52}}&{\textbf{42.58}} & {\textbf{20.90}} & {\textbf{14.69}} & {\textbf{58.75}} & {\textbf{46.81}} & {\textbf{40.13}}\\
     {Video} & {w/o Bi-GRU} & {54.35}&{42.61}&{30.04}&{76.53}&{62.90}&{50.43}&{40.99}&{19.02}&{13.18}&{56.39}&{44.72}&{38.26} \\ \cline{2-14}
     {Encoder} & {\textbf{w/ Self-attention}} & {\textbf{55.38}} & {\textbf{43.07}} & {\textbf{31.69}} & {\textbf{77.54}} & {\textbf{63.81}} & {\textbf{51.52}}&{\textbf{42.58}} & {\textbf{20.90}} & {\textbf{14.69}} & {\textbf{58.75}} & {\textbf{46.81}} & {\textbf{40.13}}\\
     ~ & {w/o Self-attention} & {53.97}&{42.14}&{29.87}&{75.32}&{61.37}&{49.92}&{40.26}&{19.23}&{13.58}&{56.30}&{45.12}&{39.28} \\ \hline
    ~ & {\textbf{w/ Bi-GRU}} &  {\textbf{55.38}} & {\textbf{43.07}} & {\textbf{31.69}} & {\textbf{77.54}} & {\textbf{63.81}} & {\textbf{51.52}}&{\textbf{42.58}} & {\textbf{20.90}} & {\textbf{14.69}} & {\textbf{58.75}} & {\textbf{46.81}} & {\textbf{40.13}}\\
     {Query}  & {w/o Bi-GRU} & {54.03}&{41.78}&{29.05}&{77.06}&{61.85}&{49.26}&{39.15}&{18.46}&{12.87}&{56.04}&{44.83}&{38.71} \\ \cline{2-14}
     {Encoder}& {\textbf{w/ Self-attention}} &   {\textbf{55.38}} & {\textbf{43.07}} & {\textbf{31.69}} & {\textbf{77.54}} & {\textbf{63.81}} & {\textbf{51.52}}&{\textbf{42.58}} & {\textbf{20.90}} & {\textbf{14.69}} & {\textbf{58.75}} & {\textbf{46.81}} & {\textbf{40.13}}\\
     ~ & {w/o Self-attention} &{54.40}&{41.92}&{29.30}&{75.48}&{61.02}&{49.85}&{39.55}&{18.94}&{13.70}&{56.81}&{44.92}&{38.45}  \\ \hline
    \hline
    \multirow{3}*{{Module}} & \multirow{3}*{{Changes}} & \multicolumn{6}{c|}{{C$\rightarrow$T}}&\multicolumn{6}{c}{{T$\rightarrow$C}}\\\cline{3-14}
    &&{R@1,} & {R@1,}& {R@1,} & {R@5,} & {R@5,} & {R@5,} &{R@1,} & {R@1,}& {R@1,} & {R@5,} & {R@5,} & {R@5,} \\
     ~ & ~ & {IoU=0.1} &{IoU=0.3} & {IoU=0.5}& {IoU=0.1} & {IoU=0.3} & {IoU=0.5}& {IoU=0.3} &{IoU=0.5} & {IoU=0.7}& {IoU=0.3} & {IoU=0.5} & {IoU=0.7} \\ \hline
     ~& {\textbf{w/  Bi-GRU}} & {\textbf{47.97}} & {\textbf{35.41}} & {\textbf{26.88}} & {\textbf{69.53}} & {\textbf{58.10}} & {\textbf{42.62}}&{\textbf{48.69}} & {\textbf{35.17}} & {\textbf{17.28}} & {\textbf{85.40}} & {\textbf{69.93}} & {\textbf{39.74}}\\
     {Video} & {w/o Bi-GRU} &{45.23}&{34.50}&{24.77}&{67.35}&{56.84}&{41.05}&{46.82}&{34.06}&{15.27}&{83.82}&{67.19}&{37.58}  \\ \cline{2-14}
     {Encoder} & {\textbf{w/ Self-attention}} & {\textbf{47.97}} & {\textbf{35.41}} & {\textbf{26.88}} & {\textbf{69.53}} & {\textbf{58.10}} & {\textbf{42.62}}&{\textbf{48.69}} & {\textbf{35.17}} & {\textbf{17.28}} & {\textbf{85.40}} & {\textbf{69.93}} & {\textbf{39.74}}\\
     ~ & {w/o Self-attention} & {45.83}&{34.26}&{25.19}&{68.03}&{57.15}&{41.33}&{47.26}&{33.08}&{16.42}&{84.96}&{68.49}&{37.80} \\ \hline
    ~ & {\textbf{w/ Bi-GRU}} & {\textbf{47.97}} & {\textbf{35.41}} & {\textbf{26.88}} & {\textbf{69.53}} & {\textbf{58.10}} & {\textbf{42.62}}&{\textbf{48.69}} & {\textbf{35.17}} & {\textbf{17.28}} & {\textbf{85.40}} & {\textbf{69.93}} & {\textbf{39.74}}\\
     {Query}  & {w/o Bi-GRU} & {46.08}&{34.12}&{23.75}&{67.82}&{56.71}&{40.79}&{46.34}&{34.18}&{16.85}&{84.11}&{67.82}&{37.43} \\ \cline{2-14}
     {Encoder}& {\textbf{w/ Self-attention}} & {\textbf{47.97}} & {\textbf{35.41}} & {\textbf{26.88}} & {\textbf{69.53}} & {\textbf{58.10}} & {\textbf{42.62}}&{\textbf{48.69}} & {\textbf{35.17}} & {\textbf{17.28}} & {\textbf{85.40}} & {\textbf{69.93}} & {\textbf{39.74}}\\
     ~ & {w/o Self-attention} & {45.16}&{33.71}&{24.06}&{67.44}&{56.23}&{40.10}&{47.02}&{33.28}&{15.35}&{83.74}&{68.42}&{36.85} \\ \hline
    \end{tabular}}
    \label{tab:ablation2}
\end{table*}

\begin{table*}[t!]
    \centering
    \caption{{Ablation study on the intra-sample distribution function $\mu(\cdot)$   and the inter-sample distribution function $\sigma(\cdot)$ on all the tasks.}}
    \vspace{-10pt}
    \setlength{\tabcolsep}{1mm}{
    \begin{tabular}{c|c|cc|cc|cc|cc|cc|ccccccccccc}
    \hline
     \multirow{3}*{{Loss}} & \multirow{3}*{{Changes}}& \multicolumn{2}{c|}{{{A$\rightarrow$C}}}&\multicolumn{2}{c|}{{{C$\rightarrow$A}}}&\multicolumn{2}{c|}{{A$\rightarrow$T}}&\multicolumn{2}{c|}{{T$\rightarrow$A}}&\multicolumn{2}{c|}{{C$\rightarrow$T}}&\multicolumn{2}{c}{{T$\rightarrow$C}}\\\cline{3-14}
    & & {R@1,} & {R@1,}& {R@1,} & {R@1,} & {R@1,} & {R@1,}& {R@1,} & {R@1,}& {R@1,} & {R@1,}& {R@1,} & {R@1,} \\
    &&   {IoU=0.3} & {IoU=0.5} & {IoU=0.3} & {IoU=0.5} &   {IoU=0.3} & {IoU=0.5} & {IoU=0.3} & {IoU=0.5}&{IoU=0.3} & {IoU=0.5} & {IoU=0.3} & {IoU=0.5} \\\hline
     \multirow{4}*{{$\mathcal{L}_{DV}$}}& \textbf{{w/  $\mu(\cdot)$}} & {\textbf{71.58}} & {\textbf{54.80}} & {\textbf{70.25}} & {\textbf{52.68}} & {\textbf{43.07}} &{\textbf{31.69}}&{\textbf{42.58}}&{\textbf{20.90}}&{\textbf{35.41}}&{\textbf{26.88}}&{\textbf{48.69}}&{\textbf{35.17}}\\
    ~& {w/o $\mu(\cdot)$} & {69.31} &{51.27} &{68.97} &{51.35}&{41.88}&{28.46}&{39.20}&{18.72}&{33.98}&{25.03}&{46.85}&{32.34} \\ \cline{2-14}
     ~ & {\textbf{w/ $\sigma(\cdot)$}} &   {\textbf{71.58}} & {\textbf{54.80}} & {\textbf{70.25}} & {\textbf{52.68}}&{\textbf{43.07}} &{\textbf{31.69}}&{\textbf{42.58}}&{\textbf{20.90}}&{\textbf{35.41}}&{\textbf{26.88}}&{\textbf{48.69}}&{\textbf{35.17}} \\
     ~ & {w/o $\sigma(\cdot)$} &  {69.16}&{51.59}&{68.84}&{51.42}&{41.76}&{27.42}&{38.99}&{19.03}&{34.05}&{24.87}&{46.62}&{32.01}\\ \hline
    \multirow{4}*{{$\mathcal{L}_{DQ}$}}& {\textbf{w/  $\mu(\cdot)$}} & {\textbf{71.58}} & {\textbf{54.80}} & {\textbf{70.25}} & {\textbf{52.68}}&{\textbf{43.07}} &{\textbf{31.69}}&{\textbf{42.58}}&{\textbf{20.90}}&{\textbf{35.41}}&{\textbf{26.88}}&{\textbf{48.69}}&{\textbf{35.17}} \\
     ~  & {w/o $\mu(\cdot)$} & {69.84}&{51.02}&{69.24}&{51.76}&{42.03}&{29.70}&{39.58}&{19.17}&{34.25}&{24.99}&{47.06}&{32.94}\\ \cline{2-14}
     ~& {\textbf{w/ $\sigma(\cdot)$}} &  {\textbf{71.58}} & {\textbf{54.80}} & {\textbf{70.25}} & {\textbf{52.68}}&{\textbf{43.07}} &{\textbf{31.69}}&{\textbf{42.58}}&{\textbf{20.90}}&{\textbf{35.41}}&{\textbf{26.88}}&{\textbf{48.69}}&{\textbf{35.17}} \\
     ~ & {w/o $\sigma(\cdot)$} &{68.92}&{50.40}&{68.84}&{51.03}&{41.92}&{28.85}&{39.01}&{19.22}&{34.07}&{25.12}&{47.23}&{33.85}\\ \hline
    \multirow{4}*{{$\mathcal{L}_{M1}$}} & {\textbf{w/  $\mu(\cdot)$}} & {\textbf{71.58}} & {\textbf{54.80}} & {\textbf{70.25}} & {\textbf{52.68}}&{\textbf{43.07}} &{\textbf{31.69}}&{\textbf{42.58}}&{\textbf{20.90}}&{\textbf{35.41}}&{\textbf{26.88}}&{\textbf{48.69}}&{\textbf{35.17}} \\
     ~  & {w/o $\mu(\cdot)$} & {70.04}&{49.76}&{67.81}&{50.35}&{42.01}&{29.53}&{39.60}&{19.28}&{34.59}&{25.05}&{47.35}&{33.29}\\\cline{2-14}
     ~& {\textbf{w/ $\sigma(\cdot)$}} &  {\textbf{71.58}} & {\textbf{54.80}} & {\textbf{70.25}} & {\textbf{52.68}}&{\textbf{43.07}} &{\textbf{31.69}}&{\textbf{42.58}}&{\textbf{20.90}}&{\textbf{35.41}}&{\textbf{26.88}}&{\textbf{48.69}}&{\textbf{35.17}} \\
     ~ & {w/o $\sigma(\cdot)$} &{69.42}&{49.25}&{68.34}&{51.05}&{42.17}&{30.88}&{40.56}&{19.04}&{34.72}&{25.06}&{47.38}&{34.01}\\ \hline
    \end{tabular}}
    \label{tab:intra-inter}
\end{table*}

\noindent \textbf{Analysis on hyperparameters.} As shown in Table \ref{tab:chaocan}, we analyze the impact of three hyperparameters: $\gamma_1$, $\gamma_2$, and $\gamma_3$. It can be observed that, with the increase of $\gamma_1$, $\gamma_2$ and $\gamma_3$, their performance follows a general trend, \textit{i.e.}, rises at first and then starts to decline. It is because when these three hyperparameters  are relatively small, our MMCDA pays more attention to the fully-supervised learning during the pre-training phase in the source domain, and it can obtain more annotation knowledge from the source domain. As the increase of these  hyperparameters,  MMCDA adds the attention to the multi-modal cross-domain alignment module during the main training phase, which helps MMCDA transfer the annotation knowledge from the source domain to the target domain. However, when these  hyperparameters are too large, MMCDA may have difficulty in obtaining the annotation knowledge, which will lead the performance drop. Therefore, in our paper, we set $\gamma_1=1.0$, $\gamma_2=0.5$ and $\gamma_3=0.2$, where our MMCDA can obtain the best performance.

\noindent \textbf{Analysis on the multi-modal encoders.}
As shown in Table \ref{tab:ablation2},
{we investigate different variants of multi-modal encoders on all the tasks.  In the A$\rightarrow$C task, for the video encoder, we can observe that  Bi-GRU  brings the improvement of 1.19\%, 1.76\%,  1.90\%, 2.34\%, 1.48\% and  0.44\% in all metrics, which demonstrates that Bi-GRU  can extract the sequential information among these consecutive frames. Also, the self-attention module  improves  performance (2.16\%, 1.04\%,  0.88\%, 1.28\%, 2.59\% and 1.38\%) to the full model. The  performance boost is due to its ability to capture the intra-video contexts. As for the query encoder, we could find that if we remove the Bi-GRU module, the full model will decrease the performance of 2.30\%, 2.08\%,  1.60\%, 1.64\%, 1.74\% and  1.09\% in all metrics, which presents the effectiveness of Bi-GRU in incorporating the contextual textual information. Moreover, the self-attention module improves the accuracy of  2.30\%, 2.51\%,  2.02\%, 1.64\%,  2.98\% and  1.89\% since it can obtain the semantic dependencies in the long query.}
\begin{table*}[t!]
    \centering
    \caption{{Performance comparison of our MMCDA with state-of-the-art multi-modal cross-domain algorithms on all the tasks. Note that we only compare our MMCDA to state-of-the-art open-source algorithms. In \cite{silva2021embracing,zhao2021cross}, these authors do not name their models, for convenience, we name their designed ``Cross-domain Fake News Detection using Multi-modal Data'' framework as ``CFND'' and ``Cross-Modal Retrieval and Model Adaptation'' as ``CRMA''.}}
    \vspace{-10pt}
    \setlength{\tabcolsep}{1.5mm}{
    \begin{tabular}{c|cc|cc|cc|cc|cc|ccccccccccccccc}
    \hline
    \multirow{3}*{Model} & \multicolumn{2}{c|}{A$\rightarrow$C}&\multicolumn{2}{c|}{C$\rightarrow$A}&\multicolumn{2}{c|}{A$\rightarrow$T}&\multicolumn{2}{c|}{T$\rightarrow$A}&\multicolumn{2}{c|}{C$\rightarrow$T}&\multicolumn{2}{c}{T$\rightarrow$C}\\\cline{2-13}
    ~&R@1, & R@1, & R@1, & R@1,& R@1, & R@1, &R@1, & R@1, & R@1, & R@1,& R@1, & R@1,  \\
     ~ &  IoU=0.3 & IoU=0.5 &IoU=0.3 & IoU=0.5 &   IoU=0.3 & IoU=0.5 & IoU=0.3 & IoU=0.5 &   IoU=0.3 & IoU=0.5 &IoU=0.3 & IoU=0.5  \\ \hline
     xMUDA \cite{jaritz2020xmuda}&66.33&46.94&66.10&48.02&{38.45}&{26.92}&{36.45}&{17.26}&{30.17}&{23.10}&{40.52}&{29.33}\\
     MM-SADA \cite{munro2020multi}&67.56&48.05&67.30&48.86&{37.97}&{26.43}&{36.08}&{18.25}&{31.50}&{23.38}&{44.81}&{30.15}\\
     PMC \cite{zhang2021progressive}&68.91&48.36&68.27&49.87&{39.72}&{26.05}&{38.24}&{17.83}&{30.81}&{23.96}&{45.86}&{29.74}\\
   {CrossCLR \cite{zolfaghari2021crossclr}}&{69.07}&{48.41}&{68.54}&{50.08}&{40.13}&{26.84}&{38.50}&{18.15}&{32.96}&{24.11}&{45.25}&{30.82}\\
  {CFND \cite{silva2021embracing}} &{69.12}&{49.50}&{68.71}&{50.23}&{41.06}&{26.45}&{37.82}&{18.01}&{33.42}&{24.53}&{45.84}&{31.53}\\
   {CRMA \cite{zhao2021cross}}&{69.35}&{50.88}&{68.84}&{50.59}&{40.85}&{27.06}&{38.94}&{18.23}&{33.86}&{24.72}&{46.50}&{31.95}\\\hline
     \textbf{Our MMCDA}&\textbf{71.58} & \textbf{54.80} & \textbf{70.25} & \textbf{52.68}& {\textbf{43.07}} &{\textbf{31.69}}&{\textbf{42.58}}&{\textbf{20.90}}&{\textbf{35.41}}&{\textbf{26.88}}&{\textbf{48.69}}&{\textbf{35.17}}\\
      \hline
    \end{tabular}}
    \label{tab:ablation_xingnengduibi}
\end{table*}
\begin{figure*}[t!]
\centering
\includegraphics[width=0.8\textwidth]{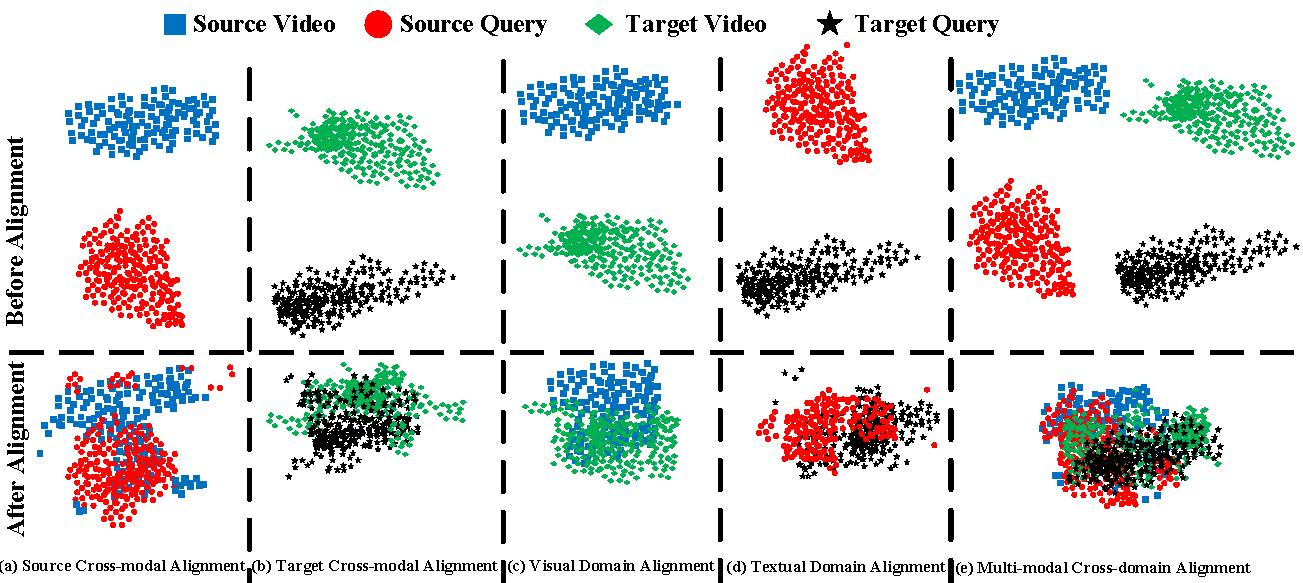}
\vspace{-12pt}
\caption{{T-SNE visualization on multi-modal cross-domain features on the A$\rightarrow$C task. (a) and (b) are the visualization of cross-modal alignment for individual domains, and each domain contains the features from videos and queries. Similarly, in (c) and (d), for each modality, we visualize the domain alignment, where the features from both source and target domains are utilized. In (e), we integrate all the features from two domains and two modalities based on both cross-modal alignment and domain alignment. Best viewed in color and each color means a kind of feature.}}
\label{fig:visualization_alignment}
\end{figure*}

\noindent {\textbf{Analysis on intra-and inter-sample distribution functions.} To evaluate the power of our newly proposed distribution functions (intra-sample function $\mu(\cdot)$ and inter-sample function $\sigma(\cdot)$), we conduct an ablation study on all the tasks. Since only three losses ($\mathcal{L}_{DV}$, $\mathcal{L}_{DQ}$ and $\mathcal{L}_{M1}$) contains $\mu(\cdot)$ and $\sigma(\cdot)$, we analyze the performance of $\mu(\cdot)$ and $\sigma(\cdot)$ on these three losses. As shown in Table~\ref{tab:intra-inter}, both $\mu(\cdot)$ and $\sigma(\cdot)$ improve the performance by large margins. Particularly, on the A$\rightarrow$C task, $\mu(\cdot)$ and $\sigma(\cdot)$ of $\mathcal{L}_{DV}$ achieve the performance improvement by 2.27\% and 2.42\% in terms of ``R@1, IoU=0.3'' respectively. As for the C$\rightarrow$A task, $\mu(\cdot)$ and $\sigma(\cdot)$ of $\mathcal{L}_{M1}$ improve the retrieval performance by 2.39\% and 2.03\% ``R@1, IoU=0.7'' respectively. The significant performance shows the effectiveness of $\mu(\cdot)$ and $\sigma(\cdot)$ on the domain alignment and cross-modal alignment.}

\subsection{Do Other Multi-modal Cross-domain methods Work Well?} \label{section:other}
Someone might ask if other multi-modal cross-domain methods can also work well. To evaluate the performance of different multi-modal cross-domain methods in the VMR task, we compare our MMCDA with multiple multi-modal cross-domain methods (xMUDA \cite{jaritz2020xmuda}, MM-SADA \cite{munro2020multi}, and PMC \cite{zhang2021progressive}) on different tasks. Table~\ref{tab:ablation_xingnengduibi} reports experimental results.

Since these state-of-the-art multi-modal cross-domain methods cannot be directly used for the VMR task, we first add our pre-training loss $\mathcal{L}_{SL}$ to learn the annotation knowledge. Then, we use these methods to find some possible moment-query pairs in the target domain. Finally, we utilize our inference to obtain the target moment boundary of these methods.

From Table~\ref{tab:ablation_xingnengduibi}, for the A$\rightarrow$C and C$\rightarrow$A tasks, we can notice that our MMCDA outperforms other  methods by a large margin. {Especially, compared with the best baseline method CRMA, MMCDA raises the performance around 2.23\%, 3.92\%,  1.41\%  and 2.09\% in all metrics. The significant improvement further shows the effectiveness of  MMCDA}.

\begin{figure}[t!]
\centering
\subfloat[]{\label{fig:video-based}\includegraphics[width=0.23\textwidth]{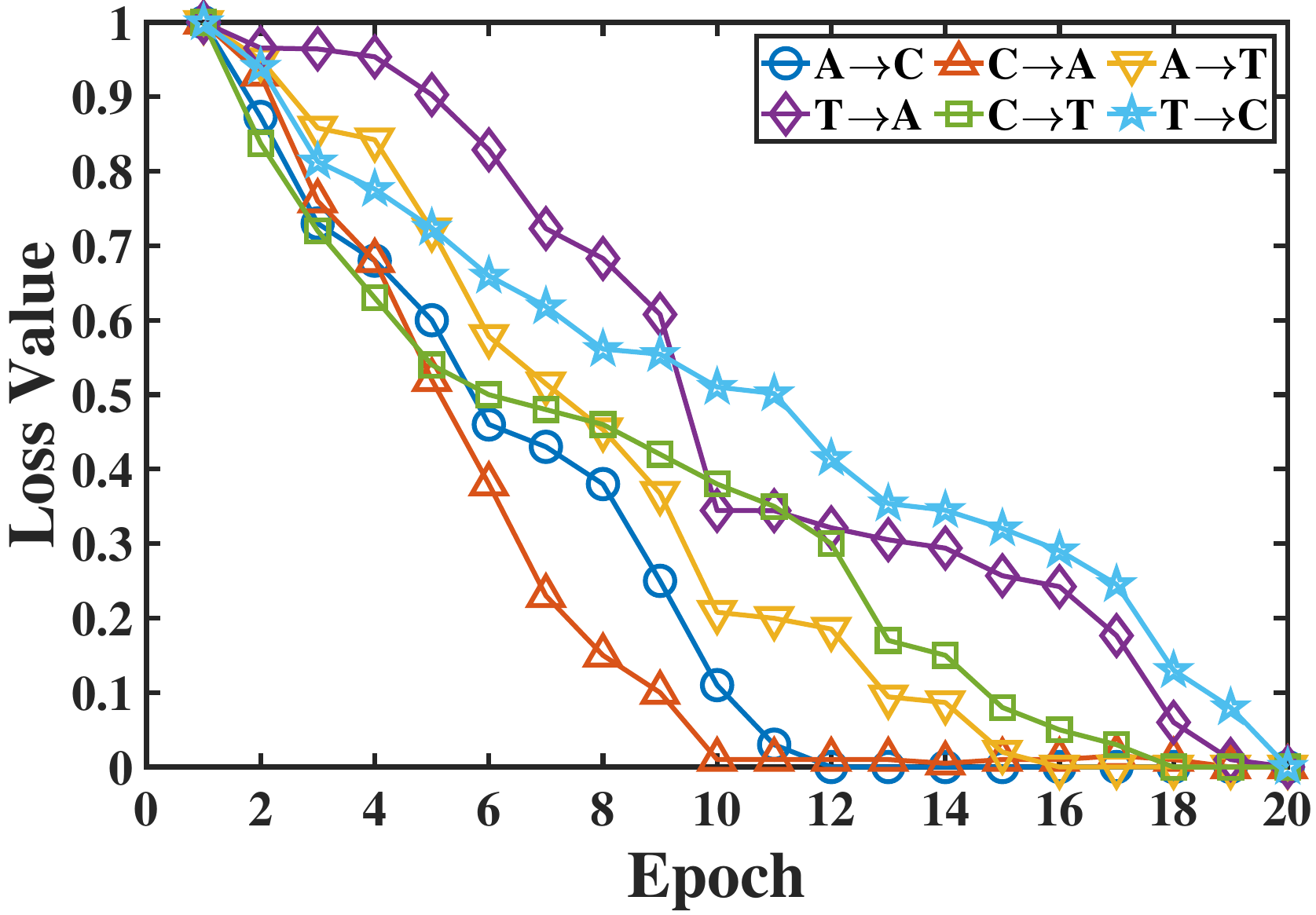} }
\subfloat[]{\label{fig:video-based}\includegraphics[width=0.23\textwidth]{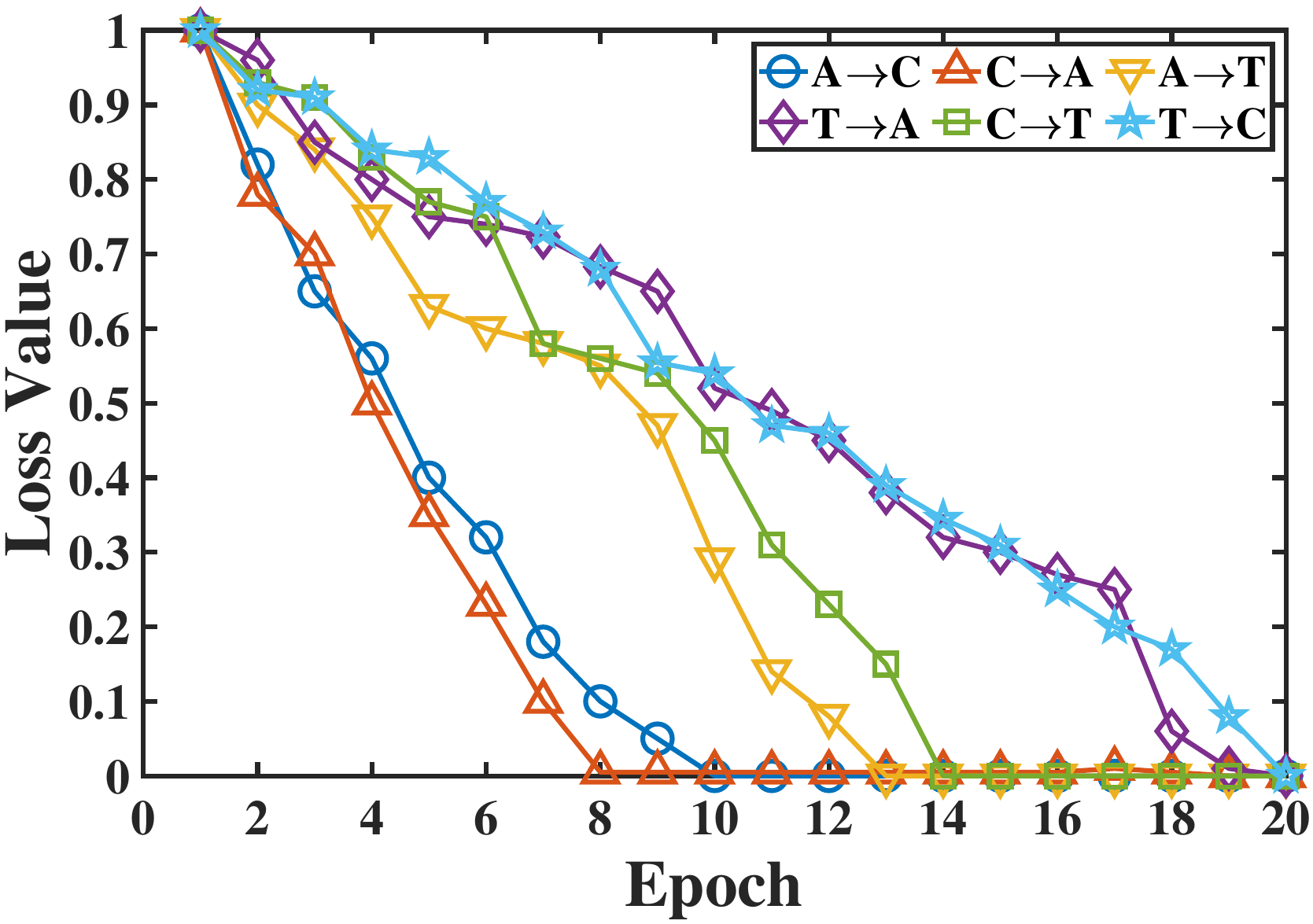}}
 \vspace{-12pt}
\caption{{Visualization of domain gap on different transfer tasks for each modality during training epochs (left: visual domain gap, right: textual domain gap). Note that we use regularized loss function values as the metric of the domain gap, where larger loss value means the larger domain gap. ``loss value=1'' means the largest domain gap and ``loss value=0'' means the smallest domain gap. Best viewed in color.}}
\label{fig:loss_value}
\end{figure}
\begin{figure}[t!]
\centering
\includegraphics[width=0.5\textwidth]{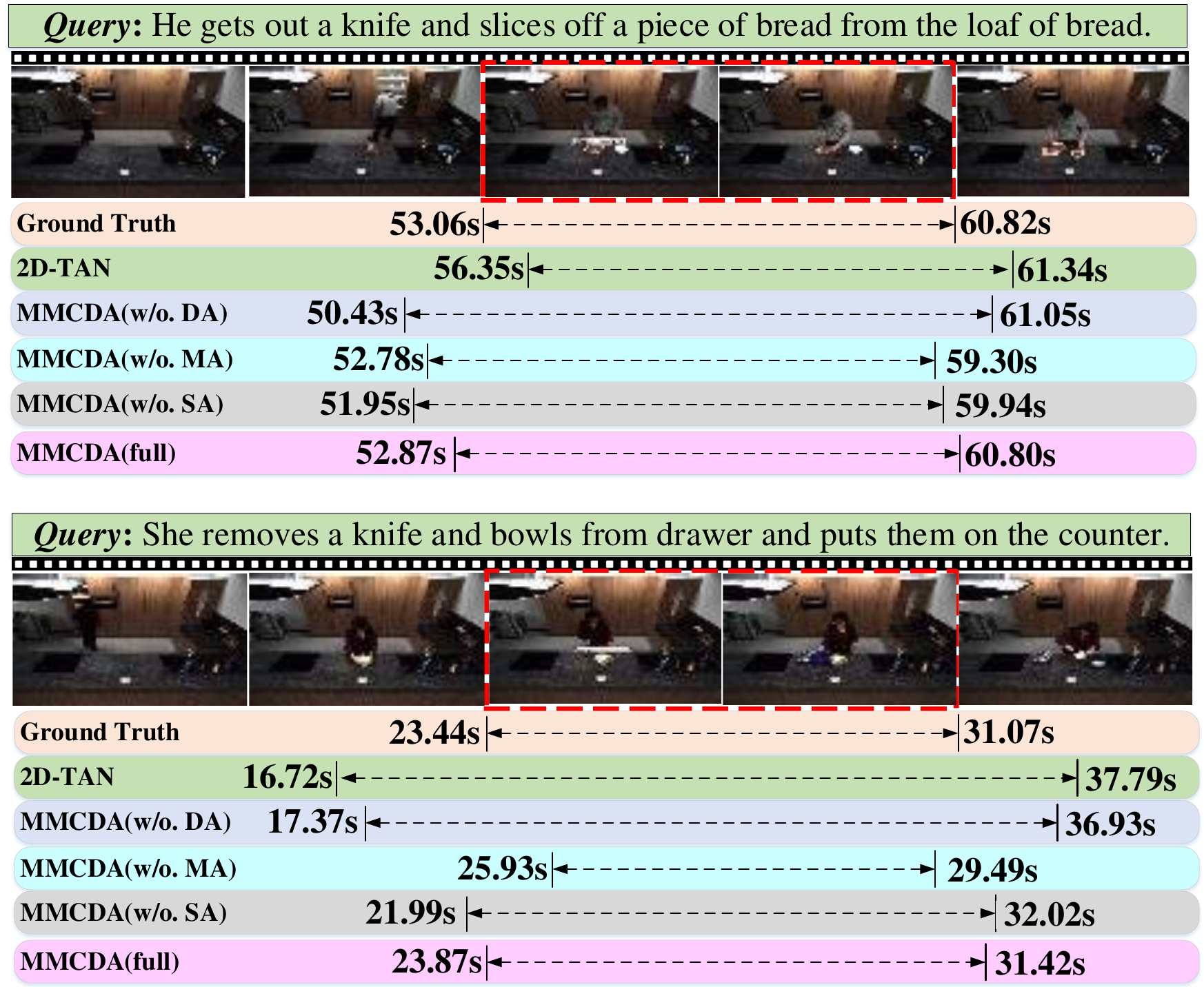}
\vspace{-25pt}
\caption{Qualitative results sampled from the TACoS dataset (top: A$\rightarrow$T, bottom: TACoS).}
\vspace{-15pt}
\label{fig:qualitative_fujia-tacos}
\end{figure}

\subsection{{Visualization on Multi-modal Cross-domain Alignment}}\label{subsection:attention_map}
{Since multi-modal cross-domain alignment is an important module, we provide the T-SNE visualization to analyze the semantic alignment between visual and textual representations. As shown in Fig.~\ref{fig:visualization_alignment}, after performing each alignment module, the corresponding features across modalities/domains are clearly clustered tighter, which illustrates the effectiveness of our each alignment loss.}

\begin{figure}[t!]
\centering
\includegraphics[width=0.5\textwidth]{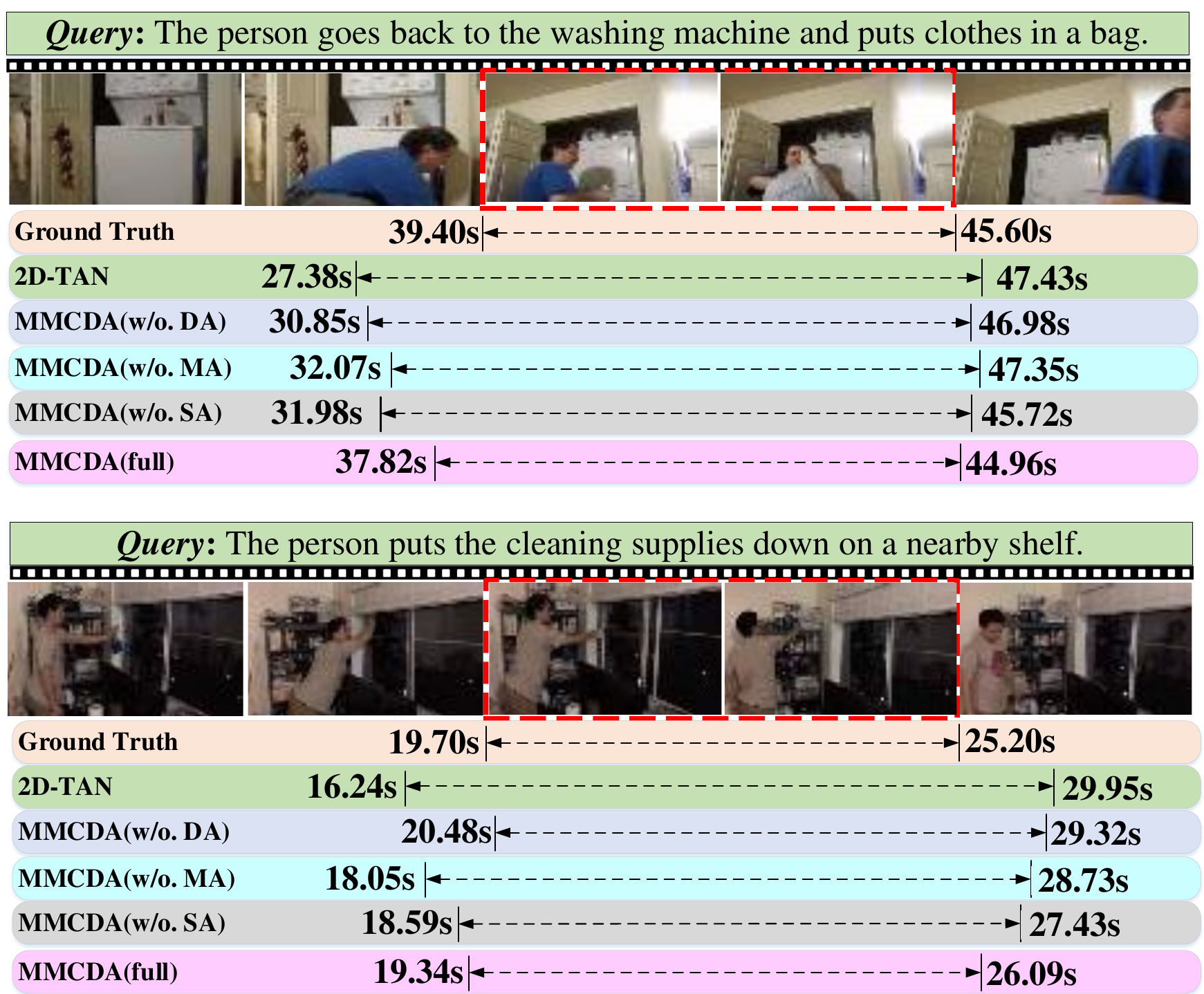}
\vspace{-23pt}
\caption{Qualitative results sampled from the Charades-STA dataset (top: A$\rightarrow$C, bottom: T$\rightarrow$C).}
\label{fig:qualitative_fujia-cha}
\end{figure}
\begin{figure}[t!]
\centering
\includegraphics[width=0.5\textwidth]{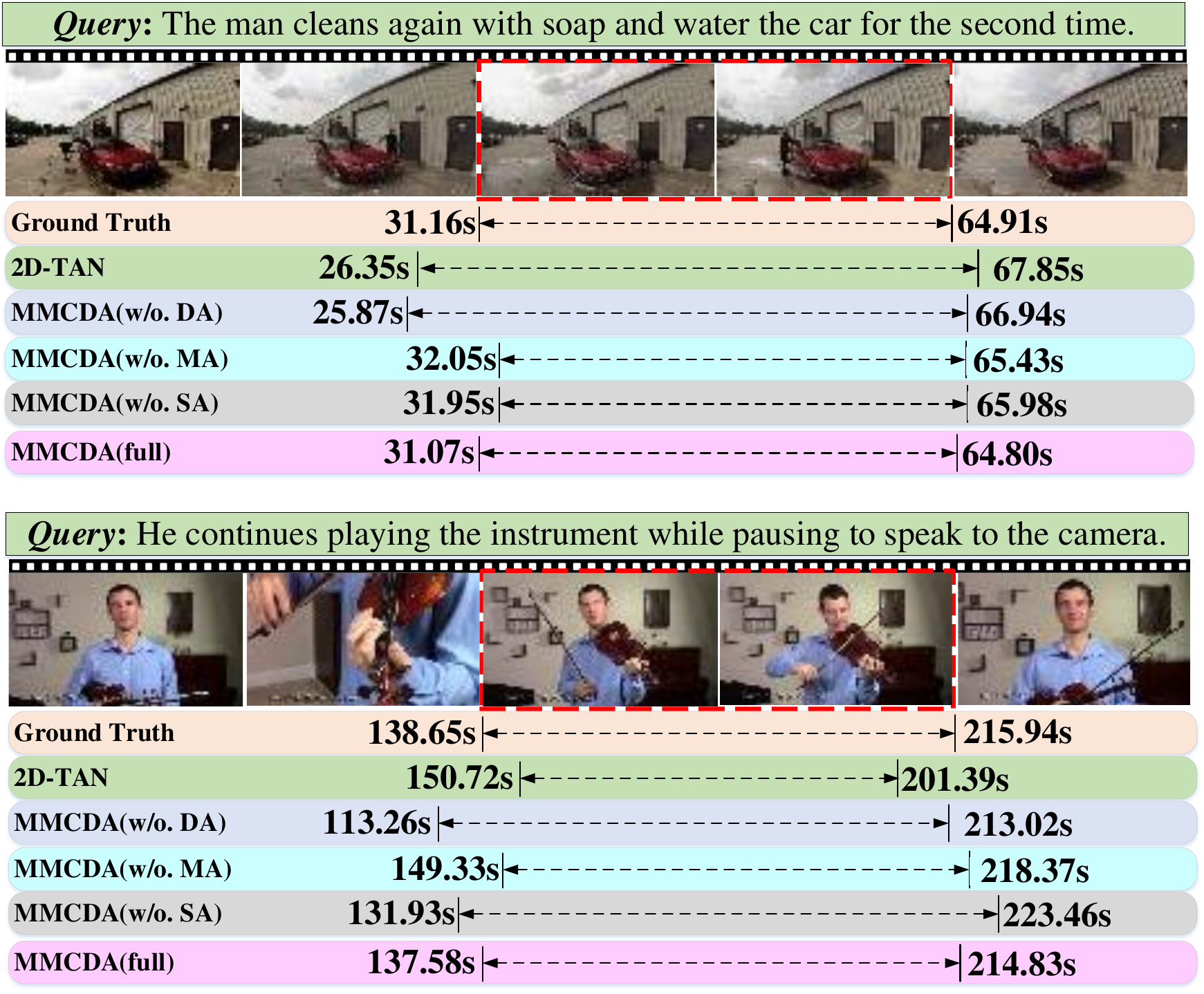}
\vspace{-25pt}
\caption{Qualitative results sampled from the ActivityNet Captions dataset (top: C$\rightarrow$A, bottom: T$\rightarrow$A).}
\label{fig:qualitative_fujia-act}
\end{figure}

{In Fig. \ref{fig:loss_value}, the faster convergence curve indicates that our method is better able to reduce the domain gap of the two datasets in the transfer task. For example, in the A$\rightarrow$C and C$\rightarrow$A tasks, their corresponding curves converge within 12 epochs (for the visual domain gap) and 10 epochs (for the textual domain gap). It is because our MMCDA can effectively and efficiently transfer the annotation knowledge from the souce dataset to the target dataset by our domain alignment module, which close the domain gap between the ActivityNet Captions and Charades-STA datasets. When we use the TACoS dataset as the source dataset, the curves in the T$\rightarrow$C and T$\rightarrow$A tasks  often converge slowly. The main reason is that the TACoS dataset contains significantly little annotation knowledge, which limits the performance on the target datasets. On the contrary, when we utilize the TACoS dataset as the target dataset, our MMCDA can obtain enough annotation knowledge by the domain alignment, which speeds the curve convergence.}

\subsection{Qualitative Analysis}\label{subsection:qualitative}

To qualitatively investigate the effectiveness of our MMCDA,  we report some representative examples from three datasets on six transfer tasks as shown in Fig.~\ref{fig:qualitative_fujia-tacos}-\ref{fig:qualitative_fujia-act}, where the ground truth and the fully-supervised method 2D-TAN are two baselines.
Obviously, our  MMCDA localizes more accurately than  2D-TAN, which verifies the effectiveness of MMCDA.

For our designed ablation models, we can find the following interesting points: (i) MMCDA(full) achieves more precise localization than each ablation model, which shows the effectiveness of each alignment module. (ii) Compared with MMCDA(w/o. DA), MMCDA(full) achieves significant improvement because MMCDA(full) leverages the domain alignment to alleviate the domain discrepancy between source and target domains. (iii) Compared with MMCDA(w/o. MA),  MMCDA(full) obtains satisfactory improvement, since MMCDA(full) utilizes the cross-modal alignment to bridge the semantic gaps between videos and queries in the target domain. (iv) Compared with MMCDA(w/o. SA), MMCDA(full) also gains better performance. One possible reason is that MMCDA(full) employs the specific alignment to obtain a more accurate moment boundary.



\section{Conclusion}  \label{section:con}
In this paper, we make the first attempt for the cross-domain video moment retrieval task. Thanks to our well-designed three alignment modules that align data distribution, video-query features, and context semantics across the source and target domains, our proposed MMCDA can easily adapt to unseen domain data without manual annotations. Extensive experiments demonstrate that MMCDA despites being cross-domain significantly outperforms state-of-the-art single-domain methods. 
In the future, we will apply MMCDA to other tasks/datasets \cite{li2020hero,lei2020tvr} to further improve its generalization.
 Meanwhile, exploring different ways to align source/target domains to improve the model is also our aimed work.

\bibliographystyle{IEEEtran}
\bibliography{egbib_new}

@article{zhang2022latent,
  title={Latent Domain Generation for Unsupervised Domain Adaptation Object Counting},
  author={Zhang, Anran and Yang, Yandan and Xu, Jun and Cao, Xianbin and Zhen, Xiantong and Shao, Ling},
  journal={IEEE TMM},
  year={2022}
}

@article{tao2022dreamt,
  title={DREAMT: Diversity Enlarged Mutual Teaching for Unsupervised Domain Adaptive Person Re-Identifcation},
  author={Tao, Yusheng and Zhang, Jian and Hong, Jiajing and Zhu, Yuesheng},
  journal={IEEE TMM},
  year={2022}
}

@article{wang2022uncertainty,
  title={Uncertainty-aware clustering for unsupervised domain adaptive object re-identification},
  author={Wang, Pengfei and Ding, Changxing and Tan, Wentao and Gong, Mingming and Jia, Kui and Tao, Dacheng},
  journal={IEEE TMM},
  year={2022}
}

@article{jing2022adversarial,
  title={Adversarial Mixup Ratio Confusion for Unsupervised Domain Adaptation},
  author={Jing, Mengmeng and Meng, Lichao and Li, Jingjing and Zhu, Lei and Shen, Heng Tao},
  journal={IEEE TMM},
  year={2022}
}

@article{wang2022information,
  title={Information Maximizing Adaptation Network With Label Distribution Priors for Unsupervised Domain Adaptation},
  author={Wang, Pei and Yang, Yun and Xia, Yuelong and Wang, Kun and Zhang, Xingyi and Wang, Song},
  journal={IEEE TMM},
  year={2022}
}

@InProceedings{GaoSYN17,
  author    = {Jiyang Gao and Chen Sun and Zhenheng Yang and Ram Nevatia},
  booktitle = {ICCV},
  title     = {{TALL:} Temporal Activity Localization via Language Query},
  year      = {2017},
  pages     = {5277--5285},
}

@InProceedings{KrishnaHRFN17,
  author    = {Ranjay Krishna and Kenji Hata and Frederic Ren and Li Fei{-}Fei and Juan Carlos Niebles},
  booktitle = {ICCV},
  title     = {Dense-Captioning Events in Videos},
  year      = {2017},
  pages     = {706--715},
}

@Article{RegneriRWTSP13,
  author  = {Michaela Regneri and Marcus Rohrbach and Dominikus Wetzel and Stefan Thater and Bernt Schiele and Manfred Pinkal},
  journal = {TACL},
  title   = {Grounding Action Descriptions in Videos},
  year    = {2013},
  pages   = {25--36},
  volume  = {1},
}

@InProceedings{SigurdssonVWFLG16,
  author    = {Gunnar A. Sigurdsson and G{\"{u}}l Varol and Xiaolong Wang and Ali Farhadi and Ivan Laptev and Abhinav Gupta},
  booktitle = {ECCV},
  title     = {Hollywood in Homes: Crowdsourcing Data Collection for Activity Understanding},
  year      = {2016},
  pages     = {510--526},
}

@InProceedings{Na_2021_CVPR,
    author    = {Na, Jaemin and Jung, Heechul and Chang, Hyung Jin and Hwang, Wonjun},
    title     = {FixBi: Bridging Domain Spaces for Unsupervised Domain Adaptation},
    booktitle = {CVPR},
    year      = {2021},
}

@InProceedings{Liu_2021_CVPR,
    author    = {Liu, Daizong and Qu, Xiaoye and Dong, Jianfeng and Zhou, Pan and Cheng, Yu and Wei, Wei and Xu, Zichuan and Xie, Yulai},
    title     = {Context-Aware Biaffine Localizing Network for Temporal Sentence Grounding},
    booktitle = {CVPR},
    month     = {June},
    year      = {2021},
    pages     = {11235-11244}
}

@inproceedings{xiao2021boundary,
  title={Boundary Proposal Network for Two-Stage Natural Language Video Localization},
  author={Xiao, Shaoning and Chen, Long and Zhang, Songyang and Ji, Wei and Shao, Jian and Ye, Lu and Xiao, Jun},
  booktitle={AAAI},
  year={2021}
}

@inproceedings{li2021proposal,
  title={Proposal-Free Video Grounding with Contextual Pyramid Network},
  author={Li, Kun and Guo, Dan and Wang, Meng},
  booktitle={AAAI},
  pages={1902--1910},
  year={2021}
}

@inproceedings{wang2021structured,
  title={Structured Multi-Level Interaction Network for Video Moment Localization via Language Query},
  author={Wang, Hao and Zha, Zheng-Jun and Li, Liang and Liu, Dong and Luo, Jiebo},
  booktitle={CVPR},
  year={2021}
}

@article{li2021semisupervised,
  title={Semisupervised Human Activity Recognition With Radar Micro-Doppler Signatures},
  author={Li, Xinyu and He, Yuan and Fioranelli, Francesco and Jing, Xiaojun},
  journal={IEEE TGRS},
  year={2021}
}

@article{song2020weakly,
  title={Weakly-supervised multi-level attentional reconstruction network for grounding textual queries in videos},
  author={Song, Yijun and Wang, Jingwen and Ma, Lin and Yu, Zhou and Yu, Jun},
  journal={arXiv preprint arXiv:2003.07048},
  year={2020}
}

@article{zhang2020counterfactual,
  title={Counterfactual Contrastive Learning for Weakly-Supervised Vision-Language Grounding},
  author={Zhang, Zhu and Zhao, Zhou and Lin, Zhijie and He, Xiuqiang and others},
  journal={NeurIPS},
  year={2020}
}

@inproceedings{zhang2021multi,
  title={Multi-Stage Aggregated Transformer Network for Temporal Language Localization in Videos},
  author={Zhang, Mingxing and Yang, Yang and Chen, Xinghan and Ji, Yanli and Xu, Xing and Li, Jingjing and Shen, Heng Tao},
  booktitle={CVPR},
  pages={12669--12678},
  year={2021}
}

@inproceedings{zhao2021cascaded,
  title={Cascaded Prediction Network via Segment Tree for Temporal Video Grounding},
  author={Zhao, Yang and Zhao, Zhou and Zhang, Zhu and Lin, Zhijie},
  booktitle={CVPR},
  pages={4197--4206},
  year={2021}
}

@inproceedings{rodriguez2020proposal,
  title={Proposal-free temporal moment localization of a natural-language query in video using guided attention},
  author={Rodriguez, Cristian and Marrese-Taylor, Edison and Saleh, Fatemeh Sadat and Li, Hongdong and Gould, Stephen},
  booktitle={WACV},
  year={2020}
}

@InProceedings{ZhangSJZ20,
  author    = {Hao Zhang and Aixin Sun and Wei Jing and Joey Tianyi Zhou},
  booktitle = {ACL},
  title     = {Span-based Localizing Network for Natural Language Video Localization},
  year      = {2020},
  pages     = {6543--6554},
}

@InProceedings{LiuWN0CC18,
  author    = {Meng Liu and Xiang Wang and Liqiang Nie and Xiangnan He and Baoquan Chen and Tat{-}Seng Chua},
  booktitle = {SIGIR},
  title     = {Attentive Moment Retrieval in Videos},
  year      = {2018},
  pages     = {15--24},
}

@InProceedings{ge2019mac,
  author    = {Runzhou {Ge} and Jiyang {Gao} and Kan {Chen} and Ram {Nevatia}},
  booktitle = {WACV},
  title     = {MAC: Mining Activity Concepts for Language-Based Temporal Localization},
  year      = {2019},
  pages     = {245--253},
}

@InProceedings{MithunPR19,
  author    = {Niluthpol Chowdhury Mithun and Sujoy Paul and Amit K. Roy{-}Chowdhury},
  booktitle = {CVPR},
  title     = {Weakly Supervised Video Moment Retrieval From Text Queries},
  year      = {2019}
}

@inproceedings{hendricks2018localizing,
  title={Localizing Moments in Video with Temporal Language},
  author={Hendricks, Lisa Anne and Wang, Oliver and Shechtman, Eli and Sivic, Josef and Darrell, Trevor and Russell, Bryan},
  booktitle={EMNLP},
  year={2018}
}

@InProceedings{ChenKN17,
  author    = {Kan Chen and Rama Kovvuri and Ram Nevatia},
  booktitle = {ICCV},
  title     = {Query-Guided Regression Network with Context Policy for Phrase Grounding},
  year      = {2017},
  pages     = {824--832},
}

@InProceedings{Yu0SYLBB18,
  author    = {Licheng Yu and Zhe Lin and Xiaohui Shen and Jimei Yang and Xin Lu and Mohit Bansal and Tamara L. Berg},
  booktitle = {CVPR},
  title     = {MAttNet: Modular Attention Network for Referring Expression Comprehension},
  year      = {2018},
  pages     = {1307--1315},
}

@InProceedings{ChoMGBBSB14,
  author    = {Kyunghyun Cho and Bart van Merrienboer and {\c{C}}aglar G{\"{u}}l{\c{c}}ehre and Dzmitry Bahdanau and Fethi Bougares and Holger Schwenk and Yoshua Bengio},
  booktitle = {EMNLP},
  title     = {Learning Phrase Representations using {RNN} Encoder-Decoder for Statistical Machine Translation},
  year      = {2014},
  pages     = {1724--1734},
}

@InProceedings{ZhangLZX19,
  author    = {Zhu Zhang and Zhijie Lin and Zhou Zhao and Zhenxin Xiao},
  booktitle = {SIGIR},
  title     = {Cross-Modal Interaction Networks for Query-Based Moment Retrieval in Videos},
  year      = {2019},
  pages     = {655--664},
}

@InProceedings{LiuQLDZX20,
  author    = {Daizong Liu and Xiaoye Qu and Xiao{-}Yang Liu and Jianfeng Dong and Pan Zhou and Zichuan Xu},
  booktitle = {MM},
  title     = {Jointly Cross- and Self-Modal Graph Attention Network for Query-Based Moment Localization},
  year      = {2020},
  pages     = {4070--4078},
}

@InProceedings{DuanHGW0H18,
  author    = {Xuguang Duan and Wen{-}bing Huang and Chuang Gan and Jingdong Wang and Wenwu Zhu and Junzhou Huang},
  booktitle = {NeurIPS},
  title     = {Weakly Supervised Dense Event Captioning in Videos},
  year      = {2018}
}

@InProceedings{LinZZWL20,
  author    = {Zhijie Lin and Zhou Zhao and Zhu Zhang and Qi Wang and Huasheng Liu},
  booktitle = {AAAI},
  title     = {Weakly-Supervised Video Moment Retrieval via Semantic Completion Network},
  year      = {2020}
}

@InProceedings{WangHW19,
  author    = {Weining Wang and Yan Huang and Liang Wang},
  booktitle = {CVPR},
  title     = {Language-Driven Temporal Activity Localization: {A} Semantic Matching Reinforcement Learning Model},
  year      = {2019},
}

@InProceedings{Xu0PSSS19,
  author    = {Huijuan Xu and Kun He and Bryan A. Plummer and Leonid Sigal and Stan Sclaroff and Kate Saenko},
  booktitle = {AAAI},
  title     = {Multilevel Language and Vision Integration for Text-to-Clip Retrieval},
  year      = {2019},
  pages     = {9062--9069},
}

@InProceedings{ZhangDWWD19,
  author    = {Da Zhang and Xiyang Dai and Xin Wang and Yuan{-}Fang Wang and Larry S. Davis},
  booktitle = {CVPR},
  title     = {{MAN:} Moment Alignment Network for Natural Language Moment Retrieval via Iterative Graph Adjustment},
  year      = {2019},
  pages     = {1247--1257},
}

@InProceedings{ChenJ19a,
  author    = {Shaoxiang Chen and Yu{-}Gang Jiang},
  booktitle = {AAAI},
  title     = {Semantic Proposal for Activity Localization in Videos via Sentence Query},
  year      = {2019},
  pages     = {8199--8206},
}

@InProceedings{ZhangPFL20,
  author    = {Songyang Zhang and Houwen Peng and Jianlong Fu and Jiebo Luo},
  booktitle = {AAAI},
  title     = {Learning 2D Temporal Adjacent Networks for Moment Localization with Natural Language},
  year      = {2020},
}

@InProceedings{ChenMLW19,
  author    = {Zhenfang Chen and Lin Ma and Wenhan Luo and Kwan{-}Yee Kenneth Wong},
  booktitle = {ACL},
  title     = {Weakly-Supervised Spatio-Temporally Grounding Natural Sentence in Video},
  year      = {2019},
}

@InProceedings{ZhangLZZH20,
  author    = {Zhu Zhang and Zhijie Lin and Zhou Zhao and Jieming Zhu and Xiuqiang He},
  booktitle = {MM},
  title     = {Regularized Two-Branch Proposal Networks for Weakly-Supervised Moment Retrieval in Videos},
  year      = {2020},
  pages     = {4098--4106},
}

@InProceedings{2021LoGAN,
  author    = {Tan, Reuben and Xu, Huijuan and Saenko, Kate and Plummer, Bryan A.},
  booktitle = {WACV},
  title     = {LoGAN: Latent Graph Co-Attention Network for Weakly-Supervised Video Moment Retrieval},
  year      = {2021},
}

@InProceedings{MaYKLKY20,
  author    = {Minuk Ma and Sunjae Yoon and Junyeong Kim and Youngjoon Lee and Sunghun Kang and Chang D. Yoo},
  booktitle = {ECCV},
  title     = {VLANet: Video-Language Alignment Network for Weakly-Supervised Video Moment Retrieval},
  year      = {2020},
  pages     = {156--171},
}

@InProceedings{YuanMWL019,
  author    = {Yitian Yuan and Lin Ma and Jingwen Wang and Wei Liu and Wenwu Zhu},
  booktitle = {NeurIPS},
  title     = {Semantic Conditioned Dynamic Modulation for Temporal Sentence Grounding in Videos},
  year      = {2019},
}

@InProceedings{LiuQDZ20,
  author    = {Daizong Liu and Xiaoye Qu and Jianfeng Dong and Pan Zhou},
  booktitle = {COLING},
  title     = {Reasoning Step-by-Step: Temporal Sentence Localization in Videos via Deep Rectification-Modulation Network},
  year      = {2020},
  pages     = {1841--1851},
}

@InProceedings{TzengHSD17,
  author    = {Eric Tzeng and Judy Hoffman and Kate Saenko and Trevor Darrell},
  booktitle = {CVPR},
  title     = {Adversarial Discriminative Domain Adaptation},
  year      = {2017},
  pages     = {2962--2971},
}

@InProceedings{Zhong0LL019,
  author    = {Zhun Zhong and Liang Zheng and Zhiming Luo and Shaozi Li and Yi Yang},
  booktitle = {CVPR},
  title     = {Invariance Matters: Exemplar Memory for Domain Adaptive Person Re-Identification},
  year      = {2019},
}

@InProceedings{TranBFTP15,
  author    = {Du Tran and Lubomir D. Bourdev and Rob Fergus and Lorenzo Torresani and Manohar Paluri},
  booktitle = {ICCV},
  title     = {Learning Spatiotemporal Features with 3D Convolutional Networks},
  year      = {2015},
}

@InProceedings{VaswaniSPUJGKP17,
  author    = {Ashish Vaswani and Noam Shazeer and Niki Parmar and Jakob Uszkoreit and Llion Jones and Aidan N. Gomez and Lukasz Kaiser and Illia Polosukhin},
  booktitle = {NIPS},
  title     = {Attention is All you Need},
  year      = {2017},
}

@InProceedings{PenningtonSM14,
  author    = {Jeffrey Pennington and Richard Socher and Christopher D. Manning},
  booktitle = {EMNLP},
  title     = {Glove: Global Vectors for Word Representation},
  year      = {2014},
  pages     = {1532--1543},
}

@InProceedings{chung2014empirical,
  author    = {Chung, Junyoung and Gulcehre, Caglar and Cho, KyungHyun and Bengio, Yoshua},
  booktitle = {NIPS},
  title     = {Empirical evaluation of gated recurrent neural networks on sequence modeling},
  year      = {2014},
}

@Article{GrettonBRSS12,
  author  = {Arthur Gretton and Karsten M. Borgwardt and Malte J. Rasch and Bernhard Scholkopf and Alexander J. Smola},
  journal = {JMLR},
  title   = {A Kernel Two-Sample Test},
  year    = {2012},
  pages   = {723--773},
  volume  = {13},
}

@article{guan2021uncertainty,
  title={Uncertainty-aware unsupervised domain adaptation in object detection},
  author={Guan, Dayan and Huang, Jiaxing and Xiao, Aoran and Lu, Shijian and Cao, Yanpeng},
  journal={IEEE TMM},
  year={2021}
}

@article{zhang2020temporal,
  title={Temporal textual localization in video via adversarial bi-directional interaction networks},
  author={Zhang, Zijian and Zhao, Zhou and Zhang, Zhu and Lin, Zhijie and Wang, Qi and Hong, Richang},
  journal={IEEE TMM},
  volume={23},
  pages={3306--3317},
  year={2020}
}

@article{yang2022video,
  title={Video Moment Retrieval with Cross-Modal Neural Architecture Search},
  author={Yang, Xun and Wang, Shanshan and Dong, Jian and Dong, Jianfeng and Wang, Meng and Chua, Tat-Seng},
  journal={IEEE TIP},
  year={2022}
}

@article{zhao2021cross,
          title={Cross-Domain Image Captioning via Cross-Modal Retrieval and Model Adaptation}, 
          author={Wentian Zhao and Xinxiao Wu and Jiebo Luo},
          journal={IEEE TIP}, 
          year={2021},
          volume={30},
          pages={1180-1192}
        }

@inproceedings{silva2021embracing,
  title={Embracing domain differences in fake news: Cross-domain fake news detection using multi-modal data},
  author={Silva, Amila and Luo, Ling and Karunasekera, Shanika and Leckie, Christopher},
  booktitle={AAAI},
  volume={35},
  number={1},
  pages={557--565},
  year={2021}
}

@inproceedings{zolfaghari2021crossclr,
  title={Crossclr: Cross-modal contrastive learning for multi-modal video representations},
  author={Zolfaghari, Mohammadreza and Zhu, Yi and Gehler, Peter and Brox, Thomas},
  booktitle={ICCV},
  pages={1450--1459},
  year={2021}
}

@article{liu2022unsupervised,
  title={Unsupervised temporal video grounding with deep semantic clustering},
  author={Liu, Daizong and Qu, Xiaoye and Wang, Yinzhen and Di, Xing and Zou, Kai and Cheng, Yu and Xu, Zichuan and Zhou, Pan},
  journal={arXiv preprint arXiv:2201.05307},
  year={2022}
}

@article{wang2022multi,
  title={Multi-Modal Domain Adaptation Variational Autoencoder for EEG-Based Emotion Recognition},
  author={Wang, Yixin and Qiu, Shuang and Li, Dan and Du, Changde and Lu, Bao-Liang and He, Huiguang},
  journal={JAS},
  year={2022}
}

@article{hu2021coarse,
  title={Coarse-to-fine semantic alignment for cross-modal moment localization},
  author={Hu, Yupeng and Nie, Liqiang and Liu, Meng and Wang, Kun and Wang, Yinglong and Hua, Xian-Sheng},
  journal={IEEE TIP},
  volume={30},
  pages={5933--5943},
  year={2021}
}

@article{sun2021maban,
  title={MABAN: Multi-agent boundary-aware network for natural language moment retrieval},
  author={Sun, Xiaoyang and Wang, Hanli and He, Bin},
  journal={IEEE TIP},
  volume={30},
  pages={5589--5599},
  year={2021}
}

@inproceedings{munro2020multi,
  title={Multi-modal domain adaptation for fine-grained action recognition},
  author={Munro, Jonathan and Damen, Dima},
  booktitle={CVPR},
  pages={122--132},
  year={2020}
}

@inproceedings{chen2021mind,
  title={Mind-the-Gap! Unsupervised Domain Adaptation for Text-Video Retrieval},
  author={Chen, Qingchao and Liu, Yang and Albanie, Samuel},
  booktitle={AAAI},
  pages={1072--1080},
  year={2021}
}

@inproceedings{kim2021learning,
  title={Learning cross-modal contrastive features for video domain adaptation},
  author={Kim, Donghyun and Tsai, Yi-Hsuan and Zhuang, Bingbing and Yu, Xiang and Sclaroff, Stan and Saenko, Kate and Chandraker, Manmohan},
  booktitle={ICCV},
  pages={13618--13627},
  year={2021}
}

@inproceedings{song2021spatio,
  title={Spatio-temporal contrastive domain adaptation for action recognition},
  author={Song, Xiaolin and Zhao, Sicheng and Yang, Jingyu and Yue, Huanjing and Xu, Pengfei and Hu, Runbo and Chai, Hua},
  booktitle={CVPR},
  pages={9787--9795},
  year={2021}
}

@article{zeng2022moment,
  title={Moment is Important: Language-Based Video Moment Retrieval via Adversarial Learning},
  author={Zeng, Yawen and Cao, Da and Lu, Shaofei and Zhang, Hanling and Xu, Jiao and Qin, Zheng},
  journal={ACM TOMM},
  volume={18},
  number={2},
  pages={1--21},
  year={2022},
}

@article{wang2022cross,
  title={Cross-modal Dynamic Networks for Video Moment Retrieval with Text Query},
  author={Wang, Gongmian and Xu, Xing and Shen, Fumin and Lu, Huimin and Ji, Yanli and Shen, Heng Tao},
  journal={IEEE TMM},
  year={2022}
}

@article{zhang2021natural,
  title={Natural language video localization: A revisit in span-based question answering framework},
  author={Zhang, Hao and Sun, Aixin and Jing, Wei and Zhen, Liangli and Zhou, Joey Tianyi and Goh, Rick Siow Mong},
  journal={IEEE TPAMI},
  year={2021},
}

@article{tang2021frame,
  title={Frame-wise Cross-modal Matching for Video Moment Retrieval},
  author={Tang, Haoyu and Zhu, Jihua and Liu, Meng and Gao, Zan and Cheng, Zhiyong},
  journal={IEEE TMM},
  year={2021},
}

@InProceedings{zhu2020vision,
  author    = {Zhu, Fengda and Zhu, Yi and Chang, Xiaojun and Liang, Xiaodan},
  booktitle = {CVPR},
  title     = {Vision-language navigation with self-supervised auxiliary reasoning tasks},
  year      = {2020},
  pages     = {10012--10022},
}

@article{long2018conditional,
  title={Conditional adversarial domain adaptation},
  author={Long, Mingsheng and Cao, Zhangjie and Wang, Jianmin and Jordan, Michael I},
  journal={NeurIPS},
  volume={31},
  year={2018}
}

@inproceedings{long2018deep,
  title={Deep domain adaptation hashing with adversarial learning},
  author={Long, Fuchen and Yao, Ting and Dai, Qi and Tian, Xinmei and Luo, Jiebo and Mei, Tao},
  booktitle={ACM SIGIR},
  pages={725--734},
  year={2018}
}

@inproceedings{ganin2015unsupervised,
  title={Unsupervised domain adaptation by backpropagation},
  author={Ganin, Yaroslav and Lempitsky, Victor},
  booktitle={ICML},
  pages={1180--1189},
  year={2015}
}

@Article{ganin2016domain,
  author  = {Ganin, Yaroslav and Ustinova, Evgeniya and Ajakan, Hana and Germain, Pascal and Larochelle, Hugo and Laviolette, Fran{\c{c}}ois and Marchand, Mario and Lempitsky, Victor},
  journal = {JMLR},
  title   = {Domain-adversarial training of neural networks},
  year    = {2016},
  number  = {1},
  pages   = {2096--2030},
  volume  = {17},
}

@InProceedings{hosseini2018augmented,
  author    = {Hosseini-Asl, Ehsan and Zhou, Yingbo and Xiong, Caiming and Socher, Richard},
  booktitle = {ICLR},
  title     = {Augmented Cyclic Adversarial Learning for Low Resource Domain Adaptation},
  year      = {2018},
}

@Article{huang2018mhtn,
  author  = {Huang, Xin and Peng, Yuxin and Yuan, Mingkuan},
  journal = {IEEE TCYB},
  title   = {Mhtn: Modal-adversarial hybrid transfer network for cross-modal retrieval},
  year    = {2018},
  pages   = {1047--1059},
}

@InProceedings{wang2020temporally,
  author    = {Wang, Jingwen and Ma, Lin and Jiang, Wenhao},
  booktitle = {AAAI},
  title     = {Temporally grounding language queries in videos by contextual boundary-aware prediction},
  year      = {2020},
}

@InProceedings{yuan2019find,
  author    = {Yuan, Yitian and Mei, Tao and Zhu, Wenwu},
  booktitle = {AAAI},
  title     = {To find where you talk: Temporal sentence localization in video with attention based location regression},
  year      = {2019},
}

@inproceedings{li2020hero,
  title={HERO: Hierarchical Encoder for Video+ Language Omni-representation Pre-training},
  author={Li, Linjie and Chen, Yen-Chun and Cheng, Yu and Gan, Zhe and Yu, Licheng and Liu, Jingjing},
  booktitle={EMNLP},
  year={2020}
}

@article{zhang2021progressive,
  title={Progressive modality cooperation for multi-modality domain adaptation},
  author={Zhang, Weichen and Xu, Dong and Zhang, Jing and Ouyang, Wanli},
  journal={IEEE TIP},
  volume={30},
  pages={3293--3306},
  year={2021}
}

@inproceedings{jaritz2020xmuda,
  title={xmuda: Cross-modal unsupervised domain adaptation for 3d semantic segmentation},
  author={Jaritz, Maximilian and Vu, Tuan-Hung and Charette, Raoul de and Wirbel, Emilie and P{\'e}rez, Patrick},
  booktitle={CVPR},
  pages={12605--12614},
  year={2020}
}

@inproceedings{lei2020tvr,
  title={TVR: A Large-Scale Dataset for Video-Subtitle Moment Retrieval},
  author={Lei, Jie and Yu, Licheng and Berg, Tamara L and Bansal, Mohit},
  booktitle={ECCV},
  year={2020}
}

@InProceedings{zeng2021multi,
  author    = {Zeng, Yawen and Cao, Da and Wei, Xiaochi and Liu, Meng and Zhao, Zhou and Qin, Zheng},
  booktitle = {CVPR},
  title     = {Multi-Modal Relational Graph for Cross-Modal Video Moment Retrieval},
  year      = {2021},
}

@InProceedings{cao2020strong,
  author    = {Cao, Da and Zeng, Yawen and Liu, Meng and He, Xiangnan and Wang, Meng and Qin, Zheng},
  booktitle = {MM},
  title     = {Strong: Spatio-temporal reinforcement learning for cross-modal video moment localization},
  year      = {2020},
  pages     = {4162--4170},
}

@Article{wang2021robust,
  author  = {Wang, Zhikang and He, Lihuo and Tu, Xiaoguang and Zhao, Jian and Gao, Xinbo and Shen, Shengmei and Feng, Jiashi},
  journal = {IEEE TCSVT},
  title   = {Robust Video-based Person Re-Identification by Hierarchical Mining},
  year    = {2021},
}

@article{fang2025your,
  title={Your data is not perfect: Towards cross-domain out-of-distribution detection in class-imbalanced data},
  author={Fang, Xiang and Easwaran, Arvind and Genest, Blaise and Suganthan, Ponnuthurai Nagaratnam},
  journal={Expert Systems with Applications},
  year={2025}
}

@article{fang2023hierarchical,
  title={Hierarchical local-global transformer for temporal sentence grounding},
  author={Fang, Xiang and Liu, Daizong and Zhou, Pan and Xu, Zichuan and Li, Ruixuan},
  journal={IEEE Transactions on Multimedia},
  year={2023},
  publisher={IEEE}
}

@inproceedings{fang2026cogniVerse,
  title={CogniVerse: Revolutionizing Multi-modal Retrieval-Augmented Generation with Cognitive Reflection and Geometric Reasoning},
  author={Fang, Xiang and Fang, Wanlong and Wang, Changshuo},
  booktitle={Proceedings of the IEEE/CVF Conference on Computer Vision and Pattern Recognition},
  year={2026}
}

@inproceedings{fang2023you,
  title={You can ground earlier than see: An effective and efficient pipeline for temporal sentence grounding in compressed videos},
  author={Fang, Xiang and Liu, Daizong and Zhou, Pan and Nan, Guoshun},
  booktitle={Proceedings of the IEEE/CVF Conference on Computer Vision and Pattern Recognition},
  pages={2448--2460},
  year={2023}
}

@inproceedings{fang2025hierarchical,
  title={Hierarchical Semantic-Augmented Navigation: Optimal Transport and Graph-Driven Reasoning for Vision-Language Navigation},
  author={Fang, Xiang and Fang, Wanlong and Wang, Changshuo},
  booktitle={Advances in Neural Information Processing Systems},
  year={2025}
}

@inproceedings{fang2025adaptive,
  title={Adaptive Multi-prompt Contrastive Network for Few-shot Out-of-distribution Detection},
  author={Fang, Xiang and Easwaran, Arvind and Genest, Blaise},
  booktitle={International Conference on Machine Learning},
  year={2025}
}

@inproceedings{fang2026slap,
  title={SLAP: The Semantic Least Action Principle for Variational Video-Language Modeling},
  author={Fang, Xiang and Fang, Wanlong},
  booktitle={International Conference on Machine Learning},
  year={2026}
}

@inproceedings{fang2026immuno,
  title={Immuno-VLM: Immunizing Large Vision-Language Models via Generative Semantic Antibodies for Open-World Trustworthiness},
  author={Fang, Xiang and Fang, Wanlong and Ji, Wei},
  booktitle={International Conference on Machine Learning},
  year={2026}
}

@inproceedings{fang2026disentangling,
  title={Disentangling Adversarial Prompts: A Semantic-Graph Defense for Robust LLM Security},
  author={Fang, Xiang and Fang, Wanlong},
 booktitle={Proceedings of the AAAI Conference on Artificial Intelligence},
year={2026}
}

@inproceedings{fang2026advancing,
  title={Advancing Out-of-Distribution Detection Across Diverse Scenarios},
  author={Fang, Xiang},
  booktitle={Proceedings of the AAAI Conference on Artificial Intelligence},
  volume={40},
  number={48},
  pages={41042--41043},
  year={2026}
}

@inproceedings{fang2026unveiling,
  title={Unveiling the Fragility of Vision-Language Models: Multi-Modal Adversarial Synergy via Texture-Constrained Perturbations and Cross-Modal Optimization},
  author={Fang, Xiang and Fang, Wanlong and Wang, Changshuo},
 booktitle={Proceedings of the AAAI Conference on Artificial Intelligence},
year={2026}
}

@inproceedings{fang2026rethinking,
  title={Rethinking Video-language Model From the Language Input Perspective},
  author={Fang, Xiang and Fang, Wanlong and Wang, Changshuo and Qu, Xiaoye and Liu, Daizong},
 booktitle={Proceedings of the AAAI Conference on Artificial Intelligence},
year={2026}
}

@inproceedings{fang2026towards,
  title={Towards Unified Vision-Language Models With Incomplete Multi-Modal Inputs},
  author={Fang, Xiang and Fang, Wanlong and Wang, Changshuo and Tang, Keke and Liu, Daizong and Wang, Siyi and Ji, Wei},
 booktitle={Proceedings of the AAAI Conference on Artificial Intelligence},
year={2026}
}

@inproceedings{fang2025multi,
  title={Multi-pair temporal sentence grounding via multi-thread knowledge transfer network},
  author={Fang, Xiang and Fang, Wanlong and Wang, Changshuo and Liu, Daizong and Tang, Keke and Dong, Jianfeng and Zhou, Pan and Li, Beibei},
  booktitle={Proceedings of the AAAI Conference on Artificial Intelligence},
  volume={39},
  number={3},
  pages={2915--2923},
  year={2025}
}

@inproceedings{fang2024fewer,
  title={Fewer Steps, Better Performance: Efficient Cross-Modal Clip Trimming for Video Moment Retrieval Using Language},
  author={Fang, Xiang and Liu, Daizong and Fang, Wanlong and Zhou, Pan and Xu, Zichuan and Xu, Wenzheng and Chen, Junyang and Li, Renfu},
  booktitle={Proceedings of the AAAI Conference on Artificial Intelligence},
  volume={38},
  number={2},
  pages={1735--1743},
  year={2024}
}

@inproceedings{fang2024multi,
  title={Multi-Pair Temporal Sentence Grounding via Multi-Thread Knowledge Transfer Network},
  author={Fang, Xiang and Fang, Wanlong and Wang, Changshuo and Liu, Daizong and Tang, Keke and Dong, Jianfeng and Zhou, Pan and Li, Beibei},
  booktitle={Proceedings of the AAAI Conference on Artificial Intelligence},
  year={2025}
}

@inproceedings{fang2025turing,
  title={Turing Patterns for Multimedia: Reaction-Diffusion Multi-Modal Fusion for Language-Guided Video Moment Retrieval},
  author={Fang, Xiang and Fang, Wanlong and Ji, Wei and Chua, Tat-Seng},
  booktitle={ACM International Conference on Multimedia},
  year={2025}
}

@inproceedings{fang2024not,
  title={Not all inputs are valid: Towards open-set video moment retrieval using language},
  author={Fang, Xiang and Fang, Wanlong and Liu, Daizong and Qu, Xiaoye and Dong, Jianfeng and Zhou, Pan and Li, Renfu and Xu, Zichuan and Chen, Lixing and Zheng, Panpan and others},
  booktitle={Proceedings of the 32nd ACM International Conference on Multimedia},
  pages={28--37},
  year={2024}
}

@inproceedings{fang2024rethinking,
  title={Rethinking Weakly-supervised Video Temporal Grounding From a Game Perspective},
  author={Fang, Xiang and Xiong, Zeyu and Fang, Wanlong and Qu, Xiaoye and Chen, Chen and Dong, Jianfeng and Tang, Keke and Zhou, Pan and Cheng, Yu and Liu, Daizong},
  booktitle={European Conference on Computer Vision},
  year={2024},
  organization={Springer}
}

@inproceedings{fang2023annotations,
  title={Annotations Are Not All You Need: A Cross-modal Knowledge Transfer Network for Unsupervised Temporal Sentence Grounding},
  author={Fang, Xiang and Liu, Daizong and Fang, Wanlong and Zhou, Pan and Cheng, Yu and Tang, Keke and Zou, Kai},
  booktitle={Findings of the Association for Computational Linguistics: EMNLP 2023},
  pages={8721--8733},
  year={2023}
}

@article{fang2021unbalanced,
  title={Unbalanced incomplete multi-view clustering via the scheme of view evolution: Weak views are meat; strong views do eat},
  author={Fang, Xiang and Hu, Yuchong and Zhou, Pan and Wu, Dapeng Oliver},
  journal={IEEE Transactions on Emerging Topics in Computational Intelligence},
  volume={6},
  number={4},
  pages={913--927},
  year={2021},
  publisher={IEEE}
}

@article{fang2025adaptivetai,
  title={Adaptive Hierarchical Graph Cut for Multi-granularity Out-of-distribution Detection},
  author={Fang, Xiang and Easwaran, Arvind and Genest, Blaise and Suganthan, Ponnuthurai Nagaratnam},
  journal={IEEE Transactions on Artificial Intelligence},
  year={2025}
}

@article{fang2021animc,
  title={Animc: A soft approach for autoweighted noisy and incomplete multiview clustering},
  author={Fang, Xiang and Hu, Yuchong and Zhou, Pan and Wu, Dapeng},
  journal={IEEE Transactions on Artificial Intelligence},
  volume={3},
  number={2},
  pages={192--206},
  year={2021},
  publisher={IEEE}
}

@article{fang2020v,
  title={V3H: View variation and view heredity for incomplete multiview clustering},
  author={Fang, Xiang and Hu, Yuchong and Zhou, Pan and Wu, Dapeng Oliver},
  journal={IEEE Transactions on Artificial Intelligence},
  volume={1},
  number={3},
  pages={233--247},
  year={2020},
  publisher={IEEE}
}

@article{fang2020double,
  title={Double self-weighted multi-view clustering via adaptive view fusion},
  author={Fang, Xiang and Hu, Yuchong},
  journal={arXiv preprint arXiv:2011.10396},
  year={2020}
}

@article{liu2023exploring,
  title={Exploring optical-flow-guided motion and detection-based appearance for temporal sentence grounding},
  author={Liu, Daizong and Fang, Xiang and Hu, Wei and Zhou, Pan},
  journal={IEEE Transactions on Multimedia},
  volume={25},
  pages={8539--8553},
  year={2023},
  publisher={IEEE}
}

@inproceedings{wang2025taylor,
  title={Taylor series-inspired local structure fitting network for few-shot point cloud semantic segmentation},
  author={Wang, Changshuo and He, Shuting and Fang, Xiang and Wu, Meiqing and Lam, Siew-Kei and Tiwari, Prayag},
  booktitle={Proceedings of the AAAI Conference on Artificial Intelligence},
  volume={39},
  number={7},
  pages={7527--7535},
  year={2025}
}

@inproceedings{wang2025point,
  title={Point clouds meets physics: Dynamic acoustic field fitting network for point cloud understanding},
  author={Wang, Changshuo and He, Shuting and Fang, Xiang and Han, Jiawei and Liu, Zhonghang and Ning, Xin and Li, Weijun and Tiwari, Prayag},
  booktitle={Proceedings of the Computer Vision and Pattern Recognition Conference},
  pages={22182--22192},
  year={2025}
}

@inproceedings{wang2025dypolyseg,
  title={DyPolySeg: Taylor Series-Inspired Dynamic Polynomial Fitting Network for Few-shot Point Cloud Semantic Segmentation},
  author={Wang, Changshuo and Fang, Xiang and Tiwari, Prayag},
  booktitle={Forty-second International Conference on Machine Learning},
  year={2025}
}

@article{wang2026reasoning,
  title={Reasoning beyond points: A visual introspective approach for few-shot 3d segmentation},
  author={Wang, Changshuo and He, Shuting and Fang, Xiang and Hu, Zhijian and Huang, Jia-Hong and Shen, Yixian and Tiwari, Prayag},
  journal={Advances in Neural Information Processing Systems},
  volume={38},
  pages={117394--117414},
  year={2026}
}

@article{wang2026from,
  title={From Coarse to Fine: Deep Prototype Refinement Network for Few-Shot Point Cloud Semantic Segmentation},
  author={Wang, Changshuo and He, Shuting and Fang, Xiang and Li, Weijun and Gao, Xingyu and Liu, Zhonghang and Tiwari, Prayag and Kanoulas, Dimitrios},
  journal={International Conference on Machine Learning},
  year={2026}
}

@article{wang2026topadapter,
  title={TopAdapter: Topology-Aware Prompt Tuning for Efficient Point Cloud Understanding},
  author={Wang, Changshuo and He, Shuting and Fang, Xiang and Li, Weijun and Shen, Yixian and Xu, Mingkun and Sun, Zhongtian and Tiwari, Prayag},
  journal={International Conference on Machine Learning},
  year={2026}
}

@inproceedings{wang2026biologically,
  title={Biologically-Inspired Evolutionary Domain Symbiosis for Few-shot and Zero-shot Point Cloud Semantic Segmentation},
  author={Wang, Changshuo and Hu, Zhijian and Fang, Xiang and Yu, Zai Yang and Wu, Yibin and Xu, Mingkun and Wang, Yusong and Gao, Xingyu and Tiwari, Prayag},
  booktitle={Proceedings of the AAAI Conference on Artificial Intelligence},
  volume={40},
  number={12},
  pages={9666--9674},
  year={2026}
}

@inproceedings{yang2025eood,
  title={EOOD: Entropy-based Out-of-distribution Detection},
  author={Yang, Guide and Hou, Chao and Peng, Weilong and Fang, Xiang and Nie, Yongwei and Zhu, Peican and Tang, Keke},
  booktitle={2025 International Joint Conference on Neural Networks (IJCNN)},
  pages={1--8},
  year={2025},
  organization={IEEE}
}

@inproceedings{wang2025reducing
,
  title={Reducing T-Depth and T-Count in Quantum Multiplication Using Compressor Primitives},
  author={Wang, Siyi and Dutta, Suman and Lee, Wei Jie Bryan and Feng, Jerrie and Fang, Xiang and Chattopadhyay, Anupam},
  booktitle={Proceedings of the Great Lakes Symposium on VLSI 2025},
  pages={35--40},
  year={2025}
}

@inproceedings{lei2025exploring,
  title={Exploring Disentangled Appearance-Motion Contexts for Temporal Activity Localization},
  author={Lei, Huashuo and Cai, Xiaowen and Liu, Daizong and Fang, Xiang and Qu, Xiaoye and Dong, Jianfeng and Yu, Jixiang and Jin, Keyan},
  booktitle={2025 International Joint Conference on Neural Networks (IJCNN)},
  pages={1--8},
  year={2025},
  organization={IEEE}
}

@inproceedings{zhang2025monoattack,
  title={MonoAttack: A Strong Attack Framework with Depth-Migration and Attribute-Tampering for Monocular 3D Object Detection},
  author={Zhang, Xiayue and Lei, Huashuo and Liu, Daizong and Qu, Xiaoye and Fang, Xiang and Guan, Runwei and Jin, Keyan},
  booktitle={2025 International Joint Conference on Neural Networks (IJCNN)},
  pages={1--8},
  year={2025},
  organization={IEEE}
}

@inproceedings{zhang2025manipulating,
  title={Manipulating the Bounding Box: Multimodal Controlled Backdoor Attacks on 3D Visual Grounding Models},
  author={Zhang, Xiayue and Lei, Huashuo and Liu, Daizong and Qu, Xiaoye and Fang, Xiang and Guan, Runwei and Jin, Keyan},
  booktitle={2025 International Joint Conference on Neural Networks (IJCNN)},
  pages={1--8},
  year={2025},
  organization={IEEE}
}

@article{wang2025prototype,
  title={Prototype-driven structure synergy network for remote sensing images segmentation},
  author={Wang, Junyi and Li, Jinjiang and Fan, Guodong and Ju, Yakun and Fang, Xiang and Kot, Alex C},
  journal={IEEE Transactions on Geoscience and Remote Sensing},
  year={2025},
  publisher={IEEE}
}

@inproceedings{wang2025seeing,
  title={Seeing the Overlooked: Bio-Visual Inspired Weak Saliency Feedback Transformer for Person Re-identification},
  author={Wang, Changshuo and He, Shuting and Fang, Xiang and Nan, Fangzhe and Tiwari, Prayag},
  booktitle={Proceedings of the 33rd ACM International Conference on Multimedia},
  pages={3192--3201},
  year={2025}
}

@inproceedings{fang2026align,
  title={To align or not to align: Strategic multimodal representation alignment for optimal performance},
  author={Fang, Wanlong and Zhang, Tianle and Chan, Alvin},
  booktitle={Proceedings of the AAAI Conference on Artificial Intelligence},
  volume={40},
  number={25},
  pages={21056--21064},
  year={2026}
}

@article{liu2023conditional,
  title={Conditional video diffusion network for fine-grained temporal sentence grounding},
  author={Liu, Daizong and Zhu, Jiahao and Fang, Xiang and Xiong, Zeyu and Wang, Huan and Li, Renfu and Zhou, Pan},
  journal={IEEE Transactions on Multimedia},
  volume={26},
  pages={5461--5476},
  year={2023},
  publisher={IEEE}
}

@article{liu2024pandora,
  title={Pandora's box: Towards building universal attackers against real-world large vision-language models},
  author={Liu, Daizong and Yang, Mingyu and Qu, Xiaoye and Zhou, Pan and Fang, Xiang and Tang, Keke and Wan, Yao and Sun, Lichao},
  journal={Advances in Neural Information Processing Systems},
  volume={37},
  pages={52127--52158},
  year={2024}
}

@inproceedings{liu2026attacking,
  title={Attacking Gray-Box Large Vision-Language Models with Adaptive SVD-Structured Adversarial Alignment},
  author={Liu, Daizong and Cai, Xiaowen and Dong, Junhao and Guo, Zhongliang and Qu, Xiaoye and Guan, Runwei and Fang, Xiang and Ye, Dengpan},
  booktitle={International Conference on Machine Learning},
  year={2026}
}

@inproceedings{liu2024unsupervised,
  title={Unsupervised domain adaptative temporal sentence localization with mutual information maximization},
  author={Liu, Daizong and Fang, Xiang and Qu, Xiaoye and Dong, Jianfeng and Yan, He and Yang, Yang and Zhou, Pan and Cheng, Yu},
  booktitle={Proceedings of the AAAI Conference on Artificial Intelligence},
  volume={38},
  number={4},
  pages={3567--3575},
  year={2024}
}

@inproceedings{liu2023hypotheses,
  title={Hypotheses tree building for one-shot temporal sentence localization},
  author={Liu, Daizong and Fang, Xiang and Zhou, Pan and Di, Xing and Lu, Weining and Cheng, Yu},
  booktitle={Proceedings of the AAAI Conference on Artificial Intelligence},
  volume={37},
  number={2},
  pages={1640--1648},
  year={2023}
}

@inproceedings{tang2024reparameterization,
  title={Reparameterization head for efficient multi-input networks},
  author={Tang, Keke and Zhao, Wenyu and Peng, Weilong and Fang, Xiang and Cui, Xiaodong and Zhu, Peican and Tian, Zhihong},
  booktitle={ICASSP 2024-2024 IEEE International Conference on Acoustics, Speech and Signal Processing (ICASSP)},
  pages={6190--6194},
  year={2024},
  organization={IEEE}
}

@article{xiong2024rethinking,
  title={Rethinking video sentence grounding from a tracking perspective with memory network and masked attention},
  author={Xiong, Zeyu and Liu, Daizong and Fang, Xiang and Qu, Xiaoye and Dong, Jianfeng and Zhu, Jiahao and Tang, Keke and Zhou, Pan},
  journal={IEEE Transactions on Multimedia},
  volume={26},
  pages={11204--11218},
  year={2024},
  publisher={IEEE}
}

@inproceedings{tang2025simplification,
  title={Simplification is all you need against out-of-distribution overconfidence},
  author={Tang, Keke and Hou, Chao and Peng, Weilong and Fang, Xiang and Wu, Zhize and Nie, Yongwei and Wang, Wenping and Tian, Zhihong},
  booktitle={Proceedings of the Computer Vision and Pattern Recognition Conference},
  pages={5030--5040},
  year={2025}
}

@article{cai2026towards,
  title={Towards building model/prompt-transferable attackers against large vision-language models},
  author={Cai, Xiaowen and Liu, Daizong and Qu, Xiaoye and Fang, Xiang and Dong, Jianfeng and Tang, Keke and Zhou, Pan and Sun, Lichao and Hu, Wei},
  journal={Advances in Neural Information Processing Systems},
  volume={38},
  pages={174022--174058},
  year={2026}
}

@article{yan2026fit,
  title={Fit the distribution: Cross-image/prompt adversarial attacks on multimodal large language models},
  author={Yan, Hai and Ma, Haijian and Cai, Xiaowen and Liu, Daizong and Yuan, Zenghui and Qu, Xiaoye and Dong, Jianfeng and Guan, Runwei and Fang, Xiang and He, Hongyang and others},
  journal={Advances in Neural Information Processing Systems},
  volume={38},
  pages={75204--75247},
  year={2026}
}

@inproceedings{liu2024towards,
  title={Towards robust temporal activity localization learning with noisy labels},
  author={Liu, Daizong and Qu, Xiaoye and Fang, Xiang and Dong, Jianfeng and Zhou, Pan and Nan, Guoshun and Tang, Keke and Fang, Wanlong and Cheng, Yu},
  booktitle={Proceedings of the 2024 Joint International Conference on Computational Linguistics, Language Resources and Evaluation (LREC-COLING 2024)},
  pages={16630--16642},
  year={2024}
}

@inproceedings{cai2025imperceptible,
  title={Imperceptible Beam-Sensitive Adversarial Attacks for LiDAR-based Object Detection in Autonomous Driving},
  author={Cai, Fuyao and Liu, Daizong and Fang, Xiang and Yu, Jixiang and Tang, Keke and Zhou, Pan},
  booktitle={2025 IEEE International Conference on Multimedia and Expo (ICME)},
  pages={1--6},
  year={2025},
  organization={IEEE}
}

@article{kuai2026dynamic,
  title={Dynamic Graph-enhanced Event Refinement for Temporal Sentence Grounding of Micro-moments},
  author={Kuai, Mingjin and Qin, You and Fang, Xiang and Ji, Wei and Zimmermann, Roger},
  journal={IEEE Transactions on Multimedia},
  year={2026},
  publisher={IEEE}
}

@inproceedings{fang2026towardsicml,
  title={Towards Understanding Modality Interaction in Multimodal Language Models via Partial Information Decomposition},
  author={Fang, Wanlong and Zhang, Tianle and Tao, Wen and Chan, Alvin},
  booktitle={International Conference on Machine Learning},
  year={2026}
}
\end{document}